\documentclass[11pt,letterpaper]{article}
\usepackage[letterpaper,margin=0.98in]{geometry}

\usepackage{microtype}
\usepackage{graphicx}
\usepackage{subcaption} 
\usepackage{booktabs}
\usepackage{amsmath,amsthm}
\usepackage{amssymb}
\usepackage{bm}
\usepackage{bbm}
\usepackage{xspace}
\usepackage{float}
\usepackage{dsfont}
\usepackage{multirow}
\usepackage{comment}
\usepackage{algorithm}
\usepackage{algpseudocode}
\usepackage{hyperref}
\usepackage{colortbl}
\usepackage[table,xcdraw]{xcolor}
\usepackage{arydshln}
\usepackage{adjustbox}
\usepackage{bigints}
\usepackage{mathtools}
\usepackage{enumitem}
\usepackage{textcomp}
\usepackage{color}
\usepackage{tcolorbox}
\usepackage{authblk}

\usepackage[style=alphabetic, maxbibnames=15, giveninits=true, maxcitenames=15, natbib=true,
  maxalphanames=10, backend=biber, sorting=nty, uniquename=false]{biblatex}
\DefineBibliographyStrings{english}{backrefpage = {page},backrefpages = {pages},}
\DeclareNameAlias{sortname}{given-family}
\renewbibmacro{in:}{\ifentrytype{article}{}{\printtext{\bibstring{in}\intitlepunct}}}

\DeclareMathOperator*{\argmax}{argmax}

\DeclareMathOperator{\Eval}{Eval}

\newcommand{\Var}{\operatorname{Var}}
\newcommand{\Cov}{\operatorname{Cov}}

\newcommand{\diag}{\operatorname{diag}}

\newcommand{\norm}[2]{\xmath{\left\| #1 \right\|_{#2}}}

\newcommand{\innerprod}[2]{\langle #1, #2 \rangle}

\newcommand{\1}{\blmath{1}}

\long\def\red#1{\bgroup\color{red}#1\egroup}
\long\def\blue#1{\bgroup\color{blue}#1\egroup}

\newcommand{\abs}[1]{\left| #1 \right|}

\newcommand{\indic}[1]{\mathds{1}\{#1\}}
\renewcommand{\Re}{\mathbb{R}}

\newcommand{\grad}{\xmath{\nabla}}

\newcommand{\EV}[2]{\xmath{\mathbb{E}_{#1}\left[ #2 \right]}}

\newcommand{\KL}[2]{{\mathrm{KL}}\left(#1 \,\|\, #2\right)}

\newcommand{\xmath}[1] {\ensuremath{#1}\xspace}
\newcommand{\blmath}[1] {\xmath{\bm{#1}}}
\newcommand{\paren}[1]{\left( #1 \right)}
\newcommand{\parenb}[1]{\big( #1 \big)}

\newcommand{\highlight}[2]{%
  \begingroup
  \setlength{\fboxsep}{0pt}
  \colorbox{#1}{\strut\kern0.15em #2\kern0.15em}%
  \endgroup
}

\renewcommand{\S}{\xmath{\mathcal{S}}}

\newcommand{\Vocab}{\mathcal{V}}

\newcommand{\placeholder}{\mathord{\makebox[\widthof{$x$}]{$\cdot$}}}

\newcommand{\hist}{\xmath{h}}
\newcommand{\histt}{\xmath{\hist^{(t)}}}
\newcommand{\histtt}{\xmath{\hist^{(t+1)}}}
\newcommand{\nof}{\raisebox{.25ex}{$\scriptscriptstyle\backslash$}f}

\newcommand{\thistsup}[1]{\xmath{\mathrlap{\phantom{h}_{\nof}}\smash[t]{h^{\raisebox{.15em}{$\scriptstyle #1$}}}\vphantom{h^{#1}}\vphantom{h_f}}}

\newcommand{\thisttt}{\xmath{\thistsup{(t+1)}}}
\newcommand{\thisttj}{\xmath{\thistsup{(t,j)}}}

\newcommand{\ft}{\xmath{f^{(t)}}}
\newcommand{\fonet}{\xmath{f^{(1:t)}}}
\newcommand{\pit}{\xmath{\pi^{(t)}}}
\newcommand{\pitt}{\xmath{\pi^{(t+1)}}}

\newcommand{\rhot}{\xmath{\rho^{(t)}}}
\newcommand{\rhott}{\xmath{\rho^{(t+1)}}}

\newcommand{\rlm}{\xmath{r_{<m}}}

\newcommand{\calA}{\xmath{\mathcal{A}}}

\newcommand{\rv}[1]{\xmath{\mathsf{#1}}}
\newcommand{\Rollout}{\rv{R}}     
\newcommand{\Tok}{\rv{V}}         
\newcommand{\simplex}{\Delta^{+}(\Vocab)}

\newcommand{\sigtt}{\xmath{s^{(t+1)}}}
\newcommand{\rollt}{r^{(t)}}

\newcommand{\rollonet}{r^{(1:t)}}

\newcommand{\mypromptbox}[1]{
\begin{tcolorbox}[
  width=0.92\linewidth,
  colback=gray!5,
  colframe=gray!50,
  boxrule=0.4pt,
  arc=2pt,
  left=7pt,
  right=7pt,
  top=6pt,
  bottom=6pt
]
#1
\end{tcolorbox}
}

\newcounter{myparagraph}[subsubsection]

\newtheorem{theorem}{Theorem}[section]
\newtheorem*{theorem*}{Theorem}
\newtheorem{corollary}{Corollary}[theorem]
\newtheorem{lemma}[theorem]{Lemma}
\newtheorem{proposition}[theorem]{Proposition}
\newtheorem*{proposition*}{Proposition}

\theoremstyle{remark}
\newtheorem{remark}[theorem]{Remark}
\hypersetup{colorlinks=true, urlcolor=blue, linkcolor=blue, citecolor=blue}

\title{Guiding LLMs at Inference Time via Bregman Tilting \vspace{3mm}

\footnotetext{Corresponding author: \texttt{jblock@utexas.edu}}}

\renewcommand*{\Affilfont}{\normalsize}

\renewcommand*{\Affilfont}{\normalsize}

\makeatletter
\renewcommand\AB@affilsepx{\quad \protect\Affilfont}
\renewcommand\AB@affilsep{\protect\Affilfont}
\makeatother

\author[$\ast$]{Jacob L. Block}
\author[$\ast \, \dagger$]{Aryan Mokhtari}
\author[$\ast$]{Sanjay Shakkottai}

\affil[$\ast$]{UT Austin}
\affil[$\dagger$]{Google Research}

\date{}

\begin{document}

\maketitle

\begin{abstract}
We study inference-time guidance for large language models (LLMs) in iterative problem solving, where the LLM generates an attempt, receives text feedback from the environment, and tries again. While input prompting, which puts prior attempts and feedback in context, is a standard approach, it gives no control over how the LLM uses the feedback in its next attempt. Thus, we introduce \emph{Bregman tilting}, which complements any input-prompting approach by directly modifying the LLM's next-token distribution. At each decoding step, it scores each token by its change in log-probability when the feedback is added to a prompt that contains the prior attempt, and updates the distribution via a mirror ascent step on the expected score, regularized by a Tsallis-entropy-based family of Bregman divergences indexed by $\alpha \in [0,1]$. At $\alpha = 1$ this recovers the exponential tilt of prior methods, while $\alpha < 1$ changes the divergence geometry to discount the influence of unlikely tokens under input prompting. The resulting update damps each token's tilt by its input-prompting likelihood, and it locally improves the true correctness probability whenever the score positively covaries with correctness under the input-prompting distribution sharpened to a lower temperature. Experiments on coding and math show that Bregman tilting outperforms input prompting in every setting, while $\alpha = 0$ performs best, beating the exponential tilt at $\alpha = 1$ and highlighting the value of our framework as a whole.
\end{abstract}

\newpage
\section{Introduction}

A common paradigm for solving reasoning problems with large language models (LLMs) is iterative generation at test time: the model generates an attempt, receives feedback on it in the form of text, and generates again. The feedback can come from the environment, such as a compiler and unit tests \citep{chen:2024:self-debug} or a proof verifier \citep{lin:2026:goedel-prover-v2}, from a human user \citep{buening:2026:aligning-llm-user-int}, from another LLM \citep{olausson:2024:is-self-repair-silver-bullet}, or from the solver itself \citep{madaan:2023:self-refine}. In each case, the goal is for the LLM to produce a better attempt at each round given the text history of its previous attempts and their feedback.

We study a training-free, inference-time setting, where the underlying model parameters are kept frozen. In this setting, the standard approach is \emph{input prompting}, which places the previous attempts and their feedback in the prompt given to the LLM at the next round and resamples \citep{madaan:2023:self-refine, shinn:2023:reflexion, venkatraman:2025:rsa, pandita:2025:pro-refine}. While this is simple to implement and can improve performance round over round, input prompting updates the distribution over rollouts only through the LLM's context. This treats the LLM as a black-box sampler conditioned on its prompt and gives no control over how strongly or faithfully the LLM incorporates the feedback into its next attempt. We address this issue with \emph{Bregman tilting}, a principled inference-time framework that complements a given input-prompting approach by using text feedback to explicitly modify the LLM’s output distribution.

\textbf{Contributions.} After each round, we search for a distribution over the next round's rollouts which improves the probability of producing an acceptable response (Section \ref{sec:pipeline}). We propose a per-token feedback score that serves as a proxy for this intractable objective, and a novel guidance method that updates the LLM's next-token distribution to increase the expected score. We detail these contributions below.

At any prefix, the optimal sampler maximizes the probability of generating a token which leads to an acceptable rollout. However, this objective is infeasible without access to a verifier of correctness. Instead, we define a per-token feedback score as a surrogate, computed as the change in the log-probability of each candidate next token when the feedback is added to the prompt \eqref{eq:score}. It compares the LLM's next-token distribution under the prompt that shows the previous attempts and their feedback to its distribution under a reference prompt that shows the same previous attempts but without the feedback. Similar contrastive scores appear in contrastive decoding \citep{li:2023-contrastive-decoding}, context-aware decoding \citep{shi:2024:context-aware-decoding}, classifier-free guidance for LLMs \citep{sanchez:2024:stay-on-topic-cfg}, and self-distillation \citep{hubotter:2026:rl-self-dist}, which contrast against a weaker LLM or the same LLM prompted without some added context. In iterative refinement, however, a reference prompt that omits the previous attempt can reward tokens for agreeing with that attempt regardless of correctness. Because the two prompts differ only in the feedback, our score isolates its effect, while the question-only reference prompt construction of prior work can fail to outperform vanilla input prompting (Table \ref{tab:reference}).

We then guide the LLM's output distribution at inference time with our proposed Bregman tilting framework. At each decoding step, Bregman tilting takes a single mirror ascent step on the objective of maximizing the expected score over next tokens \eqref{eq:mirror-step}. This regularizes the update using the Bregman divergence to the input-prompting LLM's next-token distribution, penalizing deviations in terms of the geometry induced by the mirror map.

We consider a family of mirror maps parameterized by $\alpha \in [0,1]$ based on the Tsallis entropy which smoothly interpolates between the log-barrier at $\alpha=0$ and the negative Shannon entropy at $\alpha=1$. At $\alpha=1$, the Bregman divergence is the KL divergence, and the update exponentially tilts the input-prompting LLM with the feedback score, a common update form for inference-time guidance \citep{shi:2024:context-aware-decoding,sanchez:2024:stay-on-topic-cfg,ho:2022:classifier-free-guidance}. However, setting $\alpha < 1$ leads to a generalized tilting update which discounts the effects of tokens which the input-prompting LLM considers unlikely.

We show that, in the limit as a token's probability under the input-prompting distribution goes to zero, Bregman tilting changes its log-probability by the exponential tilt's amount, damped by a factor of its input-prompting probability raised to the power $1 - \alpha$ (Proposition \ref{prop:log-ratio}). We further analyze the non-asymptotic case, upper-bounding the factor by which the update can increase the probability of any token whose input-prompting probability lies below a threshold set by the step size and the token's score. For $\alpha < 1$, this factor approaches one as the token's input-prompting probability decreases, whereas at $\alpha = 1$ it grows exponentially in the score regardless of the token's probability (Proposition \ref{prop:cap}). Thus, Bregman tilting incorporates the score in the updated distribution while decaying the tilt strength in terms of each token's probability under input prompting, providing effective guidance while avoiding extreme jumps in the probabilities of tokens which are disfavored by input prompting. Further, we derive a sufficient condition for when the update, which optimizes a surrogate, also increases the true per-token objective of maximizing the probability of leading to an acceptable rollout. We show that if the feedback score positively covaries with the probability of acceptance over tokens drawn from the input-prompting distribution sharpened with temperature $1/(2-\alpha) \leq 1$, then the Bregman tilting update with small step size provably increases this objective (Theorem \ref{thm:rate}). Smaller $\alpha$ amplifies this sharpening, concentrating the PMF on the likely tokens and lessening the dependence of this condition on potentially erroneous scores on unlikely tokens.

\textbf{Experiments.} We apply Bregman tilting at $\alpha \in \{0,1\}$ with two input-prompting methods, Self-Refine \citep{madaan:2023:self-refine} and Recursive Self-Aggregation (RSA) \citep{venkatraman:2025:rsa}, with Qwen3-4B-Instruct-2507 \citep{yang:2025:qwen3}. We use two code benchmarks, LiveCodeBench and MBPP+, with feedback from running each program on tests, and two math benchmarks, AIME 2025 and HMMT February 2025, with feedback from a stronger reviewer LLM. In all eight settings, Bregman tilting with either value of $\alpha$ improves performance over input prompting, and $\alpha = 0$ achieves the best mean pass@1 and pass@16 across all benchmarks and input-prompting methods, showing the key role played by $\alpha$ in shaping the Bregman tilting update geometry beyond the exponential tilting form of $\alpha=1$.

\subsection{Related Work}
\label{sec:related}

A large body of work improves LLM reasoning at inference time by alternating between generation and feedback. Self-Refine \citep{madaan:2023:self-refine} has the LLM critique and revise its own output, Reflexion \citep{shinn:2023:reflexion} keeps verbal reflections on past failures in context, and CRITIC \citep{gou:2024:critic} and Self-Debugging \citep{chen:2024:self-debug} incorporate feedback from external tools such as interpreters and unit tests. Recursive Self-Aggregation \citep{venkatraman:2025:rsa} refines a population of attempts by aggregating subsets of prior attempts to show to the model at each round. ProRefine \citep{pandita:2025:pro-refine} considers a prompt-optimization approach, using one LLM to generate natural language feedback and another to refine the prompt given to the solver at the next round. A related line of work optimizes the prompt itself through evolutionary search \citep{fernando:2024:promptbreeder,agrawal:2026:gepa}. All of these methods act on the LLM only through its context and treat it as a black-box sampler. In contrast, Bregman tilting directly modifies the LLM's next-token distribution and complements any input-prompting method, which we demonstrate on top of Self-Refine and RSA.

Prior inference-time guidance methods modify the LLM's next-token distribution directly, most commonly via exponential tilting. This form arises in posterior sampling, where Bayes' rule expresses the target as the original, or prior, distribution tilted by a learned classifier \citep{dhariwal:2021:classifier-guidance,yang:2021:fudge,block:2026:T3-unlearning}, and in KL-regularized RL, where the optimal policy tilts the original distribution by a per-token reward or value function \citep{mudgal:2024:controlled-decoding,xu:2025:genarm,rashid:2025:tokenwise-reward-guiding}. Related reward-guided decoding methods apply the same tilt, adding a reward model's score of each candidate token to its log-probability \citep{khanov:2024:args}. Other guidance methods take the score to be the log-ratio of two LLM next-token distributions. These can come from a strong and a weak LLM, as in contrastive decoding \citep{li:2023-contrastive-decoding}, or from the same LLM under two prompts: with and without retrieved context \citep{shi:2024:context-aware-decoding}, with and without an attribute to condition on \citep{ho:2022:classifier-free-guidance,sanchez:2024:stay-on-topic-cfg}, or against a deliberately corrupted instruction in instructive decoding \citep{kim:2024:instructive-decoding}. Bregman tilting differs from these methods in two ways. First, its reference prompt contains the question and the previous rollout but not the feedback, so the score isolates the effect of the feedback rather than rewarding tokens which simply copy the previous rollout shown to the LLM. Second, it replaces the exponential tilt with the solution of a mirror ascent step under a family of Bregman divergences, which recovers the exponential tilt as a special case while inducing a family of more performant updates.

Bregman tilting performs a single step of mirror ascent \citep{nemirovsky:1983:problem-complexity,beck:2003:mirror-descent-nonlinear-proj}, which generalizes the gradient ascent update to non-Euclidean geometries via the Bregman divergence of a strictly convex function, called the mirror map. We define a family of mirror maps $\psi_\alpha$ for $\alpha \in [0, 1]$. On the interior $\alpha \in (0, 1)$, $\psi_\alpha$ is, up to an additive constant, the negative Tsallis entropy of index $\alpha$ \citep{tsallis:1988:gen-entropy}, scaled by $1/\alpha$, while at the endpoints we define $\psi_0$ as the log barrier and $\psi_1$ as the negative Shannon entropy respectively. This is a natural extension of the Tsallis family: up to additive constants, $\psi_0$ and $\psi_1$ are the limits of $\psi_\alpha$ as $\alpha \to 0$ and $\alpha \to 1$, and the induced Bregman divergence of $\psi_\alpha$ varies continuously in $\alpha \in [0,1]$ (Proposition \ref{prop:tsallis-bregman-limits}). This family of mirror maps is standard in online learning over the simplex, notably for multi-armed bandits \citep{wei:2018:more-algs-for-bandits,zimmert:2021:tsallis}. We bring this geometry to inference-time guidance, leading to our generalized tilting update.

Lastly, while we focus on inference-time methods, there is substantial prior work on training-based methods which update the LLM parameters. Reinforcement learning with a KL penalty to the original LLM, whose optimal policy is an exponential tilt of the original LLM by the reward, maximizes a reward that is learned, implicit in preference data, or given by a verifier \citep{schulman:2017:ppo,rafailov:2023:dpo,shao:2024:deepseekmath}. When text feedback or reference solutions are available, self-distillation methods instead train the LLM to match its own next-token distribution when it is additionally given this privileged information in context \citep{zhao:2026:self-distill-reasoner,penaloza:2026:priv-info-dist,hubotter:2026:rl-self-dist}, extending context distillation \citep{snell:2022:learning-by-distill-context,ye:2026:onpolicy-context-distillation-llm}.

With the reverse KL divergence as its objective, self-distillation is a policy gradient method whose per-token advantage is the log-ratio of the LLM's next-token distributions with and without the feedback in its prompt \citep{hubotter:2026:rl-self-dist}, a contrastive signal of the same form as our score. However, self-distillation uses this signal to move the effect of the privileged information into the weights, so that the LLM prompted with the question alone matches the LLM when given the feedback in-context. Its target is therefore the feedback-prompted LLM itself, and its goal is to learn weights that generalize to new questions. Test-time variants apply the same update repeatedly to a single question \citep{hubotter:2026:rl-self-dist}, but they still aim to distill the feedback into the weights and sample rollouts by prompting the LLM with the question alone. In contrast, our setting keeps the weights frozen, places the history of attempts and feedback in the prompt, and focuses on how to utilize this history to further update the feedback-prompted LLM distribution when generating the next attempt. Bregman tilting therefore starts from the feedback-prompted LLM, which self-distillation uses as its training target, and tilts it further along the score without any backpropagation through the LLM.

\textbf{Notation.} Let $\Vocab$ be the vocabulary and $M$ the maximum sequence length, and write $\Vocab^M$ for the set of token sequences of length $M$.\footnote{A sequence shorter than $M$ is represented with an end-of-sequence token followed by padding, and we implicitly restrict our focus to autoregressive PMFs which generate only padding after this token.} For a sequence $x$, $x_m$ is its $m$th token and $x_{<m} \coloneq (x_1, \ldots, x_{m-1})$ is its prefix of length $m - 1$ (empty for $m = 1$); $(x, y)$ is the concatenation of two sequences, and $[n] \coloneq \{1, \ldots, n\}$.

We write $\simplex \coloneq \{p \in \Re^{\Vocab} : \sum_{v \in \Vocab} p(v) = 1, \ p(v) > 0 \text{ for every } v \in \Vocab\}$ for the set of probability mass functions (PMFs) over the vocabulary with full support. Every next-token PMF of an LLM lies in $\simplex$, since it is a softmax over the LLM's logits. An \emph{autoregressive} PMF $\rho$ assigns a next-token PMF $\rho(\placeholder \mid h) \in \simplex$ to every prefix of tokens $h$. Sampling $n$ tokens one at a time from $\rho$ after the prefix $h$ induces a PMF on continuations $r \in \Vocab^n$, $\rho(r \mid h) \coloneq \prod_{m=1}^{n} \rho\paren{r_m \mid h, r_{<m}}$. We overload notation and write both the PMF over next tokens and over continuations as $\rho(\placeholder \mid h)$, where the surrounding context makes the distinction clear. We write $\rho(r)$ for the PMF over rollouts with the empty prefix. We use $\pi$ to denote the specific autoregressive PMF induced by the LLM; $\rho$ refers to an arbitrary autoregressive PMF that need not be the output of any LLM. In contrast, $p$ and $q$ denote standalone PMFs in $\simplex$.

We keep the following conventions throughout. Lower-case letters denote realized quantities: $h$ is a prompt, $r$ a rollout, $f$ a feedback text, and $v$ a token. The sans-serif capitals $\Rollout$ and $\Tok$ denote a random rollout and a random token. $\mathbb{P}$ and $\mathbb{E}$ are the probability and expectation operators, always with the sampling law in the subscript, as in $\mathbb{P}_{\Rollout \sim \rho}(\Rollout \in \calA)$ or $\EV{\Tok \sim q}{\placeholder}$. A subscript on a sequence or a PMF indexes a token position, as in $r_m$ or $r_{<m}$; the round $t$ and, when several chains run in parallel, the chain $j$ are superscripts in parentheses, as in $h^{(t)}$ and $r^{(t, j)}$. Vectors indexed by $\Vocab$, such as PMFs, are written as functions of $v$, so every operation on them is componentwise, and $\1 \in \Re^{\Vocab}$ is the all-ones vector.

\section{Inference-Time Pipeline}
\label{sec:pipeline}

We first describe the inference-time pipeline which encompasses our method. Given an LLM $\pi$, we aim to solve a fixed question by iteratively generating answer attempts, which we refer to as rollouts, and receiving feedback on them. We assume access to a feedback mechanism $\Eval$ that takes a rollout $r$ and returns a text $\Eval(r)$ about its quality. The mechanism can be a system such as a compiler or a test suite, another LLM, or a human.

At each round of the iterative pipeline, we prompt the LLM to generate rollouts, observe the feedback, and then update the distribution used for the next round's rollouts. We focus on tasks such as math and code, where correctness of each rollout is deterministically verifiable, for example by comparing its final answer to a reference or running its program on a set of tests, even though the pipeline itself cannot access this verifier. Our goal is simply to produce acceptable rollouts for the task.

\subsection{Setup}
\label{sec:setup}

We first define the pipeline mechanics. The pipeline runs for $T$ rounds with $Q$ chains per question. A chain is one sequence of attempts at the question, in which each round produces one rollout and its feedback. We write $r^{(t,j)}$ and $f^{(t,j)} = \Eval(r^{(t,j)})$ for the rollout and feedback of chain $j \in [Q]$ at round $t$. For brevity, we describe the pipeline for a single chain and drop the chain index $j$.

After round $t$, we have available the LLM $\pi$, the zero-shot prompt $h^{(1)}$, which contains the question and any instructions on the format of the answer, and the rollouts $r^{(1)}, \ldots, \rollt$ with their feedback $f^{(1)}, \ldots, \ft$. We use this information to construct an autoregressive PMF $\rhott$ for the next round, with the goal of increasing the probability of generating an acceptable rollout.

Formally, we define $\calA \subseteq \Vocab^{M}$ as the fixed set of acceptable rollouts. For an autoregressive PMF $\rho$, we define the acceptance probability as
\begin{equation}
\label{eq:objective}
A(\rho) \coloneq \mathbb{P}_{\Rollout \sim \rho}\paren{\Rollout \in \calA},
\end{equation}
and more generally, for a prefix $x$ and autoregressive PMF $\rho$, we write the prefix-conditioned version
\begin{equation*}
A(\rho \, ; x) \coloneq \mathbb{P}_{\Rollout \sim \rho(\placeholder \mid x)}\paren{(x, \Rollout) \in \calA}.
\end{equation*}
For the empty prefix $x$, $A(\rho) = A(\rho \, ; x)$. Our goal is then to construct the PMF $\rhott$ for the next round so that $A(\rhot) \leq A(\rhott)$. However, since $\calA$ is unknown, we cannot optimize \eqref{eq:objective} directly. Instead, we use the rollouts and feedback observed so far to update the LLM's rollout distribution with the goal of increasing the value of $A(\placeholder)$ across rounds.

\subsection{Input Prompting}
\label{sec:input-prompting}

We describe the standard input-prompting approach which our Bregman tilting method is designed to complement. The simplest way to construct $\rhott$ is input prompting, which places the observed rollouts and their feedback in the prompt of the next round. At round 1 there is no feedback yet, and prompting the LLM with the zero-shot prompt gives the autoregressive PMF
\begin{equation}
\label{eq:zero-shot}
\rho^{(1)}(r) = \pi(r \mid h^{(1)}).
\end{equation}
An input-prompting method is defined by its prompt update, which constructs the next prompt $\histtt$ from the previous prompts, rollouts, and feedback \citep{madaan:2023:self-refine,shinn:2023:reflexion,venkatraman:2025:rsa}. Formally, the method defines a function $\mathrm{PromptUpdate}\paren{\placeholder}$ and sets
\begin{equation*}
\histtt = \mathrm{PromptUpdate}\paren{h^{(1:t)}, \rollonet, \fonet},
\qquad
\rhott(r) = \pi(r \mid \histtt).
\end{equation*}
In RSA \citep{venkatraman:2025:rsa}, the update can also depend on the rollouts of the other chains of the question, but we describe the single-chain case here for brevity. We write $\pit$ for the \emph{input-prompting LLM} at round $t$, the autoregressive PMF obtained from the same LLM $\pi$ by fixing the prompt $\histt$. Formally, for every prefix $x$ of a rollout, we write
\begin{equation}
\label{eq:input-prompting-llm}
\pit(\placeholder \mid x) \coloneq \pi(\placeholder \mid \histt, x).
\end{equation}
Thus, input prompting sets $\rhot = \pit$ for each round $t$. This updates each $\rhot$ only through the context of the LLM, allowing the model to see its past mistakes and reason about a better solution. While we cannot guarantee that the LLM improves its rollouts based on implicit reasoning given prior attempts, prior work on iterative prompting \citep{madaan:2023:self-refine, shinn:2023:reflexion, venkatraman:2025:rsa} as well as our experiments in Section \ref{sec:experiments} demonstrate that existing iterative prompting methods improve performance round over round and significantly outperform the round-1 zero-shot model $\pi^{(1)}$. Thus, it serves as a sound starting point for our proposed framework.

\section{Our Proposed Framework: Bregman Tilting}
\label{sec:output-update}

Although input prompting lets the LLM see its prior mistakes in the updated context $\histtt$, it gives no control over how the LLM uses this information in its next rollout. Therefore, we complement a given input-prompting method by also directly modifying the next-token PMF of the LLM under the prompt $\histtt$. To do so, we define a feedback score that measures how the feedback changes the likelihood of each candidate token, acting as a surrogate for the true acceptance probability. We then propose \emph{Bregman tilting}, an inference-time guidance framework that takes a mirror ascent step from the input-prompting PMF, with the expected feedback score as the objective and a Tsallis-entropy-based mirror map (Section \ref{sec:mirror-ascent}).

\subsection{Per-Token Surrogate Objective}
\label{sec:per-token-surrogate}

We first define the per-token counterpart $\bar a$ of the sequence-level objective $A$ in \eqref{eq:objective}. Fix the prompt $\histtt$, and consider a prefix $x$ of the rollout being generated. For each token $v$, define the probability that the input-prompting LLM $\pitt$ defined in \eqref{eq:input-prompting-llm} completes the prefix $(x, v)$ with a rollout in $\calA$ as
\begin{equation}
\label{eq:token-value}
a(v \, ; x) \coloneq A\paren{\pitt \, ; (x, v)}.
\end{equation}
For any autoregressive PMF $\rho$, we define the per-token objective $\bar a(\rho \, ; x)$ as the expected value of $a(\Tok \, ; x)$ under its next-token PMF $\rho(\placeholder \mid x)$,
\begin{equation}
\label{eq:objective-token}
\bar a(\rho \, ; x) \coloneq \EV{\Tok \sim \rho(\placeholder \mid x)}{a(\Tok \, ; x)}.
\end{equation}
Importantly, the gain in the sequence-level objective $A$ of \eqref{eq:objective} over the input-prompting LLM $\pitt$ can be expressed as the sum of expected per-token gains in $\bar a$.
\begin{proposition}
\label{prop:pdl}
For every autoregressive PMF $\rho$,
\begin{equation}
\label{eq:pdl}
A(\rho) - A(\pitt) = \sum_{m=1}^{M} \EV{\Rollout \sim \rho}{ \bar a\paren{\rho \, ; \Rollout_{<m}} - \bar a(\pitt \, ; \Rollout_{<m})}.
\end{equation}
\end{proposition}
It therefore suffices to study the update at a single position $m$ with a fixed prefix $\rlm$. At this prefix, $\bar a(\rho \, ; \rlm)$ depends on $\rho$ only through its next-token PMF $\rho(\placeholder \mid \rlm)$, so we treat this PMF as a free variable $p \in \simplex$ and define
\begin{equation*}
A_m(p) \coloneq \EV{\Tok \sim p}{a(\Tok \, ; \rlm)} = \sum_{v \in \Vocab} p(v) \, a(v \, ; \rlm).
\end{equation*}
The subscript $m$ marks the dependence on the fixed prefix $\rlm$, so that $\bar a(\rho \, ; \rlm) = A_m\paren{\rho(\placeholder \mid \rlm)}$.

While $A_m$ is the ideal per-token objective, like $A$ it is unobserved. We thus replace it with a tractable surrogate. We define a per-token feedback score $\sigtt(v \mid \rlm)$ which computes the log-probability gain for the token $v$ when the LLM $\pi$ is given the updated prompt $\histtt$, which contains the previous rollout $\rollt$ and its feedback $\ft$, compared to when it is given a modified reference prompt $\thisttt$ containing $\rollt$ \emph{but not the feedback $\ft$}. Formally, for a token $v$ and prefix $\rlm$,
\begin{equation}
\label{eq:score}
\sigtt(v \mid \rlm) \coloneq \ln \pi(v \mid \histtt, \rlm) - \ln \pi(v \mid \thisttt, \rlm).
\end{equation}
See Appendix \ref{app:exp-prompts} for the exact prompts we use. Going forward, we abbreviate the input-prompting PMF and the score given the fixed prefix $\rlm$ as
\begin{equation}
\label{eq:shorthand}
\pi_m \coloneq \pitt(\placeholder \mid \rlm) \in \simplex,
\qquad
s_m \coloneq \sigtt(\placeholder \mid \rlm) \in \Re^{\Vocab}.
\end{equation}

This score $s_m(v)$ is positive when the feedback makes $v$ more likely and negative when it makes $v$ less likely at the prefix $\rlm$. For each next token of the round-$(t+1)$ rollout, we use this score in place of $a$ in \eqref{eq:objective-token}, replacing the intractable objective with the following surrogate.
\begin{equation}
\label{eq:surrogate}
\tilde A_m(p) \coloneq \EV{\Tok \sim p}{s_m(\Tok)} = \sum_{v \in \Vocab} p(v) \, s_m(v).
\end{equation}
Since $\tilde A_m$ is a surrogate, we do not maximize it outright, as this would simply place all probability mass on the highest scoring token. Instead, we search for a distribution which remains close to the input-prompting PMF $\pi_m$ while increasing the value of $\tilde A_m$, leading to our mirror-ascent-based Bregman tilting update.

\subsection{Bregman Tilting Update}
\label{sec:mirror-ascent}

The standard way to perturb $\pi_m$ toward the tokens that the feedback favors is through exponential tilting, which returns the updated next-token PMF $\hat p_m(v) \propto \pi_m(v) \, e^{\eta \, s_m(v)}$ \citep{shi:2024:context-aware-decoding,ho:2022:classifier-free-guidance,sanchez:2024:stay-on-topic-cfg}. It reweights the probability assigned to each token by the factor $e^{\eta \, s_m(v)}$, potentially increasing the probability of a token that $\pi_m$ considers unlikely by a large factor. We instead propose \emph{Bregman tilting}, which returns the output of a single mirror ascent step on the surrogate objective $\tilde A_m$. We consider a family of mirror maps $\psi_{\alpha}$ for $\alpha \in [0,1]$ which strictly generalize the exponential tilting update and result in a \emph{$\pi_m$-dependent tilt} which adapts the update geometry to better resist changes to low-probability tokens under $\pi_m$.

Treating the optimization variable $p$ as a vector of next-token probabilities, we can write 
\begin{equation*}
    \tilde A_m(p) = \innerprod{s_m}{p}.
\end{equation*}
We maximize this objective over the simplex $\simplex$, regularized by the Bregman divergence $D_\psi(p,q)$ of a strictly convex, differentiable function $\psi$, defined by
\begin{equation}
\label{eq:bregman}
D_\psi(p,q) \coloneq \psi(p) - \psi(q) - \innerprod{\grad \psi(q)}{p - q}.
\end{equation}
The \emph{Bregman tilting} update is then
\begin{equation}
\label{eq:mirror-step}
\hat p_m \coloneq \argmax_{p \in \simplex} \; \innerprod{s_m}{p} - \frac{1}{\eta} D_\psi(p, \pi_m).
\end{equation}
Since $s_m = \grad \tilde A_m(p)$, Equation \eqref{eq:mirror-step} can be interpreted as a single step of mirror ascent \citep{nemirovsky:1983:problem-complexity} applied to the objective $\tilde A_m(p)$. We consider a particular family of mirror maps $\psi_\alpha$ parameterized by $\alpha \in [0,1]$. On the interior $\alpha \in (0,1)$ we define
\begin{equation}
\label{eq:tsallis}
\psi_\alpha(p) \coloneq -\frac{1}{\alpha (1 - \alpha)} \sum_{v \in \Vocab} p(v)^{\alpha}.
\end{equation}
This is the negative Tsallis entropy of index $\alpha$ \citep{tsallis:1988:gen-entropy}, scaled by $1/\alpha$ and shifted by a constant. We then define the endpoints $\psi_0$ and $\psi_1$ as the log-barrier and negative Shannon entropy respectively,
\begin{equation}
\label{eq:endpoints}
\psi_0(p) \coloneq -\sum_{v \in \Vocab} \ln p(v),
\qquad
\psi_1(p) \coloneq \sum_{v \in \Vocab} p(v) \ln p(v).
\end{equation}
By design, this family of strictly convex functions $\psi_\alpha$ for $\alpha \in [0,1]$ is continuous in terms of the Bregman divergence, the key operation which defines the Bregman tilting update in Equation \eqref{eq:mirror-step}.
\begin{proposition}
\label{prop:tsallis-bregman-limits}
Consider the family of mirror maps $\psi_{\alpha}$ defined by Equations \eqref{eq:tsallis}, \eqref{eq:endpoints}, and arbitrary PMFs $p,q \in \simplex$.
\begin{enumerate}[label=(\roman*), topsep=1pt, itemsep=1pt]
\item On the interior $\alpha \in (0,1)$, the Bregman divergence of $\psi_\alpha$ is
\begin{equation}
\label{eq:tsallis-bregman}
D_{\psi_\alpha}(p, q) = \frac{1}{\alpha (1 - \alpha)} \sum_{v \in \Vocab} \paren{q(v)^\alpha - p(v)^\alpha + \alpha \, q(v)^{\alpha - 1} \paren{p(v) - q(v)}}.
\end{equation}
\vspace{-5mm}
\item The Bregman divergence $D_{\psi_\alpha}$ is a continuous function of $\alpha$ at the endpoints $\{0,1\}$, with limits
\begin{align}
\lim_{\alpha \to 0} D_{\psi_\alpha}(p, q) &= D_{\psi_0}(p, q) = \sum_{v \in \Vocab} \paren{\frac{p(v)}{q(v)} - \ln \frac{p(v)}{q(v)} - 1}, \label{eq:limit-logbar} \\
\lim_{\alpha \to 1} D_{\psi_\alpha}(p, q) &= D_{\psi_1}(p, q) = \sum_{v \in \Vocab} p(v) \ln \frac{p(v)}{q(v)} = \KL{p}{q}. \label{eq:limit-kl}
\end{align}
\end{enumerate}
\end{proposition}
\vspace{-5mm}
For this family of Bregman divergences, we compute the update rule below.
\begin{proposition}
\label{prop:closed-form}
For $\alpha \in [0,1]$ and $\eta > 0$, Equation \eqref{eq:mirror-step} with the mirror map $\psi_\alpha$ has a unique solution, which we denote $\hat p_m(\placeholder \, ;\alpha)$. Moreover, there exists a scalar $\hat\lambda$ such that $(p,\lambda) = (\hat p_m(\placeholder \, ;\alpha), \hat\lambda)$ is the unique pair in $\simplex \times \Re$ which satisfies
\begin{equation}
\label{eq:stationarity}
\grad \psi_\alpha(p) = \grad \psi_\alpha(\pi_m) + \eta \, s_m - \lambda \1.
\end{equation}
\end{proposition}
While Proposition \ref{prop:closed-form} implicitly defines the update, we further compute it in closed form. For $\alpha = 1$, the Bregman divergence $D_{\psi_1}(p, q)$ is the KL divergence $\KL{p}{q}$ (Equation \eqref{eq:limit-kl}), so Equation \eqref{eq:mirror-step} is the familiar KL-regularized reward maximization problem with the score $s_m$ as the reward \citep{rafailov:2023:dpo}. Substituting $[\grad \psi_1(p)](v) = \ln p(v) + 1$ into Equation \eqref{eq:stationarity} gives
\begin{equation}
\label{eq:closed-form-kl}
\hat p_m(v \, ; 1) = \pi_m(v) \, e^{\eta \, s_m(v) - \hat \lambda}.
\end{equation}
Thus, $\alpha=1$ recovers standard exponential tilting, reweighting the probability of each token under $\pi_m$ by $e^{\eta \, s_m(v) - \hat \lambda}$. For $\alpha \in [0, 1)$, substituting $[\grad \psi_\alpha(p)](v) = -\tfrac{1}{1-\alpha}p(v)^{\alpha - 1}$ into Equation \eqref{eq:stationarity} gives
\begin{equation}
\label{eq:closed-form}
\hat p_m(v \, ; \alpha) = \frac{\pi_m(v)}{\paren{1 + (1-\alpha) \pi_m(v)^{1 - \alpha} \paren{\hat \lambda - \eta \, s_m(v)}}^{\frac{1}{1-\alpha}}},
\end{equation}
which recovers Equation \eqref{eq:closed-form-kl} in the limit $\alpha \to 1$. This update for $\alpha < 1$ reweights the probability $\pi_m(v)$ by the $\pi_m$-dependent tilt $\paren{1 + (1-\alpha) \pi_m(v)^{1 - \alpha} (\hat \lambda - \eta \, s_m(v))}^{-\frac{1}{1-\alpha}}$. Similar to the $\alpha=1$ case, the tilt factor is increasing in the score $s_m(v)$, thereby boosting the probability of tokens with higher scores. However, the tilt factor is independent of $\pi_m(v)$ only in the special case $\alpha = 1$; for $\alpha < 1$, the update modulates the tilt strength in terms of the probability assigned by the input-prompting PMF $\pi_m(v)$. We quantify the consequences of the choice of $\alpha$ in the next section.

\begin{remark}[Solving for $\hat \lambda$]
\label{rem:lambda}
Computing the update requires the multiplier $\hat \lambda$ that ensures $\hat p_m(\placeholder \, ; \alpha)$ sums to one. For $\alpha = 1$ it is the log-normalizer $\hat \lambda = \ln \sum_{v} \pi_m(v) \, e^{\eta \, s_m(v)}$. For $\alpha <1$ it has no closed form, so we resort to numerical approximation. As a function of $\lambda$, the total mass of the right-hand side of Equation \eqref{eq:closed-form} is strictly decreasing and convex, and $\hat \lambda$ lies in a known interval. Thus, a few steps of bisection and Newton's method can efficiently approximate $\hat \lambda$, while adding negligible cost to the LLM decoding step. See Appendix \ref{app:pf-lambda} for details.
\end{remark}

\section{Theoretical Analysis}
\label{sec:interpreting-alpha}

We analyze the Bregman tilting update in two parts. First, we show how $\alpha$ shapes the geometry of the Bregman divergence regularizer, and how this results in a damped exponential tilt update that limits the change in probabilities of tokens that are unlikely under the input-prompting PMF $\pi_m$. Second, we connect the update, which ascends the surrogate objective $\tilde A_m$, to the true per-token objective $A_m$, deriving a sufficient condition for ascending the true objective. Below, we write $\hat p_m(\placeholder \, ; \alpha, \eta)$ for the solution of Equation \eqref{eq:mirror-step} with the mirror map $\psi_\alpha$ and step size $\eta \geq 0$, and $a_m(v) \coloneq a(v \, ; \rlm)$, so that $A_m(p) = \EV{\Tok \sim p}{a_m(\Tok)}$. All other notation follows Section \ref{sec:mirror-ascent}.

\subsection{\texorpdfstring{The Geometry Induced by $\alpha$}{Geometry induced by alpha}}
\label{sec:role-alpha}
We study how the index $\alpha$ of the mirror maps \eqref{eq:tsallis} and \eqref{eq:endpoints} shapes the Bregman tilting update. Importantly, we highlight how smaller values of $\alpha$ smoothly damp the update strength for each token in terms of its probability under the input-prompting PMF $\pi_m$, incorporating the feedback score into the update without drastically increasing the probability of unlikely tokens.

We first analyze the local behavior of the divergence around $\pi_m$. For small perturbations, the divergence acts as the squared weighted Euclidean norm induced by the positive definite Hessian of the mirror map $\psi_\alpha$.
\begin{proposition}
\label{prop:breg-div-second-order}
Let $\alpha \in [0,1]$. For every $p \in \simplex$, $\grad^2 \psi_\alpha(p) \!=\! \diag \!\paren{p(v)^{\alpha - 2}}_{v \in \Vocab}$, and for every perturbation $u \in \Re^{\Vocab}$ with $\sum_v u(v) = 0$,
\begin{equation}
\label{eq:hessian}
D_{\psi_\alpha}(\pi_m + u, \pi_m)
= \frac{1}{2} \sum_{v \in \Vocab} \pi_m(v)^{\alpha} \paren{\frac{u(v)}{\pi_m(v)}}^2 + \mathcal{O}\paren{\norm{u}{}^3} \quad \text{as } u \to 0.
\end{equation}
\end{proposition}
The divergence locally penalizes the perturbation $u(v)$ of each token by its squared relative change $(u(v) / \pi_m(v))^2$, weighted by $\pi_m(v)^\alpha$. At $\alpha = 1$, the weight is $\pi_m(v)$ itself, so changing the probability of an unlikely token by a large factor costs almost nothing, but as $\alpha$ decreases to $0$ the weights approach one, and at $\alpha = 0$ every weight is exactly one. Since an unlikely token can only gain significant probability by a large relative change, a smaller $\alpha$ makes such a change much more costly.

While this result concerns the regularizer, we show that the update inherits this behavior by analyzing how it changes the probability of a single token $v$ asymptotically as $\pi_m(v)$ goes to 0.
\begin{proposition}
\label{prop:log-ratio}
Let $\alpha \in [0, 1]$ and $\eta > 0$, and fix the score $s_m$. For every token $v \in \Vocab$, the log-ratio of the mass assigned to $v$ by the update $\hat p_m(v \, ; \alpha)$ relative to $\pi_m(v)$ satisfies
\begin{equation}
\label{eq:log-ratio}
\ln \frac{\hat p_m(v \, ; \alpha)}{\pi_m(v)}
= \pi_m(v)^{1 - \alpha} \paren{\eta \, s_m(v) - \hat \lambda} + \mathcal{O}\paren{\pi_m(v)^{2(1 - \alpha)}}
\quad \text{as } \pi_m(v) \to 0.
\end{equation}
\end{proposition}
Thus, to leading order, Bregman tilting changes the log-probability of an unlikely token by $\eta \, s_m(v) - \hat \lambda$, the exact amount of the exponential tilt \eqref{eq:closed-form-kl}, but \emph{damped} by the factor $\pi_m(v)^{1 - \alpha}$. At $\alpha = 1$ the damping factor is one, and every token is tilted with the same step size $\eta$ no matter how unlikely it is. For $\alpha < 1$, the damping factor decreases smoothly to zero as $\pi_m(v)$ decreases, and at $\alpha = 0$ it is $\pi_m(v)$ itself. Thus, the step size $\eta$ sets the strength of the tilt, while $\alpha$ controls how quickly this strength decays on the tokens that the input-prompting PMF $\pi_m$ considers unlikely.

While Proposition \ref{prop:log-ratio} describes the limit $\pi_m(v) \to 0$, we lastly give a non-asymptotic bound on how far the update can promote a token, which holds at every step size $\eta$ for every token whose probability under $\pi_m$ is small relative to its score. To state it, we define the sharpened PMF $\pi_m^{[\alpha]} \in \simplex$ and the expected score $\bar s_m^{[\alpha]}$ under it,
\begin{equation}
\label{eq:sharp-mean-score}
\pi_m^{[\alpha]}(v) \coloneq \frac{\pi_m(v)^{2 - \alpha}}{\sum_{u \in \Vocab} \pi_m(u)^{2 - \alpha}}, \qquad
\bar s_m^{[\alpha]} \coloneq \EV{\Tok \sim \pi_m^{[\alpha]}}{s_m(\Tok)}.
\end{equation}
The sharpened PMF $\pi_m^{[\alpha]}$ raises each probability to the power $2 - \alpha \geq 1$ and renormalizes, which is equivalent to applying the temperature $1 / (2 - \alpha) \leq 1$ to $\pi_m$. At $\alpha = 1$ it is $\pi_m$ itself, and for $\alpha < 1$ it moves probability mass from unlikely tokens to likely ones.
\begin{proposition}
\label{prop:cap}
Let $\alpha \in [0, 1]$ and $\eta > 0$. If a token $v \in \Vocab$ satisfies
\begin{equation}
\label{eq:cap-condition}
(1 - \alpha) \, \eta \, \pi_m(v)^{1 - \alpha} \paren{s_m(v) - \bar s_m^{[\alpha]}} < 1,
\end{equation}
then
\begin{equation}
\label{eq:cap}
\frac{\hat p_m(v \, ; \alpha)}{\pi_m(v)} \leq
\begin{cases}
\paren{1 - (1 - \alpha) \, \eta \, \pi_m(v)^{1 - \alpha} \paren{s_m(v) - \bar s_m^{[\alpha]}}}^{-\frac{1}{1 - \alpha}} & \text{if } \alpha \in [0, 1), \\[2mm]
e^{\eta \paren{s_m(v) - \bar s_m^{[1]}}} & \text{if } \alpha = 1.
\end{cases}
\end{equation}
\end{proposition}
This bounds the factor by which the update can increase the probability of a token which has small probability relative to its score as defined by the condition in \eqref{eq:cap-condition}. At $\alpha = 1$ the cap grows exponentially in the score and is independent of $\pi_m(v)$, so an unlikely token with a large score can gain probability by a large factor. In contrast, for $\alpha < 1$ the score enters the bound only through the damped quantity $\eta \,\pi_m(v)^{1 - \alpha} (s_m(v) - \bar s_m^{[\alpha]})$, making the bound close to one when this quantity is small. For example, at $\alpha = 0$ the bound is $1 / \!\paren{1 - \eta \, \pi_m(v) (s_m(v) - \bar s_m^{[0]})}$, so even a large step size increases the probability of an unlikely token by only a small factor unless its score is extreme.

Notably, the condition \eqref{eq:cap-condition} always holds at $\alpha = 1$, where its left-hand side is zero, as well as when $\alpha<1$ for every token $v$ whose score is at most the sharpened mean, $s_m(v) \leq \bar s_m^{[\alpha]}$. In general, it holds for every token whose probability under $\pi_m$ lies below a constant determined by $\eta$ and its centered score $s_m(v) - \bar s_m^{[\alpha]}$, and it fails only for tokens that are both relatively likely and favored by the feedback, which are the tokens the tilt is meant to move.

Together, these results show that $\alpha$ smoothly controls the strength of the update for each token $v$ in terms of the input-prompting probability $\pi_m(v)$. At $\alpha = 1$, every token is tilted by the same rule regardless of its probability, while a smaller $\alpha$ concentrates the tilt on the likely tokens and restricts it on the unlikely ones. This is preferable when the score is reliable on the tokens that the input-prompting LLM considers likely but possibly unreliable on the rest, since a smaller $\alpha$ prevents a large, possibly erroneous score on an unlikely token from greatly increasing its probability. Section \ref{sec:rate} makes this interpretation precise, showing that the same discounting appears in a sufficient condition for guaranteeing that the update ascends the true objective.

\subsection{Ascending the True Objective}
\label{sec:rate}

We connect the Bregman tilting update on the surrogate $\tilde A_m$ to the true objective $A_m$. We first show that Bregman tilting is guaranteed to ascend the surrogate, a standard result for mirror ascent.

\begin{proposition}
\label{prop:ascent}
Let $\alpha \in [0, 1]$ and $\eta > 0$, and write $\hat p_m = \hat p_m(\placeholder \, ; \alpha, \eta)$. Then
\begin{equation}
\label{eq:ascent}
\tilde A_m(\hat p_m) - \tilde A_m(\pi_m) = \frac{1}{\eta} \Bigl( D_{\psi_\alpha}(\hat p_m, \pi_m) + D_{\psi_\alpha}(\pi_m, \hat p_m) \Bigr) \geq 0.
\end{equation}
Further, if the score $s_m$ is non-constant over $\Vocab$, the gain is strictly positive, $\tilde A_m(\hat p_m) - \tilde A_m(\pi_m) > 0$.
\end{proposition}
To relate this gain to $A_m$, we compare the rates at which the two objectives increase as the step size $\eta$ grows from zero. For $\alpha \in [0, 1]$, we now define $\Gamma_m(\alpha)$ and $\tilde \Gamma_m(\alpha)$ as the derivatives with respect to $\eta$ of $A_m$ and $\tilde A_m$ along the update, evaluated at $\eta = 0$,
\begin{equation}
\label{eq:rates}
\Gamma_m(\alpha) \coloneq \frac{d}{d\eta} \, A_m\paren{\hat p_m(\placeholder \, ; \alpha, \eta)} \Big|_{\eta = 0},
\qquad
\tilde \Gamma_m(\alpha) \coloneq \frac{d}{d\eta} \, \tilde A_m\paren{\hat p_m(\placeholder \, ; \alpha, \eta)} \Big|_{\eta = 0}.
\end{equation}
We show that the ratio of the rates $\Gamma_m(\alpha)$ and $\tilde \Gamma_m(\alpha)$ is determined by the covariance and variance of a token drawn from the sharpened input-prompting PMF $\pi_m^{[\alpha]}$ defined in \eqref{eq:sharp-mean-score}.
\begin{theorem}
\label{thm:rate}
Let $\alpha \in [0, 1]$, and let the score $s_m$ be non-constant over the vocabulary $\Vocab$. Then the derivatives in Equation \eqref{eq:rates} exist, $\tilde \Gamma_m(\alpha) > 0$, and
\begin{equation}
\label{eq:rate-ratio}
\frac{\Gamma_m(\alpha)}{\tilde \Gamma_m(\alpha)}
= \frac{\Cov_{\Tok \sim \pi_m^{[\alpha]}}\paren{s_m(\Tok), \, a_m(\Tok)}}{\Var_{\Tok \sim \pi_m^{[\alpha]}}\paren{s_m(\Tok)}}.
\end{equation}
\end{theorem}
The right-hand side of Equation \eqref{eq:rate-ratio} is the coefficient of the least-squares regression of $a_m(\Tok)$ on $s_m(\Tok)$ under $\Tok \sim \pi_m^{[\alpha]}$. Thus, Theorem \ref{thm:rate} shows that, per unit of gain in the surrogate, the gain in the true objective is determined by how well the score vector $s_m$ predicts the true vector of acceptance probabilities $a_m$. Since the denominators on both sides of Equation \eqref{eq:rate-ratio} are strictly positive, this gives the following condition for ascent of the true objective.
\begin{corollary}
\label{cor:sign}
Let $\alpha \in [0, 1]$, and suppose $\Cov_{\Tok \sim \pi_m^{[\alpha]}}\paren{s_m(\Tok), \, a_m(\Tok)} > 0$. Then there exists $\eta_0 > 0$ such that $A_m\paren{\hat p_m(\placeholder \, ; \alpha, \eta)} > A_m(\pi_m)$ for every $\eta \in (0, \eta_0)$.
\end{corollary}
The Bregman tilting update ascends the true per-token objective $A_m$ at small step sizes whenever the feedback score $s_m(\Tok)$ and the acceptance probability $a_m(\Tok)$ positively covary under $\Tok \sim \pi_m^{[\alpha]}$. Since $\pi_m^{[\alpha]}(v) \propto \pi_m(v)^{1 - \alpha} \, \pi_m(v)$, the weight of each token relative to $\alpha = 1$ is multiplied by $\pi_m(v)^{1 - \alpha}$, the same damping factor as in Equation \eqref{eq:log-ratio}. Thus, in this condition as well, a smaller $\alpha$ discounts the score's errors on the tokens that input prompting considers unlikely when determining whether ascent of the surrogate carries over to the true objective.

Overall, $\alpha$ controls how far the score can move the tokens that input prompting considers unlikely (Propositions \ref{prop:log-ratio} and \ref{prop:cap}), and Corollary \ref{cor:sign} gives a sufficient condition, at small step sizes, for the update to improve the true objective. Since this condition depends on the unknown acceptance probabilities $a_m$, the theory does not determine the best $\alpha$. In Section \ref{sec:experiments}, we compare $\alpha = 1$ and $\alpha = 0$ empirically and find that $\alpha = 0$ performs best, highlighting both the value of the framework and the robustness to unlikely tokens that our analysis identifies.

\section{Experiments}
\label{sec:experiments}

\begin{algorithm}[t]
\caption{Bregman Tilting}
\label{alg:bregman-tilting}
\begin{algorithmic}[1]
\Require LLM $\pi$, feedback mechanism $\Eval$, input-prompting method, number of chains $Q$, number of rounds $T$, index $\alpha \in [0, 1]$, step size $\eta \geq 0$
\For{$t = 1, \ldots, T$}
  \State Build $h^{(t,j)}$ and $\thisttj$ for every $j \in [Q]$ with the input-prompting method
  \For{$j \in [Q]$}
    \For{$m = 1, 2, \ldots$ until the end of the sequence}
      \State $\pi_m \gets \pi(\placeholder \mid h^{(t,j)}, \rlm)$, \quad $s_m \gets \ln \pi_m - \ln \pi(\placeholder \mid \thisttj, \rlm)$
      \State $\hat p_m(\placeholder \, ; \alpha) \gets$ Bregman tilting update of $\pi_m$ with score $s_m$ \Comment{\eqref{eq:closed-form-kl}, \eqref{eq:closed-form}, Remark \ref{rem:lambda}}
      \State Draw $r_m \sim \hat p_m(\placeholder \, ; \alpha)$
    \EndFor
    \State $r^{(t,j)} \gets (r_1, r_2, \ldots)$, \quad $f^{(t,j)} \gets \Eval(r^{(t,j)})$
  \EndFor
\EndFor
\State \Return $r^{(T,j)}$, $j \in [Q]$
\end{algorithmic}
\end{algorithm}

We apply Bregman tilting with $\alpha = 1$ and $\alpha = 0$ on top of two input-prompting methods using two code and two math benchmarks. Appendix \ref{app:exp-details} gives the full experimental details.

\paragraph{Algorithm.} We present the formal Bregman tilting algorithm in Algorithm \ref{alg:bregman-tilting}. In each round $t$, the input-prompting method builds a prompt $h^{(t,j)}$ for each chain $j$ from the previous rollouts and their feedback (Section \ref{sec:input-prompting}). We pair it with its no-feedback counterpart $\thisttj$, which just removes the feedback and the instructions that refer to it. The rollout is generated one token at a time: at each position $m$, we evaluate $\pi$ on both prompts after the same prefix $\rlm$, form the score $s_m$ of Equation \eqref{eq:score}, solve for $\hat \lambda$, and sample the next token from $\hat p_m(\placeholder \, ; \alpha)$. In round 1 both prompts are the zero-shot prompt $h^{(1)}$, so the score is zero and the chain samples from the zero-shot PMF \eqref{eq:zero-shot}. See Appendix \ref{app:exp-prompts} for the exact prompts.

\paragraph{Input-Prompting Methods.} We use \emph{Self-Refine} \citep{madaan:2023:self-refine} and \emph{Recursive Self-Aggregation} (RSA) \citep{venkatraman:2025:rsa}, modified to incorporate feedback. Self-Refine has the model first critique its own previous output, then generate a refined response using this critique. We adapt this to our setting, where the prompt $\histtt$ shows the question, the chain's most recent rollout $\rollt$, and its feedback $\Eval(\rollt)$, and asks for a refined solution that addresses the feedback. For Self-Refine, the $Q$ chains of a question run independently. In contrast, RSA lets the $Q$ chains of a question interact. After each round, the RSA prompt of every chain shows $K = 4$ rollout-feedback pairs drawn at random from the previous round's population and asks for a single improved solution. This allows a correct step found by one chain to spread to the others. For both methods, the no-feedback prompt $\thisttt$ is identical to $\histtt$, except the feedback and corresponding instructions are removed (Appendix \ref{app:exp-prompts}).

\begin{table}[t]
\centering
\footnotesize
\setlength{\tabcolsep}{2.5pt}
\definecolor{tiltshade}{RGB}{251,231,214}
\setlength{\aboverulesep}{0pt}
\setlength{\belowrulesep}{0pt}
\renewcommand{\arraystretch}{1.35}
\caption{Mean pass@1 and pass@16 (\%) at round 5, averaged over four seeds, with $Q = 16$ rollouts per question, along with the standard error. Zero-shot is the round-1 solver which all methods share within a seed. The shaded columns correspond to our {\setlength{\fboxsep}{1.5pt}\colorbox{tiltshade}{Bregman tilting}}. The best score in each row and metric is highlighted in \textcolor{green!50!black}{\textbf{green}}.}
\label{tab:results}
\begin{tabular}{llcc>{\columncolor{tiltshade}}c>{\columncolor{tiltshade}}ccc>{\columncolor{tiltshade}}c>{\columncolor{tiltshade}}c}
\toprule
 & & \multicolumn{4}{c}{Pass@1} & \multicolumn{4}{c}{Pass@16} \\
\cmidrule(lr){3-6} \cmidrule(lr){7-10}
\addlinespace[2pt]
 & & & & \multicolumn{2}{c}{{\setlength{\fboxsep}{2.5pt}\colorbox{tiltshade}{Bregman tilting}}} & & & \multicolumn{2}{c}{{\setlength{\fboxsep}{2.5pt}\colorbox{tiltshade}{Bregman tilting}}} \\
\cmidrule(lr){5-6} \cmidrule(lr){9-10}
\addlinespace[2pt]
\rowcolor{white}
Benchmark & Method & Zero-shot & Prompting & $\alpha = 1$ & $\alpha = 0$ & Zero-shot & Prompting & $\alpha = 1$ & $\alpha = 0$ \\
\midrule
\multirow{2}{*}{LiveCodeBench} & Self-Refine & 34.1\,{\scriptsize$\pm$0.2} & 38.6\,{\scriptsize$\pm$0.2} & 39.9\,{\scriptsize$\pm$0.2} & \textcolor{green!50!black}{\textbf{41.0}}\,{\scriptsize$\pm$0.3} & 47.5\,{\scriptsize$\pm$0.7} & 52.1\,{\scriptsize$\pm$0.7} & 55.5\,{\scriptsize$\pm$0.8} & \textcolor{green!50!black}{\textbf{59.0}}\,{\scriptsize$\pm$0.7} \\
 & RSA & 34.1\,{\scriptsize$\pm$0.2} & 43.6\,{\scriptsize$\pm$0.5} & 45.9\,{\scriptsize$\pm$0.6} & \textcolor{green!50!black}{\textbf{47.0}}\,{\scriptsize$\pm$0.6} & 47.5\,{\scriptsize$\pm$0.7} & 47.9\,{\scriptsize$\pm$0.8} & 49.6\,{\scriptsize$\pm$0.8} & \textcolor{green!50!black}{\textbf{51.5}}\,{\scriptsize$\pm$0.8} \\
\midrule
\multirow{2}{*}{\begin{tabular}[c]{@{}l@{}}MBPP+ (hard)\end{tabular}} & Self-Refine & 40.8\,{\scriptsize$\pm$0.3} & 64.3\,{\scriptsize$\pm$0.3} & 67.1\,{\scriptsize$\pm$0.3} & \textcolor{green!50!black}{\textbf{67.8}}\,{\scriptsize$\pm$0.3} & 62.3\,{\scriptsize$\pm$0.7} & 83.4\,{\scriptsize$\pm$0.5} & 83.5\,{\scriptsize$\pm$0.5} & \textcolor{green!50!black}{\textbf{85.0}}\,{\scriptsize$\pm$0.5} \\
 & RSA & 40.8\,{\scriptsize$\pm$0.3} & 72.9\,{\scriptsize$\pm$0.6} & 77.5\,{\scriptsize$\pm$0.5} & \textcolor{green!50!black}{\textbf{78.9}}\,{\scriptsize$\pm$0.4} & 62.3\,{\scriptsize$\pm$0.7} & 77.7\,{\scriptsize$\pm$0.5} & 81.1\,{\scriptsize$\pm$0.6} & \textcolor{green!50!black}{\textbf{82.6}}\,{\scriptsize$\pm$0.4} \\
\midrule
\multirow{2}{*}{AIME 2025} & Self-Refine & 41.4\,{\scriptsize$\pm$0.6} & 70.3\,{\scriptsize$\pm$0.6} & 71.8\,{\scriptsize$\pm$0.6} & \textcolor{green!50!black}{\textbf{73.2}}\,{\scriptsize$\pm$0.6} & 67.5\,{\scriptsize$\pm$2.0} & 87.5\,{\scriptsize$\pm$1.6} & 90.0\,{\scriptsize$\pm$1.4} & \textcolor{green!50!black}{\textbf{92.5}}\,{\scriptsize$\pm$0.8} \\
 & RSA & 41.4\,{\scriptsize$\pm$0.6} & 80.7\,{\scriptsize$\pm$0.9} & 84.5\,{\scriptsize$\pm$1.0} & \textcolor{green!50!black}{\textbf{85.9}}\,{\scriptsize$\pm$1.2} & 67.5\,{\scriptsize$\pm$2.0} & 85.8\,{\scriptsize$\pm$1.6} & 89.2\,{\scriptsize$\pm$1.4} & \textcolor{green!50!black}{\textbf{90.0}}\,{\scriptsize$\pm$1.5} \\
\midrule
\multirow{2}{*}{\begin{tabular}[c]{@{}l@{}}HMMT Feb 2025\end{tabular}} & Self-Refine & 29.8\,{\scriptsize$\pm$0.5} & 55.3\,{\scriptsize$\pm$0.7} & 57.8\,{\scriptsize$\pm$0.7} & \textcolor{green!50!black}{\textbf{58.2}}\,{\scriptsize$\pm$0.7} & 52.5\,{\scriptsize$\pm$2.3} & 78.3\,{\scriptsize$\pm$1.5} & \textcolor{green!50!black}{\textbf{83.3}}\,{\scriptsize$\pm$1.5} & \textcolor{green!50!black}{\textbf{83.3}}\,{\scriptsize$\pm$1.5} \\
 & RSA & 29.8\,{\scriptsize$\pm$0.5} & 68.7\,{\scriptsize$\pm$1.6} & 74.5\,{\scriptsize$\pm$1.4} & \textcolor{green!50!black}{\textbf{74.6}}\,{\scriptsize$\pm$1.0} & 52.5\,{\scriptsize$\pm$2.3} & 75.0\,{\scriptsize$\pm$1.9} & 80.8\,{\scriptsize$\pm$1.6} & \textcolor{green!50!black}{\textbf{83.3}}\,{\scriptsize$\pm$1.5} \\
\bottomrule
\end{tabular}
\end{table}

\paragraph{Model and Sampling.} We use Qwen3-4B-Instruct-2507 as the solver, sampled at temperature 1. Every method runs for $T = 5$ rounds with $Q = 16$ chains per question, across four seeds. Within a seed, all methods start from the same zero-shot round-1 rollouts and feedback, so the comparisons are paired by question. Bregman tilting uses $\eta = 1$ at $\alpha = 1$ and $\eta = 48$ at $\alpha = 0$.

\begin{figure}[t]
\centering
\includegraphics[width=\linewidth]{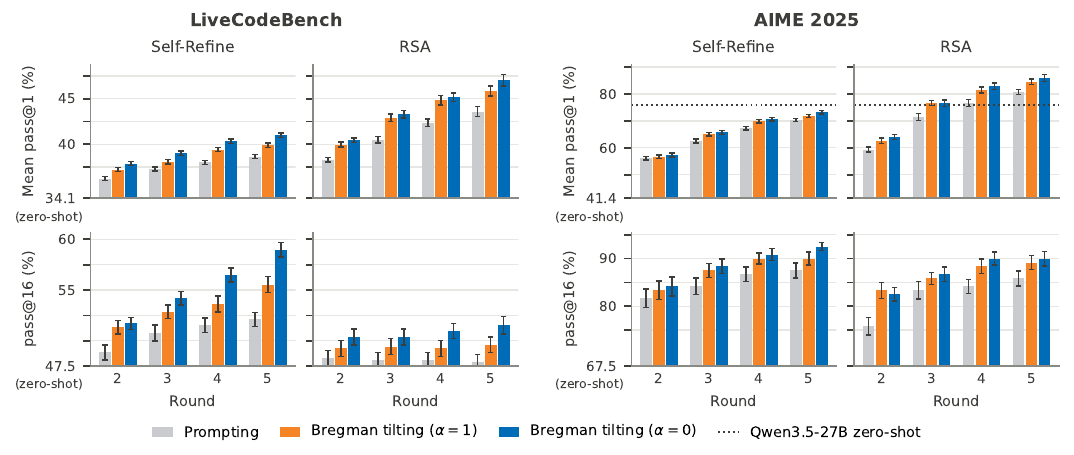}
\caption{Mean pass@1 (top) and pass@16 (bottom) by round on LiveCodeBench (left) and AIME 2025 (right) for Self-Refine and RSA. Prompting refers to input prompting only, and we compare to our Bregman tilting at $\alpha = 0$ and $\alpha = 1$. On LiveCodeBench the feedback comes from running each program on a test suite, and on AIME 2025 from Qwen3.5-27B as a reviewer. Bars are means over four seeds, with standard error bars. Within each seed, all methods use the same round-1 rollouts, drawn from the zero-shot Qwen3-4B solver, so each row's y-axis starts at the round-1 score, marked ``(zero-shot)'', and the height of a bar above it is the method's gain over zero-shot. For AIME, the dotted line is the zero-shot mean pass@1 of the Qwen3.5-27B feedback model.}
\label{fig:lcb-aime-results}
\end{figure}

\paragraph{Benchmarks.} For code, we use LiveCodeBench-v6 \citep{jain:2025:livecodebench} and MBPP+ \citep{liu:2023:evalplus}. We consider a generated program correct if it passes all exposed and hidden tests. For LiveCodeBench we use the problems published from February to April 2025, and for MBPP+ we filter out the 223 of the 378 which Qwen3-4B-Instruct-2507 already solves in all 16 of its seed-0 zero-shot rollouts. We report results on the remaining \emph{hard set} in the main body (see Appendix \ref{app:mbpp-full} for full-set performance). For math, we use AIME 2025 and HMMT February 2025 via MathArena \citep{balunovic:2025:matharena}.

\paragraph{Feedback.} On code, the feedback mechanism $\Eval(r)$ runs the program of $r$ on a set of exposed tests and reports the first test it fails, with the input, the program's output, and the expected output, or the error it raises. The benchmarks' public tests are few and miss many errors, so we expose the public tests together with a fixed random half of the benchmarks' remaining private tests for each problem as in prior work \citep{hubotter:2026:rl-self-dist}. On LiveCodeBench, this raises the share of wrong programs that the feedback flags from about 60\% to over 90\%. On math, we simulate the feedback of a human expert by prompting Qwen3.5-27B \citep{qwen:2026:35-27b} to review the rollout $r$ without seeing the reference answer and to point to the first step that fails and why, without giving an answer of its own.

\paragraph{Metrics.} We report mean pass@1, the fraction of a round's $Q$ rollouts that are correct, and pass@16, whether any of them is correct, each averaged over the questions. Under Self-Refine the $Q$ chains of a question are independent, so the rollouts of a round are independent draws from that round's distribution and pass@16 is its usual pass@$k$ with $k = Q$. Under RSA the chains share rollouts through their prompts, so pass@16 instead measures how many questions the population that RSA produces covers. In both cases, we compute pass@16 from the rollouts directly. Main results are shown with standard errors. See Appendix \ref{app:metrics} for details.

\begin{table}[t]
\centering
\footnotesize
\setlength{\tabcolsep}{5pt}
\renewcommand{\arraystretch}{1.2}
\caption{Necessity of feedback isolation in the reference prompt, on AIME 2025 with Self-Refine. We compare Context-Aware Decoding (CAD) \citep{shi:2024:context-aware-decoding}, which is Bregman tilting at $\alpha = 1$ with the zero-shot reference prompt $h^{(1)}$, to Bregman tilting at $\alpha = 1$ and $\alpha = 0$ with our no-feedback reference prompt $\thisttt$. We report mean pass@1 and pass@16 (\%) at round 5, averaged over four seeds, as well as the percentage of rollouts with a reviewer-flagged wrong final answer whose next-round rollout repeats the same answer, pooled over rounds 1 to 4 and all seeds.}
\label{tab:reference}
\begin{tabular}{llccc}
\toprule
Method & Reference prompt & Mean pass@1 ($\uparrow$) & Pass@16 ($\uparrow$) & {\renewcommand{\arraystretch}{0.9}\begin{tabular}[b]{@{}c@{}}Flagged Wrong Answer\\Repetition Rate (\%) ($\downarrow$)\end{tabular}} \\
\midrule
Prompting & --- & 70.3 & 87.5 & 44.3 \\
CAD & Zero-shot $h^{(1)}$ & 69.8 & 86.7 & 43.4 \\
Bregman tilting, $\alpha = 1$ & No feedback $\thisttt$ (ours) & 71.8 & 90.0 & 29.9 \\
Bregman tilting, $\alpha = 0$ & No feedback $\thisttt$ (ours) & \textbf{73.2} & \textbf{92.5} & \textbf{21.7} \\
\bottomrule
\end{tabular}
\end{table}

\subsection{Results}
\label{sec:results}

\paragraph{Main Results.}

Table \ref{tab:results} reports mean pass@1 and pass@16 after five rounds. Input prompting alone improves substantially over the zero-shot solver on every benchmark, while Bregman tilting further improves on top of input prompting for both values of $\alpha$. Crucially, $\alpha = 0$ achieves the best mean pass@1 and pass@16 in every setting, strictly improving over $\alpha = 1$ everywhere except HMMT Self-Refine pass@16, where they tie. Figure \ref{fig:lcb-aime-results} highlights the round-by-round performance on LiveCodeBench and AIME 2025. Throughout the rounds, both settings of Bregman tilting lead prompting from the first tilted round onward in both metrics, and $\alpha = 0$ leads $\alpha = 1$ in all but two of the 32 round-and-metric comparisons, both in intermediate rounds with RSA on AIME 2025. Since $\alpha = 1$ reduces to the exponential tilt used by prior guidance methods, applied here with our feedback score, these results demonstrate the value of the framework as a whole; the feedback score alone gives consistent gains over input prompting, while moving beyond the exponential tilt to $\alpha = 0$ improves even further. See Appendix \ref{app:by-round} for the round-by-round results on all benchmarks.

\paragraph{Necessity of Feedback Isolation.} Our feedback score \eqref{eq:score} contrasts the LLM under the prompt $\histtt$, containing the previous rollout and its feedback, against the reference prompt $\thisttt$, which contains the same rollout but no feedback. Prior contrastive scores \citep{shi:2024:context-aware-decoding,hubotter:2026:rl-self-dist} instead contrast against a reference prompt without any added context, which in our pipeline is the zero-shot prompt $h^{(1)}$. In this case, Bregman tilting at $\alpha = 1$ reduces to Context-Aware Decoding (CAD) \citep{shi:2024:context-aware-decoding} with score $\ln \pi(v \mid \histtt, \rlm) - \ln \pi(v \mid h^{(1)}, \rlm)$. Since $h^{(1)}$ omits both the previous rollout and its feedback, this score muddles the effect of showing the previous rollout with the effect of the feedback, rewarding tokens that agree with the previous rollout regardless of correctness.

To demonstrate this effect, we run Bregman tilting at $\alpha = 1$ with the zero-shot reference on AIME 2025 with Self-Refine, keeping each seed's round-1 rollouts and feedback unchanged. Along with mean pass@1 and pass@16, we measure how often the solver repeats a wrong answer when the reviewer model has flagged it as incorrect, as the reviewer is prompted to end each reply with a correctness verdict which is stripped from the returned feedback (Appendix \ref{app:exp-prompts-math}). Table \ref{tab:reference} shows that with the zero-shot reference, Bregman tilting at $\alpha=1$ is no better than input prompting in either metric, and it trails our reference prompt at every round. Interestingly, the zero-shot reference repeats flagged wrong answers as often as input prompting. Our reference prompt reduces this repetition pitfall at $\alpha=1$, with the most reduction at $\alpha = 0$. Thus, the zero-shot reference biases the tilt towards copying the previous rollout, while our reference lets the tilt incorporate the feedback more effectively. See Figure \ref{fig:reference-by-round} for round-by-round results.

\paragraph{Runtime.} We compare the runtime of Bregman tilting, input prompting, and CAD, which is Bregman tilting with $\alpha=1$ and the zero-shot reference prompt $h^{(1)}$. At each step, both CAD and Bregman tilting additionally evaluate the LLM on a reference prompt in the same batch to compute the guidance score. Bregman tilting at $\alpha=0$ also solves for $\hat \lambda$ to construct the updated distribution (Remark \ref{rem:lambda}). We use a set of fixed Self-Refine rollouts on AIME 2025, where at each position, each method computes its next-token PMF and samples a token, but decoding then continues from the recorded token rather than the sample. This makes the comparison fair, since every method decodes from the same prefixes. Table \ref{tab:runtime} reports the time per generated token on a single NVIDIA H100 GPU, for both 1 chain and 8 chains decoded in parallel. We observe that solving for $\hat \lambda$ is nearly free, as moving from $\alpha = 1$ to $\alpha = 0$ adds only 1--2\%. The rest of the overhead comes from the reference prompt. Each decode has a fixed cost shared by all sequences in the batch, so the time per token of input prompting drops from 5.46 ms with 1 chain to 1.58 ms with 8 chains. The reference prompt doubles the batch size, adding about $0.5$ ms per token in both cases, so the relative overhead for the guidance methods grows from about $1.1\times$ to $1.3\times$. See Appendix \ref{app:runtime} for details.

\begin{table}[t]
\centering
\footnotesize
\setlength{\tabcolsep}{5pt}
\renewcommand{\arraystretch}{1.2}
\caption{Runtime comparison with Qwen3-4B-Instruct-2507. Every method decodes the same AIME 2025 round-2 Self-Refine rollouts, with 1 or 8 chains decoded in parallel. We report the decoding time per generated token and its ratio to prompting, along with the time to process the prompts of a rollout before decoding and the peak GPU memory. CAD \citep{shi:2024:context-aware-decoding} is Bregman tilting at $\alpha = 1$ with the zero-shot reference prompt $h^{(1)}$.}
\label{tab:runtime}
\begin{tabular}{clcccc}
\toprule
Chains & Method & ms / token & $\times$ Prompting & Prompt processing (s) & Peak memory (GiB) \\
\midrule
\multirow{4}{*}{1} & Prompting & 5.46 & 1.00 & 0.20 & 12.4 \\
 & CAD ($\alpha = 1$, reference $h^{(1)}$) & 5.96 & 1.09 & 0.22 & 15.2 \\
 & Bregman tilting, $\alpha = 1$ & 5.96 & 1.09 & 0.39 & 15.2 \\
 & Bregman tilting, $\alpha = 0$ & 6.07 & 1.11 & 0.39 & 15.2 \\
\midrule
\multirow{4}{*}{8} & Prompting & 1.58 & 1.00 & 0.21 & 32.1 \\
 & CAD ($\alpha = 1$, reference $h^{(1)}$) & 2.04 & 1.30 & 0.23 & 54.6 \\
 & Bregman tilting, $\alpha = 1$ & 2.09 & 1.32 & 0.41 & 54.6 \\
 & Bregman tilting, $\alpha = 0$ & 2.11 & 1.34 & 0.41 & 54.6 \\
\bottomrule
\end{tabular}
\end{table}
\vspace{-1mm}
\section{Conclusion}
\vspace{-1mm}
We introduced Bregman tilting, an inference-time framework for guiding an LLM with text feedback in iterative problem solving. We defined a per-token feedback score as the change in its log-probability when the feedback was added to a prompt containing the prior attempt, isolating the effect of the feedback, and incorporated the score into the updated next-token distribution via a mirror ascent step regularized by a Bregman divergence based on the Tsallis entropy parameterized by $\alpha \in [0,1]$. Analysis showed that this update resulted in a generalized exponential tilt that damped the tilt strength on unlikely tokens under input prompting, and that at small step sizes it provably increased the true correctness probability whenever the feedback score covaried positively with correctness under a sharpened input-prompting distribution. Empirically, Bregman tilting outperformed input prompting in every setting, and $\alpha = 0$, which damped the most, performed best throughout.
\vspace{-1mm}
\section*{Acknowledgments}
\vspace{-1mm}
This work was supported in part by NSF Grants 2019844, 2107037, 2112471, and 2505865, ONR Grant N00014-19-1-2566, the Machine Learning Lab (MLL) at UT Austin, the NSF AI Institute for Foundations of Machine Learning (IFML), the NSF AI-EDGE Institute, and the Wireless Networking and Communications Group (WNCG) Industrial Affiliates Program. We are grateful for computing support on the Vista GPU Cluster through the Center for Generative AI (CGAI) and the Texas Advanced Computing Center (TACC) at the University of Texas at Austin.

\clearpage
\appendix

\makeatletter
\addtocontents{atoc}{} 

\let\orig@addcontentsline\addcontentsline
\renewcommand{\addcontentsline}[3]{\edef\@tempa{#1}\edef\@tempb{toc}\ifx\@tempa\@tempb
\orig@addcontentsline{atoc}{#2}{#3}\else
\orig@addcontentsline{#1}{#2}{#3}\fi
}
\makeatother

\clearpage
\section*{Contents of Appendix}
\setcounter{tocdepth}{3}
\makeatletter
\@starttoc{atoc}
\makeatother
\clearpage

\section{Theoretical Analysis}
\label{app:proofs}
\theoremstyle{plain}
\newtheorem*{corollary*}{Corollary}

In this section, we prove every formal statement from the main body. The proofs are organized by the sections in which they appear. Appendix \ref{app:prelims} establishes general properties of Bregman divergences and of the mirror-ascent update \eqref{eq:mirror-step}, along with basic facts about the specific family of mirror maps $\psi_\alpha$. We prove the main results in the following sections.

\subsection{Preliminaries}
\label{app:prelims}

\begin{lemma}
\label{lem:bregman-basics}
Let $\psi$ be strictly convex and differentiable on a convex open set $U \subseteq \Re^{\Vocab}$, and let $D_\psi$ be its Bregman divergence \eqref{eq:bregman}. Then the following hold for every $p, q, z \in U$.
\begin{enumerate}[label=(\roman*), topsep=1pt, itemsep=1pt]
\item $D_\psi(p, q) \geq 0$, with equality if and only if $p = q$.
\item The map $p \mapsto D_\psi(p, q)$ is strictly convex on $U$.
\item If $\tilde \psi(p) = \psi(p) + \innerprod{w}{p} + c$ for a vector $w \in \Re^{\Vocab}$ and a scalar $c \in \Re$, then $D_{\tilde \psi} = D_\psi$.
\item $D_\psi(p, q) + D_\psi(q, z) - D_\psi(p, z) = \innerprod{\grad \psi(z) - \grad \psi(q)}{p-q}$.
\end{enumerate}
\end{lemma}
\begin{proof}[Proof of Lemma \ref{lem:bregman-basics}]
(i) Since $\psi$ is strictly convex and differentiable, it lies strictly above each of its tangent lines, except at the point of reference \citep{boyd:2004:cvx}, i.e., for every $p \neq q$, 
\begin{equation*}
    \psi(p) > \psi(q) + \innerprod{\grad \psi(q)}{p - q}.
\end{equation*}
Thus for $p \neq q$,
\begin{equation*}
    D_\psi(p, q) = \psi(p) - \paren{\psi(q) + \innerprod{\grad \psi(q)}{p - q}} > 0.
\end{equation*}
Further by definition,
\begin{equation*}
    D_\psi(q, q) = \psi(q) - \paren{\psi(q) + \innerprod{\grad \psi(q)}{q - q}} = 0.
\end{equation*}

(ii) As a function of $p$, $D_\psi(p, q)$ is the sum of the strictly convex function $\psi(p)$ and the affine function $-\psi(q) - \innerprod{\grad \psi(q)}{p - q}$, so it is strictly convex. 

(iii) Using the assumption, $\grad \tilde \psi(q) = \grad \psi(q) + w$. Thus,
\begin{align*}
D_{\tilde \psi}(p, q)
&= \tilde \psi(p) - \tilde \psi(q) - \innerprod{\grad \tilde \psi(q)}{p-q} \\
&= \psi(p) + \innerprod{w}{p} - \psi(q) - \innerprod{w}{q} - \innerprod{\grad \psi(q) + w}{p-q} \\
&= \psi(p) - \psi(q) + \innerprod{w}{p-q} - \innerprod{\grad \psi(q)}{p-q} - \innerprod{w}{p-q} \\
&= \psi(p) - \psi(q) - \innerprod{\grad \psi(q)}{p-q} \\
&= D_\psi(p, q).
\end{align*}

(iv) By definition of $D_\psi$, each $\psi(p),\psi(q),\psi(z)$ cancels out, giving
\begin{align*}
D_\psi(p, q) + D_\psi(q, z) - D_\psi(p, z) &= -\innerprod{\grad \psi(q)}{p-q} - \innerprod{\grad \psi(z)}{q-z} + \innerprod{\grad \psi(z)}{p-z}\\
&=- \innerprod{\grad \psi(q)}{p-q} + \innerprod{\grad \psi(z)}{p-q} \\
&= \innerprod{\grad \psi(z) - \grad \psi(q)}{p-q}.
\qedhere
\end{align*}
\end{proof}

\begin{lemma}
\label{lem:tsallis-facts}
Let $\alpha \in [0, 1]$. The mirror map $\psi_\alpha$ of Equations \eqref{eq:tsallis} and \eqref{eq:endpoints} is infinitely differentiable on $(0, \infty)^{\Vocab}$, and for every $p \in (0, \infty)^{\Vocab}$ and $v \in \Vocab$,
\begin{equation*}
[\grad \psi_\alpha(p)](v) =
\begin{cases}
-\dfrac{p(v)^{\alpha - 1}}{1 - \alpha} & \text{if } \alpha \in [0, 1), \\[2mm]
\ln p(v) + 1 & \text{if } \alpha = 1,
\end{cases}
\qquad
\grad^2 \psi_\alpha(p) = \diag\paren{p(v)^{\alpha - 2}}_{v \in \Vocab}.
\end{equation*}
In particular, $\psi_\alpha$ is strictly convex on $(0, \infty)^{\Vocab}$, and $\lim_{p(v) \to 0^{+}} [\grad \psi_\alpha(p)](v) = -\infty$.
\end{lemma}
\begin{proof}[Proof of Lemma \ref{lem:tsallis-facts}]
By Equations \eqref{eq:tsallis} and \eqref{eq:endpoints}, $\psi_\alpha(p) = \sum_{v \in \Vocab} \phi_\alpha(p(v))$, where
\begin{equation*}
\phi_\alpha(x) \coloneq -\frac{x^\alpha}{\alpha (1 - \alpha)} \ \ \text{for } \alpha \in (0, 1),
\qquad
\phi_0(x) \coloneq -\ln x,
\qquad
\phi_1(x) \coloneq x \ln x.
\end{equation*}
Each $\phi_\alpha$ is infinitely differentiable on $(0, \infty)$, so $\psi_\alpha$ is infinitely differentiable on $(0, \infty)^{\Vocab}$, with $[\grad \psi_\alpha(p)](v) = \phi_\alpha'(p(v))$ and $\grad^2 \psi_\alpha(p) = \diag\paren{\phi_\alpha''(p(v))}_{v \in \Vocab}$.

For $\alpha \in (0, 1)$,
\begin{equation}
\label{eq:app-mirror-map-derivs}
\phi_\alpha'(x) = -\frac{\alpha \, x^{\alpha - 1}}{\alpha (1 - \alpha)} = -\frac{x^{\alpha - 1}}{1 - \alpha},
\qquad
\phi_\alpha''(x) = -\frac{(\alpha - 1) \, x^{\alpha - 2}}{1 - \alpha} = x^{\alpha - 2}.
\end{equation}
At $\alpha=0$,
\begin{equation*}
\phi_0'(x) = - \frac{1}{x},
\qquad
\phi_0''(x) = \frac{1}{x^2},
\end{equation*}
which is equivalent to \eqref{eq:app-mirror-map-derivs} evaluated at $\alpha=0$.

At $\alpha=1$,
\begin{equation*}
\phi_1'(x) = 1 + \ln(x),
\qquad
\phi_1''(x) = \frac{1}{x}.
\end{equation*}
This proves the form of the gradient and Hessian. Since the Hessian is diagonal with strictly positive entries, it is positive definite, implying strict convexity of $\psi_{\alpha}$ \citep{boyd:2004:cvx}. Lastly,
\begin{equation*}
    \lim_{p(v) \to 0^+} [\grad \psi_\alpha(p)](v) =
\begin{cases}
\lim_{x \to 0^+} -\dfrac{x^{\alpha - 1}}{1 - \alpha} = -\infty & \text{if } \alpha \in [0, 1), \\[3mm]
\lim_{x \to 0^+} 1 + \ln(x) = -\infty & \text{if } \alpha = 1.
\end{cases}
\end{equation*}
\end{proof}

\begin{lemma}
\label{lem:mirror-map-endpoint-const-limit}
For the family of mirror maps $\psi_\alpha$ for $\alpha \in [0,1]$ defined in Equations \eqref{eq:tsallis} and \eqref{eq:endpoints}, $\psi_0$ and $\psi_1$ are the limits of $\psi_\alpha$ up to additive constants. Precisely, for every $p \in \simplex$,
\begin{equation*}
\lim_{\alpha \to 0} \paren{\psi_\alpha(p) + \abs{\Vocab} \paren{\frac{1}{\alpha} + 1}} = \psi_0(p),
\qquad
\lim_{\alpha \to 1} \paren{\psi_\alpha(p) + \frac{1}{\alpha (1 - \alpha)}} = \psi_1(p).
\end{equation*}
\end{lemma}

\begin{proof}[Proof of Lemma \ref{lem:mirror-map-endpoint-const-limit}]
For the first limit, for every $v \in \Vocab$, we have that
\begin{equation*}
p(v)^\alpha = e^{\alpha \ln p(v)} = 1 + \alpha \ln p(v) + \mathcal{O}(\alpha^2).
\end{equation*}
Then, using the identity $\frac{1}{\alpha (1 - \alpha)} = \frac{1}{\alpha} + \frac{1}{1 - \alpha}$ and substituting these into Equation \eqref{eq:tsallis},
\begin{align*}
\lim_{\alpha \to 0} \paren{ \psi_\alpha(p) + \abs{\Vocab} \paren{\frac{1}{\alpha} + 1}}
&= \lim_{\alpha \to 0} \paren{ \abs{\Vocab} \paren{\frac{1}{\alpha} + 1} -\paren{\frac{1}{\alpha} + \frac{1}{1 - \alpha}} \sum_{v \in \Vocab} \paren{1 + \alpha \ln p(v) + \mathcal{O}(\alpha^2)} }\\
&= \lim_{\alpha \to 0} \paren{\abs{\Vocab} \paren{\frac{1}{\alpha} + 1} - \abs{\Vocab} \paren{\frac{1}{\alpha} + \frac{1}{1-\alpha}} - \frac{1}{1-\alpha} \sum_{v \in \Vocab} \ln p(v) + \mathcal{O}(\alpha)} \\
&= \lim_{\alpha \to 0} \paren{\abs{\Vocab} \paren{1- \frac{1}{1-\alpha}} - \sum_{v \in \Vocab} \ln p(v) + \mathcal{O}(\alpha) }\\
&= -\sum_{v \in \Vocab} \ln p(v).
\end{align*}

For the second limit, let $\epsilon \coloneq 1 - \alpha$. For every $v \in \Vocab$,
\begin{equation*}
p(v)^{1 - \epsilon} = p(v) \, e^{-\epsilon \ln p(v)} = p(v) - \epsilon \, p(v) \ln p(v) + \mathcal{O}(\epsilon^2).
\end{equation*}

Substituting this into Equation \eqref{eq:tsallis}, and using $\sum_v p(v) = 1$,
\begin{align*}
\lim_{\alpha \to 1} \paren{\psi_{\alpha}(p) + \frac{1}{\alpha(1-\alpha)}}
&= \lim_{\alpha \to 1} \paren{\frac{1}{\alpha(1-\alpha)} - \frac{1}{\alpha(1-\alpha)} \sum_{v \in \Vocab} p(v)^{\alpha} }\\
&= \lim_{\epsilon \to 0} \paren{\frac{1}{(1 - \epsilon) \epsilon} \paren{1 - \sum_{v \in \Vocab} p(v)^{1 - \epsilon}} }\\
&= \lim_{\epsilon \to 0} \paren{\frac{1}{(1 - \epsilon) \epsilon} \left[1 - \paren{\sum_{v \in \Vocab} p(v)} + \epsilon \paren{\sum_{v \in \Vocab} p(v) \ln p(v)} + \mathcal{O}(\epsilon^2) \right] }\\
&= \lim_{\epsilon \to 0} \paren{ \frac{1}{1 - \epsilon} \sum_{v \in \Vocab} p(v) \ln p(v) + \mathcal{O}(\epsilon) }\\
&= \sum_{v \in \Vocab} p(v) \ln p(v).
\end{align*}
\end{proof}

\subsection{Proofs of Main Results}
\label{app:main-proofs}

\subsubsection{Section \ref{sec:per-token-surrogate}}

\subsubsection*{Proof of Proposition \ref{prop:pdl}}
\label{app:pf-pdl}

\begin{proposition*}
For every autoregressive PMF $\rho$,
\begin{equation*}
A(\rho) - A(\pitt) = \sum_{m=1}^{M} \EV{\Rollout \sim \rho}{ \bar a\paren{\rho \, ; \Rollout_{<m}} - \bar a(\pitt \, ; \Rollout_{<m})}.
\end{equation*}
\end{proposition*}
\begin{proof}[Proof of Proposition \ref{prop:pdl}]
Fix a prefix $x$ of length less than $M$. Conditioning on the first token $v$ that $\pitt$ generates after $x$,
\begin{equation*}
A(\pitt \, ; x) = \sum_{v \in \Vocab} \pitt(v \mid x) \, A\paren{\pitt \, ; (x, v)}.
\end{equation*}
By the definition \eqref{eq:token-value}, $A\paren{\pitt \, ; (x, v)} = a(v \, ; x)$, so by the definition \eqref{eq:objective-token},
\begin{equation}
\label{eq:pdl-condition}
A(\pitt \, ; x) = \sum_{v \in \Vocab} \pitt(v \mid x) \, a(v \, ; x) = \bar a(\pitt \, ; x).
\end{equation}
We now show the result.
\begin{align}
A(\rho) - A(\pitt)
&= \EV{\Rollout \sim \rho}{\indic{\Rollout \in \calA}} - A(\pitt) \nonumber \\
\label{eq:app-per-tok-obj-chain-0}
&= \EV{\Rollout \sim \rho}{\indic{\Rollout \in \calA} - A(\pitt)},
\end{align}
since $A(\pitt)$ is a constant.

Since $\Rollout = \Rollout_{<M+1}$ is a complete rollout, we can write $\indic{\Rollout \in \calA} = A(\pitt \, ; \Rollout_{<M+1})$. Similarly, since $\Rollout_{<1}$ is empty, we have that $A(\pitt) = A(\pitt \, ; \Rollout_{<1})$. Thus,
\begin{equation}
\label{eq:app-per-tok-obj-chain-1}
\EV{\Rollout \sim \rho}{\indic{\Rollout \in \calA} - A(\pitt)} = \EV{\Rollout \sim \rho}{A(\pitt \, ; \Rollout_{<M+1}) - A(\pitt \, ; \Rollout_{<1})}.
\end{equation}
We then telescope over the prefixes of $\Rollout$, giving that
\begin{align}
& \EV{\Rollout \sim \rho}{A(\pitt \, ; \Rollout_{<M+1}) - A(\pitt \, ; \Rollout_{<1})} \nonumber \\
& \qquad \qquad = \EV{\Rollout \sim \rho}{\sum_{m=1}^{M} A(\pitt \, ; \Rollout_{<m+1}) - A(\pitt \, ; \Rollout_{<m}) } \nonumber \\
& \qquad  \qquad = \EV{\Rollout \sim \rho}{\sum_{m=1}^{M} A\paren{\pitt \, ; (\Rollout_{<m}, \Rollout_m)} - A(\pitt \, ; \Rollout_{<m})} \nonumber \\
& \qquad \qquad = \sum_{m=1}^{M} \EV{\Rollout \sim \rho}{A\paren{\pitt \, ; (\Rollout_{<m}, \Rollout_m)} - A(\pitt \, ; \Rollout_{<m})} \nonumber \\
& \qquad \qquad = \sum_{m=1}^{M} \EV{\Rollout \sim \rho}{a\paren{\Rollout_{m} \, ; \Rollout_{<m}} - A(\pitt \, ; \Rollout_{<m})} \nonumber \\
\label{eq:app-per-tok-obj-chain-2}
& \qquad \qquad = \sum_{m=1}^{M} \EV{\Rollout \sim \rho}{
\EV{\Rollout \mid \Rollout_{<m}}{
a\paren{\Rollout_{m} \, ; \Rollout_{<m}} - A(\pitt \, ; \Rollout_{<m})
}},
\end{align}
where the last line uses the law of iterated expectation, conditioning on the prefix $\Rollout_{<m}$ of $\Rollout \sim \rho$. Since $A(\pitt \, ; \Rollout_{<m})$ depends only on $\Rollout_{<m}$, which has length $m - 1 < M$, Equation \eqref{eq:pdl-condition} gives that
\[
\EV{\Rollout \mid \Rollout_{<m}}{A(\pitt \, ; \Rollout_{<m})} = A(\pitt \, ; \Rollout_{<m}) = \bar a(\pitt \, ; \Rollout_{<m}).
\]
Further, since $a\paren{\Rollout_{m} \, ; \Rollout_{<m}}$ only depends on $\Rollout_{<m}$ and $\Rollout_{m}$, and since $\Rollout_m \mid \Rollout_{<m}$ is distributed according to $\rho(\placeholder \mid \Rollout_{<m})$ for $\Rollout \sim \rho$ by definition,
\[
\EV{\Rollout \mid \Rollout_{<m}}{
a\paren{\Rollout_{m} \, ; \Rollout_{<m}}} = \EV{\Rollout_m \sim \rho(\placeholder \mid \Rollout_{<m})}{
a\paren{\Rollout_{m} \, ; \Rollout_{<m}}} = \bar a(\rho \, ; \Rollout_{<m}).
\]
Applying these two simplifications to the chain of equalities above,
\begin{equation}
\label{eq:app-per-tok-obj-chain-3}
\sum_{m=1}^{M} \EV{\Rollout \sim \rho}{
\EV{\Rollout \mid \Rollout_{<m}}{
a\paren{\Rollout_{m} \, ; \Rollout_{<m}} - A(\pitt \, ; \Rollout_{<m})}}
= \sum_{m=1}^{M} \EV{\Rollout \sim \rho}{\bar a(\rho \, ; \Rollout_{<m}) - \bar a(\pitt \, ; \Rollout_{<m})}.
\end{equation}
Then chaining the string of equalities in \eqref{eq:app-per-tok-obj-chain-0}, \eqref{eq:app-per-tok-obj-chain-1}, \eqref{eq:app-per-tok-obj-chain-2}, \eqref{eq:app-per-tok-obj-chain-3} gives the result.

\end{proof}

\subsubsection{Section \ref{sec:mirror-ascent}}

\subsubsection*{Proof of Proposition \ref{prop:tsallis-bregman-limits}}
\label{app:pf-tsallis}

\begin{proposition*}
Consider the family of mirror maps $\psi_{\alpha}$ defined by Equations \eqref{eq:tsallis}, \eqref{eq:endpoints}, and arbitrary PMFs $p,q \in \simplex$.
\begin{enumerate}[label=(\roman*), topsep=1pt, itemsep=1pt]
\item On the interior $\alpha \in (0,1)$, the Bregman divergence of $\psi_\alpha$ is
\begin{equation*}
D_{\psi_\alpha}(p, q) = \frac{1}{\alpha (1 - \alpha)} \sum_{v \in \Vocab} \paren{q(v)^\alpha - p(v)^\alpha + \alpha \, q(v)^{\alpha - 1} \paren{p(v) - q(v)}}.
\end{equation*}
\item The Bregman divergence $D_{\psi_\alpha}$ is a continuous function of $\alpha$ at the endpoints $\{0,1\}$, with limits
\begin{align*}
\lim_{\alpha \to 0} D_{\psi_\alpha}(p, q) &= D_{\psi_0}(p, q) = \sum_{v \in \Vocab} \paren{\frac{p(v)}{q(v)} - \ln \frac{p(v)}{q(v)} - 1}, \\
\lim_{\alpha \to 1} D_{\psi_\alpha}(p, q) &= D_{\psi_1}(p, q) = \sum_{v \in \Vocab} p(v) \ln \frac{p(v)}{q(v)} = \KL{p}{q}.
\end{align*}
\end{enumerate}
\end{proposition*}
\begin{proof}[Proof of Proposition \ref{prop:tsallis-bregman-limits}]
Since $p, q \in \simplex$ have full support, $\psi_\alpha$ is differentiable at both for every $\alpha \in [0, 1]$ (Lemma \ref{lem:tsallis-facts}).

(i) Let $\alpha \in (0, 1)$. By Lemma \ref{lem:tsallis-facts},
\begin{equation*}
[\grad \psi_\alpha(q)](v) = -\frac{q(v)^{\alpha - 1}}{1 - \alpha}.
\end{equation*}
Substituting this and Equation \eqref{eq:tsallis} into the definition \eqref{eq:bregman},
\begin{align*}
D_{\psi_\alpha}(p, q)
&= \psi_\alpha(p) - \psi_\alpha(q) - \innerprod{\grad \psi_\alpha(q)}{p - q} \\
&= -\frac{1}{\alpha (1 - \alpha)} \sum_{v \in \Vocab} p(v)^\alpha + \frac{1}{\alpha (1 - \alpha)} \sum_{v \in \Vocab} q(v)^\alpha + \frac{1}{1 - \alpha} \sum_{v \in \Vocab} q(v)^{\alpha - 1} \paren{p(v) - q(v)} \\
&= \frac{1}{\alpha (1 - \alpha)} \sum_{v \in \Vocab} \paren{q(v)^\alpha - p(v)^\alpha + \alpha \, q(v)^{\alpha - 1} \paren{p(v) - q(v)}}.
\end{align*}

(ii) We first compute the Bregman divergences of the endpoints. By Lemma \ref{lem:tsallis-facts}, $[\grad \psi_0(q)](v) = -1/q(v)$. Substituting this and $\psi_0$ from Equation \eqref{eq:endpoints} into the definition \eqref{eq:bregman},
\begin{align*}
D_{\psi_0}(p, q)
&= -\sum_{v \in \Vocab} \ln p(v) + \sum_{v \in \Vocab} \ln q(v) + \sum_{v \in \Vocab} \frac{p(v) - q(v)}{q(v)} \\
&= \sum_{v \in \Vocab} \paren{\frac{p(v)}{q(v)} - \ln \frac{p(v)}{q(v)} - 1}.
\end{align*}
By Lemma \ref{lem:tsallis-facts}, $[\grad \psi_1(q)](v) = \ln q(v) + 1$. Substituting this and $\psi_1$ from Equation \eqref{eq:endpoints} into the definition \eqref{eq:bregman},
\begin{align*}
D_{\psi_1}(p, q)
&= \sum_{v \in \Vocab} p(v) \ln p(v) - \sum_{v \in \Vocab} q(v) \ln q(v) - \sum_{v \in \Vocab} \paren{\ln q(v) + 1} \paren{p(v) - q(v)} \\
&= \sum_{v \in \Vocab} p(v) \ln p(v) - \sum_{v \in \Vocab} p(v) \ln q(v) - \sum_{v \in \Vocab} \paren{p(v) - q(v)} \\
&= \sum_{v \in \Vocab} p(v) \ln \frac{p(v)}{q(v)},
\end{align*}
where the last step uses $\sum_v p(v) = \sum_v q(v) = 1$.

It remains to compute the limits. Define the two functions of $\alpha$
\[w_0(\alpha) = \abs{\Vocab} \paren{\frac{1}{\alpha} + 1} \quad \text{and}  \quad w_1(\alpha) = \frac{1}{\alpha (1 - \alpha)}.
\]
Fix $k \in \{0,1\}$. By the definition of the Bregman divergence \eqref{eq:bregman},
\begin{equation*}
D_{\psi_\alpha}(p,q) = \paren{\psi_\alpha(p) + w_k(\alpha)} - \paren{\psi_\alpha(q) + w_k(\alpha)} - \innerprod{\grad \psi_\alpha(q)}{p-q},
\end{equation*}
where we add and subtract $w_k(\alpha)$. We aim to show $D_{\psi_{\alpha}}(p,q)$ converges to $D_{\psi_{k}}(p,q)$ as $\alpha \to k$.

By Lemma \ref{lem:mirror-map-endpoint-const-limit}, the first two terms converge to the value at the endpoint,
\begin{align}
\lim_{\alpha \to k} \paren{\psi_\alpha(p) + w_k(\alpha) - \psi_\alpha(q) - w_k(\alpha)}
&= \lim_{\alpha \to k} \paren{\psi_\alpha(p) + w_k(\alpha)} - \lim_{\alpha \to k} \paren{\psi_\alpha(q) + w_k(\alpha)} \nonumber \\
\label{eq:breg-div-cont-first-term}
&= \psi_k(p) - \psi_k(q).
\end{align}
It then suffices to show that $\lim_{\alpha \to k} \innerprod{\grad \psi_\alpha(q)}{p-q} = \innerprod{\grad \psi_k(q)}{p-q}$. We consider both cases.

\textbf{Case 1:} $k=0$.

By Lemma \ref{lem:tsallis-facts},
\begin{equation*}
\lim_{\alpha \to 0} [\grad \psi_\alpha(q)](v) = \lim_{\alpha \to 0} -\frac{q(v)^{\alpha-1}}{1-\alpha} = -\frac{1}{q(v)} = [\grad \psi_0(q)](v).
\end{equation*}
Thus,
\begin{equation*}
\lim_{\alpha \to 0} \innerprod{\grad \psi_\alpha(q)}{p-q} = \innerprod{\grad \psi_0(q)}{p-q}.
\end{equation*}
\textbf{Case 2:} $k=1$.

First, since $\innerprod{\1}{p-q} = 0$, we have that for every $c \in \Re$,
\begin{equation*}
\innerprod{\grad \psi_\alpha(q)}{p-q} = \innerprod{\grad \psi_\alpha(q) + c \1}{p-q}.
\end{equation*}
Take $c = \frac{1}{1-\alpha}$ and let $\epsilon \coloneq 1 - \alpha$. Since $q(v)^{-\epsilon} = 1 - \epsilon \ln q(v) + \mathcal{O}(\epsilon^2)$,
\begin{align*}
\lim_{\alpha \to 1} \paren{[\grad \psi_\alpha(q)](v) + \frac{1}{1-\alpha}}
&= \lim_{\alpha \to 1} \paren{-\frac{q(v)^{\alpha-1}}{1-\alpha} + \frac{1}{1-\alpha}} \\
&= \lim_{\epsilon \to 0} \frac{1 - q(v)^{-\epsilon}}{\epsilon} \\
&= \ln q(v).
\end{align*}
Then since $\grad \psi_1(q) = \ln q + \1$ by Lemma \ref{lem:tsallis-facts},
\begin{align*}
\lim_{\alpha \to 1} \innerprod{\grad \psi_\alpha(q)}{p-q}
&= \lim_{\alpha \to 1} \innerprod{\grad \psi_\alpha(q) + \frac{1}{1-\alpha} \cdot \1 }{p-q} \\
&= \innerprod{\ln q}{p-q} \\
&=  \innerprod{\ln q + \1}{p-q} \\
&= \innerprod{\grad \psi_1(q)}{p-q}.
\end{align*}
Combining the result of these cases with \eqref{eq:breg-div-cont-first-term}, for $k \in \{0,1\}$,
\begin{align*}
\lim_{\alpha \to k} D_{\psi_{\alpha}}(p,q)
&= \lim_{\alpha \to k} \paren{\psi_\alpha(p) + w_k(\alpha) - \psi_\alpha(q) - w_k(\alpha) - \innerprod{\grad \psi_\alpha(q)}{p-q}} \\
&= \lim_{\alpha \to k}  \paren{\psi_\alpha(p) + w_k(\alpha)} - \lim_{\alpha \to k} \paren{\psi_\alpha(q) + w_k(\alpha)} - \lim_{\alpha \to k} \paren{ \innerprod{\grad \psi_\alpha(q)}{p-q}} \\
&= \psi_k(p) - \psi_k(q) - \innerprod{\grad \psi_k(q)}{p-q} \\
&= D_{\psi_k}(p,q).
\end{align*}
\end{proof}

\subsubsection*{Proof of Proposition \ref{prop:closed-form}}
\label{app:pf-closed-form}

\begin{proposition*}
For $\alpha \in [0,1]$ and $\eta > 0$, Equation \eqref{eq:mirror-step} with the mirror map $\psi_\alpha$ has a unique solution, which we denote $\hat p_m(\placeholder \, ;\alpha)$. Moreover, there exists a scalar $\hat\lambda$ such that $(p,\lambda) = (\hat p_m(\placeholder \, ;\alpha), \hat\lambda)$ is the unique pair in $\simplex \times \Re$ which satisfies
\begin{equation*}
\grad \psi_\alpha(p) = \grad \psi_\alpha(\pi_m) + \eta \, s_m - \lambda \1 \quad \text{\normalfont (Equation \ref{eq:stationarity})}.
\end{equation*}
\end{proposition*}
\begin{proof}[Proof of Proposition \ref{prop:closed-form}]
Fix $\alpha \in [0,1]$ and $\eta > 0$. We first solve \eqref{eq:stationarity} for $p$ at a fixed $\lambda$, which reduces \eqref{eq:stationarity} to a scalar equation in $\lambda$. We then show that this scalar equation has a unique solution $\hat \lambda$, so that \eqref{eq:stationarity} has a unique solution in $\simplex \times \Re$. Finally, we show that the first entry of this solution coincides with the unique maximizer of the objective in \eqref{eq:mirror-step} over $\simplex$.

Let $(p, \lambda) \in \simplex \times \Re$. We solve \eqref{eq:stationarity} for $p$ in terms of $\lambda$. For $\alpha = 1$, the condition in \eqref{eq:stationarity} can be written as
\begin{equation*}
p(v) = \pi_m(v) \, e^{\eta \, s_m(v) - \lambda} \qquad \text{(cf. Equation \ref{eq:closed-form-kl})}.
\end{equation*}
For $\alpha \in [0,1)$, the condition in \eqref{eq:stationarity} can be written as
\begin{equation*}
p(v)^{\alpha - 1} = (1 - \alpha) \paren{\lambda - b(v)},
\qquad \text{where} \quad
b(v) \coloneq \eta \, s_m(v) - \frac{\pi_m(v)^{\alpha - 1}}{1 - \alpha}.
\end{equation*}
Since the left-hand side is positive, this can hold only if $\lambda > B \coloneq \max_{v \in \Vocab} b(v)$, in which case it is equivalent to
\begin{equation*}
p(v) = \parenb{(1 - \alpha)\paren{\lambda - b(v)}}^{-\frac{1}{1 - \alpha}} = \frac{\pi_m(v)}{\paren{1 + (1-\alpha) \pi_m(v)^{1 - \alpha} \paren{\lambda - \eta \, s_m(v)}}^{\frac{1}{1-\alpha}}} \qquad \text{(cf. Equation \ref{eq:closed-form})}.
\end{equation*}
To treat both cases at once, we write the set of feasible $\lambda$ as $\Lambda_1 \coloneq \Re$ and $\Lambda_\alpha \coloneq (B, \infty)$ for $\alpha \in [0,1)$. Then, for any $\lambda \in \Lambda_\alpha$, we write the corresponding solution to \eqref{eq:stationarity} over the set of strictly positive vectors as $\nu_\alpha(\placeholder \, ; \lambda) \in (0,\infty)^\Vocab$, with total mass $Z_\alpha(\lambda)$, defined by
\begin{equation}
\label{eq:Z-fn-def}
\nu_\alpha(v \, ; \lambda) \coloneq
\begin{cases}
\pi_m(v) \, e^{\eta \, s_m(v) - \lambda} & \text{if } \alpha = 1, \\[1mm]
\parenb{(1 - \alpha)\paren{\lambda - b(v)}}^{-\frac{1}{1 - \alpha}} & \text{if } \alpha \in [0,1),
\end{cases}
\qquad
Z_{\alpha}(\lambda) \coloneq \sum_{v \in \Vocab} \nu_\alpha(v \, ; \lambda).
\end{equation}
Thus, $(p, \lambda) \in \simplex \times \Re$ satisfies \eqref{eq:stationarity} if and only if $\lambda \in \Lambda_\alpha$ and $p = \nu_\alpha(\placeholder \, ; \lambda)$. Since $p$ is a valid PMF, this requires $Z_\alpha(\lambda) = 1$. Conversely, if $\lambda \in \Lambda_\alpha$ and $Z_\alpha(\lambda) = 1$, then $\nu_\alpha(\placeholder \, ; \lambda)$ is a PMF in $\simplex$ and the pair $(\nu_\alpha(\placeholder \, ; \lambda), \lambda)$ satisfies \eqref{eq:stationarity}. Thus, the solutions of \eqref{eq:stationarity} in $\simplex \times \Re$ are exactly the pairs
\begin{equation*}
(p, \lambda) = \parenb{\nu_\alpha(\placeholder \, ; \lambda), \, \lambda} \quad \text{with} \quad \lambda \in \Lambda_\alpha \ \ \text{and} \ \ Z_\alpha(\lambda) = 1.
\end{equation*}

We next show that the equation $Z_\alpha(\lambda) = 1$ has a unique solution $\hat \lambda \in \Lambda_\alpha$. We proceed in two cases.

\textbf{Case 1:} $\alpha = 1$.

By \eqref{eq:Z-fn-def}, for all $\lambda \in \Re$,
\begin{equation*}
Z_{1}(\lambda) = e^{-\lambda} \paren{ \sum_{v \in \Vocab} \pi_m(v) \, e^{\eta \, s_m(v)}}.
\end{equation*}
Thus, $\hat \lambda = \ln \sum_{v \in \Vocab} \pi_m(v) e^{\eta \, s_m(v)}$ satisfies $Z_1(\hat \lambda) = 1$. Since $Z_1$ is strictly decreasing over $\Re$, this solution is unique.

\textbf{Case 2:} $\alpha \in [0, 1)$.

By \eqref{eq:Z-fn-def}, $Z_\alpha$ is a continuous function of $\lambda$ on $\Lambda_\alpha = (B, \infty)$. Consider the following two values of $\lambda$:
\begin{equation}
\label{eq:lambda-lo-hi}
\lambda_{\mathrm{lo}} \coloneq B + \frac{1}{1 - \alpha},
\qquad
\lambda_{\mathrm{hi}} \coloneq \max_{v \in \Vocab} \; \eta \, s_m(v).
\end{equation}
Since $\pi_m(v) \leq 1$, we have $b(v) \leq \eta \, s_m(v) - \frac{1}{1-\alpha}$ for all $v \in \Vocab$, and thus $B = \max_{v \in \Vocab} b(v) < \lambda_{\mathrm{lo}} \leq \lambda_{\mathrm{hi}}$. Further,
\begin{equation*}
    Z_{\alpha} (\lambda_{\mathrm{hi}}) \leq 1 \leq Z_{\alpha} (\lambda_{\mathrm{lo}}).
\end{equation*}
The first inequality follows since
\begin{align*}
Z_{\alpha} (\lambda_{\mathrm{hi}}) &= \sum_{v\in\Vocab} \parenb{(1 - \alpha)\paren{\lambda_{\mathrm{hi}} - b(v)}}^{-\frac{1}{1 - \alpha}} \\
&\leq \sum_{v\in\Vocab} \parenb{(1 - \alpha)\paren{ \eta s_m(v) - b(v)}}^{-\frac{1}{1 - \alpha}} \\
&= \sum_{v\in\Vocab} \pi_m(v) = 1,
\end{align*}
and the second since keeping only the summand of $Z_\alpha(\lambda_{\mathrm{lo}})$ with $b(v) = B$ gives
\begin{equation*}
Z_{\alpha} (\lambda_{\mathrm{lo}}) \geq \paren{(1 - \alpha)\paren{B + \frac{1}{1-\alpha} - B}}^{-\frac{1}{1 - \alpha}} = 1.
\end{equation*}
Thus, by the intermediate value theorem, there must exist $\hat \lambda \in [\lambda_{\mathrm{lo}},\lambda_{\mathrm{hi}}] \mathrel{\subset} \Lambda_\alpha$ such that $Z_{\alpha}(\hat \lambda) = 1$. Lastly, for all $\lambda \in \Lambda_\alpha$, differentiating \eqref{eq:Z-fn-def} with respect to $\lambda$ gives
\begin{equation}
\label{eq:Z-deriv}
\frac{d}{d\lambda} Z_{\alpha}(\lambda) = -\sum_{v \in \Vocab} \nu_\alpha(v \, ; \lambda)^{2 - \alpha} < 0.
\end{equation}
Thus, $Z_\alpha$ is strictly decreasing on $\Lambda_\alpha$, so this solution to $Z_{\alpha}(\lambda) = 1$ must be unique.

In both cases, $Z_\alpha(\lambda) = 1$ has a unique solution $\hat \lambda \in \Lambda_\alpha$. Thus, by the above, $(p, \lambda) = (\nu_\alpha(\placeholder \, ; \hat \lambda), \hat \lambda)$ is the unique pair in $\simplex \times \Re$ which satisfies \eqref{eq:stationarity}.

Going forward, we define the first entry of the solution to \eqref{eq:stationarity} as
\begin{equation*}
\hat p \coloneq \nu_\alpha(\placeholder \, ; \hat \lambda).
\end{equation*}
We lastly show that $\hat p$ is the unique solution of \eqref{eq:mirror-step}. For $p \in \simplex$, we define $G(p)$ as the objective in \eqref{eq:mirror-step} with mirror map $\psi_{\alpha}$,
\begin{equation*}
G(p) \mathrel{\coloneq} \innerprod{s_m}{p} - \frac{1}{\eta} D_{\psi_{\alpha}}(p, \pi_m).
\end{equation*}
We must show that $\hat p$ is the unique maximizer of $G$ over $\simplex$.

By Lemma \ref{lem:tsallis-facts}, $\psi_{\alpha}$ is differentiable and strictly convex on the convex open set $(0,\infty)^{\Vocab}$, which contains $\simplex$. Then, by the three-point identity in Lemma \ref{lem:bregman-basics} (iv), for any $q \in \simplex$,
\begin{equation*}
D_{\psi_{\alpha}}(q, \hat p) + D_{\psi_{\alpha}}(\hat p, \pi_m) - D_{\psi_{\alpha}}(q,\pi_m) = \innerprod{\grad \psi_\alpha (\pi_m) - \grad \psi_\alpha(\hat p)}{q - \hat p}.
\end{equation*}
Rearranging,
\begin{equation*}
D_{\psi_{\alpha}}(q,\pi_m) - D_{\psi_{\alpha}}(\hat p, \pi_m) = D_{\psi_{\alpha}}(q, \hat p)  + \innerprod{\grad \psi_\alpha(\hat p) - \grad \psi_\alpha (\pi_m)}{q - \hat p}.
\end{equation*}
Then since $(\hat p, \hat \lambda)$ satisfies \eqref{eq:stationarity}, applying $\grad \psi_\alpha(\hat p) - \grad \psi_\alpha (\pi_m) = \eta \, s_m - \hat \lambda \1$ gives that
\begin{align*}
D_{\psi_{\alpha}}(q,\pi_m) - D_{\psi_{\alpha}}(\hat p, \pi_m)
&= D_{\psi_{\alpha}}(q, \hat p)  + \innerprod{\eta \, s_m - \hat \lambda \1}{q - \hat p} \\
&= D_{\psi_{\alpha}}(q, \hat p) + \innerprod{\eta \, s_m}{q - \hat p},
\end{align*}
since $\innerprod{\hat \lambda \1}{q-\hat p} = \hat \lambda \parenb{\innerprod{\1}{q} - \innerprod{\1}{\hat p}} = 0$, as $q,\hat p \in \simplex$ are both valid PMFs which sum to one.
Further rearranging gives that
\begin{equation*}
\innerprod{s_m}{\hat p} - \frac{1}{\eta} D_{\psi_{\alpha}}(\hat p, \pi_m) - \paren{\innerprod{s_m}{q} - \frac{1}{\eta} D_{\psi_{\alpha}}(q,\pi_m)}  = \frac{1}{\eta} D_{\psi_{\alpha}}(q, \hat p).
\end{equation*}
Applying the definition of $G$ then gives
\begin{equation*}
G(\hat p) - G(q) = \frac{1}{\eta} D_{\psi_{\alpha}}(q, \hat p).
\end{equation*}
By Lemma \ref{lem:bregman-basics} (i), $D_{\psi_{\alpha}}(q, \hat p) > 0$ for all $q \neq \hat p$, so
\begin{equation*}
    G(\hat p) > G(q) \quad \forall q \in \simplex, \, q \neq \hat p.
\end{equation*}
Thus, $\hat p$ uniquely maximizes $G$ over $\simplex$, i.e., \eqref{eq:mirror-step} has the unique solution $\hat p_m(\placeholder \, ;\alpha) = \hat p = \nu_\alpha(\placeholder \, ; \hat \lambda)$, and using the results above, $(p,\lambda) = (\hat p_m(\placeholder \, ;\alpha), \hat\lambda)$ is the unique solution to \eqref{eq:stationarity} over $\simplex \times \Re$.
\end{proof}

\subsubsection{Section \ref{sec:role-alpha}}

\subsubsection*{Proof of Proposition \ref{prop:breg-div-second-order}}
\label{app:pf-hessian}

\begin{proposition*}
Let $\alpha \in [0,1]$. For every $p \in \simplex$, $\grad^2 \psi_\alpha(p) \!=\! \diag \!\paren{p(v)^{\alpha - 2}}_{v \in \Vocab}$, and for every perturbation $u \in \Re^{\Vocab}$ with $\sum_v u(v) = 0$,
\begin{equation*}
D_{\psi_\alpha}(\pi_m + u, \pi_m)
= \frac{1}{2} \sum_{v \in \Vocab} \pi_m(v)^{\alpha} \paren{\frac{u(v)}{\pi_m(v)}}^2 + \mathcal{O}\paren{\norm{u}{}^3} \quad \text{as } u \to 0.
\end{equation*}
\end{proposition*}

\begin{proof}[Proof of Proposition \ref{prop:breg-div-second-order}]
Since $\pi_m \in \simplex$ has full support, $\pi_m + u \in (0, \infty)^{\Vocab}$ for every sufficiently small $u$. By Lemma \ref{lem:tsallis-facts}, $\psi_\alpha$ is infinitely differentiable on $(0, \infty)^{\Vocab}$, so by Taylor's theorem around $\pi_m$,
\begin{equation*}
\psi_\alpha(\pi_m + u) = \psi_\alpha(\pi_m) + \innerprod{\grad \psi_\alpha(\pi_m)}{u} + \frac{1}{2} u^\top \grad^2 \psi_\alpha(\pi_m) \, u + \mathcal{O}\paren{\norm{u}{}^3}.
\end{equation*}
Thus, by the definition \eqref{eq:bregman}, and using the fact that $\grad^2 \psi_\alpha(\pi_m) = \diag\paren{\pi_m(v)^{\alpha - 2}}_{v \in \Vocab}$ from Lemma \ref{lem:tsallis-facts},
\begin{align*}
D_{\psi_\alpha}(\pi_m + u, \pi_m)
&= \psi_\alpha(\pi_m + u) - \psi_\alpha(\pi_m) - \innerprod{\grad \psi_\alpha(\pi_m)}{u} \\
&= \frac{1}{2} u^\top \grad^2 \psi_\alpha(\pi_m) \, u + \mathcal{O}\paren{\norm{u}{}^3} \\
&= \frac{1}{2} \sum_{v \in \Vocab} \pi_m(v)^{\alpha - 2} \, u(v)^2 + \mathcal{O}\paren{\norm{u}{}^3} \\
&= \frac{1}{2} \sum_{v \in \Vocab} \pi_m(v)^{\alpha} \paren{\frac{u(v)}{\pi_m(v)}}^2 + \mathcal{O}\paren{\norm{u}{}^3}.
\end{align*}
\end{proof}

\subsubsection*{Proof of Proposition \ref{prop:log-ratio}}
\label{app:pf-log-ratio}
\begin{proposition*}
Let $\alpha \in [0, 1]$ and $\eta > 0$, and fix the score $s_m$. For every token $v \in \Vocab$, the log-ratio of the mass assigned to $v$ by the update $\hat p_m(v \, ; \alpha)$ relative to $\pi_m(v)$ satisfies
\begin{equation*}
\ln \frac{\hat p_m(v \, ; \alpha)}{\pi_m(v)}
= \pi_m(v)^{1 - \alpha} \paren{\eta \, s_m(v) - \hat \lambda} + \mathcal{O}\paren{\pi_m(v)^{2(1 - \alpha)}}
\quad \text{as } \pi_m(v) \to 0.
\end{equation*}
\end{proposition*}
\begin{proof}[Proof of Proposition \ref{prop:log-ratio}]
Fix $v \in \Vocab$, the step size $\eta > 0$ and the score $s_m$, and write $\hat p_m = \hat p_m(\placeholder \, ; \alpha)$ and $\psi = \psi_\alpha$. By Proposition \ref{prop:closed-form}, Equation \eqref{eq:stationarity} holds at $(p, \lambda) = (\hat p_m, \hat \lambda)$, so for every $u \in \Vocab$,
\begin{equation}
\label{eq:app-stationarity-coord}
[\grad \psi(\hat p_m)](u) - [\grad \psi(\pi_m)](u) = \eta \, s_m(u) - \hat \lambda.
\end{equation}

We first show that $\eta \, s_m(v) - \hat \lambda$ is bounded independently of $\pi_m(v)$. Since $\hat p_m$ and $\pi_m$ both sum to one, there exist $u, u' \in \Vocab$ with
\begin{equation*}
\hat p_m(u) \geq \pi_m(u)
\qquad \text{and} \qquad
\hat p_m(u') \leq \pi_m(u').
\end{equation*}
By Lemma \ref{lem:tsallis-facts}, $[\grad \psi(p)](u)$ depends only on $p(u)$ and is strictly increasing in it. Thus, the left-hand side of Equation \eqref{eq:app-stationarity-coord} is non-negative at $u$ and non-positive at $u'$, i.e.,
\begin{equation*}
\eta \, s_m(u') \leq \hat \lambda \leq \eta \, s_m(u).
\end{equation*}
Hence $\eta \min_{w} s_m(w) \leq \hat \lambda \leq \eta \max_{w} s_m(w)$, and so
\begin{equation}
\label{eq:app-score-bound}
\abs{\eta \, s_m(v) - \hat \lambda} \leq C \coloneq \eta \paren{\max_{w \in \Vocab} s_m(w) - \min_{w \in \Vocab} s_m(w)},
\end{equation}
where $C$ does not depend on $\pi_m(v)$. We now consider two cases.

\textbf{Case 1:} $\alpha = 1$.

Dividing Equation \eqref{eq:closed-form-kl} by $\pi_m(v)$ and taking logarithms,
\begin{equation*}
\ln \frac{\hat p_m(v)}{\pi_m(v)} = \eta \, s_m(v) - \hat \lambda = \pi_m(v)^{1 - \alpha} \paren{\eta \, s_m(v) - \hat \lambda},
\end{equation*}
since $\pi_m(v)^{0} = 1$. Thus Equation \eqref{eq:log-ratio} holds with no remainder.

\textbf{Case 2:} $\alpha \in [0, 1)$.

Define
\begin{equation*}
y \coloneq (1 - \alpha) \, \pi_m(v)^{1 - \alpha} \paren{\eta \, s_m(v) - \hat \lambda}.
\end{equation*}
Then the base of the power in Equation \eqref{eq:closed-form} is $1 - y$, and dividing Equation \eqref{eq:closed-form} by $\pi_m(v)$ gives
\begin{equation*}
\frac{\hat p_m(v)}{\pi_m(v)} = (1 - y)^{-\frac{1}{1 - \alpha}}.
\end{equation*}
Since $\hat p_m(v) > 0$, the base $1 - y$ is positive. Taking logarithms,
\begin{equation*}
\ln \frac{\hat p_m(v)}{\pi_m(v)} = -\frac{\ln(1 - y)}{1 - \alpha}.
\end{equation*}
Since $\frac{y}{1 - \alpha} = \pi_m(v)^{1 - \alpha} \paren{\eta \, s_m(v) - \hat \lambda}$, subtracting it from both sides gives
\begin{equation}
\label{eq:app-log-ratio-remainder}
\ln \frac{\hat p_m(v)}{\pi_m(v)} - \pi_m(v)^{1 - \alpha} \paren{\eta \, s_m(v) - \hat \lambda} = \frac{-\ln(1 - y) - y}{1 - \alpha}.
\end{equation}

It remains to bound the right-hand side. By Equation \eqref{eq:app-score-bound},
\begin{equation*}
\abs{y} \leq (1 - \alpha) \, C \, \pi_m(v)^{1 - \alpha},
\end{equation*}
which tends to $0$ as $\pi_m(v) \to 0$, since $1 - \alpha > 0$. Thus $\abs{y} \leq \frac{1}{2}$ for all sufficiently small $\pi_m(v)$. For such $y$, the Taylor series $-\ln(1 - y) = \sum_{k \geq 1} \frac{y^k}{k}$ converges, and
\begin{equation*}
\abs{-\ln(1 - y) - y}
= \abs{\sum_{k \geq 2} \frac{y^k}{k}}
\leq \frac{1}{2} \sum_{k \geq 2} \abs{y}^k
= \frac{y^2}{2 (1 - \abs{y})}
\leq y^2.
\end{equation*}
Substituting this into Equation \eqref{eq:app-log-ratio-remainder},
\begin{equation*}
\abs{\ln \frac{\hat p_m(v)}{\pi_m(v)} - \pi_m(v)^{1 - \alpha} \paren{\eta \, s_m(v) - \hat \lambda}}
\leq \frac{y^2}{1 - \alpha}
\leq (1 - \alpha) \, C^2 \, \pi_m(v)^{2 (1 - \alpha)},
\end{equation*}
which is $\mathcal{O}\paren{\pi_m(v)^{2(1 - \alpha)}}$ as $\pi_m(v) \to 0$. This proves Equation \eqref{eq:log-ratio}.
\end{proof}

\subsubsection*{Proof of Proposition \ref{prop:cap}}
\label{app:pf-cap}
\begin{proposition*}
Let $\alpha \in [0, 1]$ and $\eta > 0$. If a token $v \in \Vocab$ satisfies
\begin{equation*}
(1 - \alpha) \, \eta \, \pi_m(v)^{1 - \alpha} \paren{s_m(v) - \bar s_m^{[\alpha]}} < 1,
\end{equation*}
then
\begin{equation*}
\frac{\hat p_m(v \, ; \alpha)}{\pi_m(v)} \leq
\begin{cases}
\paren{1 - (1 - \alpha) \, \eta \, \pi_m(v)^{1 - \alpha} \paren{s_m(v) - \bar s_m^{[\alpha]}}}^{-\frac{1}{1 - \alpha}} & \text{if } \alpha \in [0, 1), \\[2mm]
e^{\eta \paren{s_m(v) - \bar s_m^{[1]}}} & \text{if } \alpha = 1.
\end{cases}
\end{equation*}
\end{proposition*}
\begin{proof}[Proof of Proposition \ref{prop:cap}]
Fix $v \in \Vocab$, write $\hat p_m = \hat p_m(\placeholder \, ; \alpha)$, and recall the weighted mean score $\bar s_m^{[\alpha]}$ of Equation \eqref{eq:sharp-mean-score}. In both cases, we first show that
\begin{equation}
\label{eq:app-lambda-lb}
\hat \lambda \geq \eta \, \bar s_m^{[\alpha]},
\end{equation}
and then substitute this bound into the closed form of the update.

\textbf{Case 1:} $\alpha = 1$.

By Equation \eqref{eq:closed-form-kl}, $\ln \frac{\hat p_m(u)}{\pi_m(u)} = \eta \, s_m(u) - \hat \lambda$ for every $u \in \Vocab$. Since the logarithm is concave, Jensen's inequality gives
\begin{equation*}
\sum_{u \in \Vocab} \pi_m(u) \paren{\eta \, s_m(u) - \hat \lambda}
= \sum_{u \in \Vocab} \pi_m(u) \ln \frac{\hat p_m(u)}{\pi_m(u)}
\leq \ln \sum_{u \in \Vocab} \pi_m(u) \, \frac{\hat p_m(u)}{\pi_m(u)}
= \ln 1 = 0.
\end{equation*}
Further expanding the left-hand side above,
\begin{equation*}
\sum_{u \in \Vocab} \pi_m(u) \paren{\eta \, s_m(u) - \hat \lambda}
= \eta \, \paren{\sum_{u \in \Vocab} \pi_m(u) \, s_m(u)} - \hat \lambda \, \paren{\sum_{u \in \Vocab} \pi_m(u)}
= \eta \, \bar s_m^{[1]} - \hat \lambda.
\end{equation*}
Thus, $\eta \, \bar s_m^{[1]} - \hat \lambda \leq 0$, which proves Equation \eqref{eq:app-lambda-lb}. Applying this to the ratio $\hat p_m / \pi_m$,
\begin{equation*}
\frac{\hat p_m(v)}{\pi_m(v)} = e^{\eta \, s_m(v) - \hat \lambda} \leq e^{\eta \paren{s_m(v) - \bar s_m^{[1]}}}.
\end{equation*}

\textbf{Case 2:} $\alpha \in [0, 1)$.

Dividing Equation \eqref{eq:closed-form} by $\pi_m(u)$ and raising both sides to the power $-(1 - \alpha)$, for every $u \in \Vocab$,
\begin{equation}
\label{eq:app-ratio-power}
\paren{\frac{\hat p_m(u)}{\pi_m(u)}}^{-(1 - \alpha)} = 1 + (1 - \alpha) \, \pi_m(u)^{1 - \alpha} \paren{\hat \lambda - \eta \, s_m(u)}.
\end{equation}
For $x > 0$, define
\begin{equation*}
    \phi(x) \coloneq \frac{x^{-(1 - \alpha)} - 1}{1 - \alpha}.
\end{equation*}
Then $\phi$ is convex since $\phi''(x) = (2 - \alpha) \, x^{-(3 - \alpha)} > 0$, and it satisfies $\phi(1) = 0$. Rearranging Equation \eqref{eq:app-ratio-power} and multiplying by $\pi_m(u)$,
\begin{equation*}
\pi_m(u)^{2 - \alpha} \paren{\hat \lambda - \eta \, s_m(u)} = \pi_m(u) \, \phi\paren{\frac{\hat p_m(u)}{\pi_m(u)}}.
\end{equation*}
Summing over $u \in \Vocab$ and applying Jensen's inequality,
\begin{align*}
\hat \lambda \paren{\sum_{u \in \Vocab} \pi_m(u)^{2 - \alpha}} - \eta \paren{\sum_{u \in \Vocab} \pi_m(u)^{2 - \alpha} \, s_m(u)}
&= \sum_{u \in \Vocab} \pi_m(u) \, \phi\paren{\frac{\hat p_m(u)}{\pi_m(u)}} \\
&\geq \phi\paren{\sum_{u \in \Vocab} \pi_m(u) \, \frac{\hat p_m(u)}{\pi_m(u)}}
= \phi(1) = 0.
\end{align*}
Thus,
\begin{equation*}
\hat \lambda \geq \eta \, \frac{\sum_{u \in \Vocab} \pi_m(u)^{2-\alpha} s_m(u)}{\sum_{u \in \Vocab} \pi_m(u)^{2-\alpha}} = \eta \, \bar{s}_m^{[\alpha]},
\end{equation*}
which proves Equation \eqref{eq:app-lambda-lb}. Substituting it into Equation \eqref{eq:app-ratio-power} at $u = v$,
\begin{equation*}
\paren{\frac{\hat p_m(v)}{\pi_m(v)}}^{-(1 - \alpha)} \geq 1 - (1 - \alpha) \, \eta \, \pi_m(v)^{1 - \alpha} \paren{s_m(v) - \bar s_m^{[\alpha]}}.
\end{equation*}
The right-hand side is positive by assumption, and $x \mapsto x^{-\frac{1}{1 - \alpha}}$ is decreasing on $(0, \infty)$, so raising both sides to the power $-\frac{1}{1 - \alpha}$ gives
\begin{equation*}
\frac{\hat p_m(v)}{\pi_m(v)} \leq \paren{1 - (1 - \alpha) \, \eta \, \pi_m(v)^{1 - \alpha} \paren{s_m(v) - \bar s_m^{[\alpha]}}}^{-\frac{1}{1 - \alpha}}. \qedhere
\end{equation*}
\end{proof}

\subsubsection{Section \ref{sec:rate}}

\subsubsection*{Proof of Proposition \ref{prop:ascent}}
\label{app:pf-ascent}

\begin{proposition*}
Let $\alpha \in [0, 1]$ and $\eta > 0$, and write $\hat p_m = \hat p_m(\placeholder \, ; \alpha, \eta)$. Then
\begin{equation*}
\tilde A_m(\hat p_m) - \tilde A_m(\pi_m) = \frac{1}{\eta} \Bigl( D_{\psi_\alpha}(\hat p_m, \pi_m) + D_{\psi_\alpha}(\pi_m, \hat p_m) \Bigr) \geq 0.
\end{equation*}
Further, if the score $s_m$ is non-constant over $\Vocab$, the gain is strictly positive, $\tilde A_m(\hat p_m) - \tilde A_m(\pi_m) > 0$.
\end{proposition*}
\begin{proof}[Proof of Proposition \ref{prop:ascent}]
Write $\psi = \psi_\alpha$. By the definition \eqref{eq:bregman},
\begin{align*}
D_\psi(\hat p_m, \pi_m) + & D_\psi(\pi_m,  \hat p_m) \\
&= \psi(\hat p_m) - \psi(\pi_m) - \innerprod{\grad \psi(\pi_m)}{\hat p_m - \pi_m} + \psi(\pi_m) - \psi(\hat p_m) - \innerprod{\grad \psi(\hat p_m)}{\pi_m - \hat p_m} \\
&= \innerprod{\grad \psi(\hat p_m) - \grad \psi(\pi_m)}{\hat p_m - \pi_m}.
\end{align*}
By Proposition \ref{prop:closed-form}, Equation \eqref{eq:stationarity} holds at $(p, \lambda) = (\hat p_m, \hat \lambda)$, i.e.,
\begin{equation*}
\grad \psi(\hat p_m) - \grad \psi(\pi_m) = \eta \, s_m - \hat \lambda \1.
\end{equation*}
Substituting this above,
\begin{align*}
D_\psi(\hat p_m, \pi_m) + D_\psi(\pi_m, \hat p_m)
&= \innerprod{\eta \, s_m - \hat \lambda \1}{\hat p_m - \pi_m} \\
&= \eta \innerprod{s_m}{\hat p_m - \pi_m} - \hat \lambda \innerprod{\1}{\hat p_m - \pi_m} \\
&= \eta \innerprod{s_m}{\hat p_m - \pi_m} \\
&= \eta \paren{\tilde A_m(\hat p_m) - \tilde A_m(\pi_m)},
\end{align*}
where the third step uses $\innerprod{\1}{\hat p_m} = \innerprod{\1}{\pi_m} = 1$, and the last step uses the definition of $\tilde A_m$ in Equation \eqref{eq:surrogate}. Thus,
\[
\tilde A_m(\hat p_m) - \tilde A_m(\pi_m) = \frac{1}{\eta} \paren{D_\psi(\hat p_m, \pi_m) + D_\psi(\pi_m, \hat p_m)}.
\]

By Lemma \ref{lem:bregman-basics}(i),
\[
D_\psi(\hat p_m, \pi_m) + D_\psi(\pi_m, \hat p_m) \geq 0,
\]
and further
\[
D_\psi(\hat p_m, \pi_m) + D_\psi(\pi_m, \hat p_m) = 0 \iff \hat p_m = \pi_m.
\]
To complete the proof, we show that if $s_m$ is non-constant, then $\hat p_m \neq \pi_m$ and thus $\tilde A_m(\hat p_m) - \tilde A_m(\pi_m) = \frac{1}{\eta} \paren{D_\psi(\hat p_m, \pi_m) + D_\psi(\pi_m, \hat p_m)} > 0$.

Assume for contradiction that $\hat p_m = \pi_m$. Then Equation \eqref{eq:stationarity} at $(p, \lambda) = (\hat p_m, \hat \lambda)$ reads
\begin{equation*}
\grad \psi(\pi_m) = \grad \psi(\pi_m) + \eta \, s_m - \hat \lambda \1,
\end{equation*}
which implies that
\[
s_m = \frac{\hat \lambda}{\eta} \1,
\]
which contradicts that $s_m$ is non-constant. Thus $\hat p_m \neq \pi_m$, so by Lemma \ref{lem:bregman-basics}(i) both divergences are strictly positive, and thus, $\tilde A_m(\hat p_m) - \tilde A_m(\pi_m) > 0$.
\end{proof}

\subsubsection*{Proof of Theorem \ref{thm:rate}}
\label{app:pf-rate}

\begin{theorem*}
Let $\alpha \in [0, 1]$, and let the score $s_m$ be non-constant over the vocabulary $\Vocab$. Then the derivatives in Equation \eqref{eq:rates} exist, $\tilde \Gamma_m(\alpha) > 0$, and
\begin{equation*}
\frac{\Gamma_m(\alpha)}{\tilde \Gamma_m(\alpha)}
= \frac{\Cov_{\Tok \sim \pi_m^{[\alpha]}}\paren{s_m(\Tok), \, a_m(\Tok)}}{\Var_{\Tok \sim \pi_m^{[\alpha]}}\paren{s_m(\Tok)}}.
\end{equation*}
\end{theorem*}
\begin{proof}[Proof of Theorem \ref{thm:rate}]
Write $\psi = \psi_\alpha$. For $\eta > 0$, write $p(\eta) \coloneq \hat p_m(\placeholder \, ; \alpha, \eta)$ and $\lambda(\eta)$ for its multiplier $\hat \lambda$ in Proposition \ref{prop:closed-form}, and set $p(0) \coloneq \pi_m$ and $\lambda(0) \coloneq 0$. The derivatives in Equation \eqref{eq:rates} are taken from the right at $\eta = 0$. We first show that $\eta \mapsto p(\eta)$ is differentiable from the right at $\eta = 0$ and compute its derivative, and then compute $\Gamma_m(\alpha)$ and $\tilde \Gamma_m(\alpha)$.

Define $F : (0, \infty)^{\Vocab} \times \Re \times \Re \to \Re^{\abs{\Vocab} + 1}$ by
\begin{equation*}
F(p, \lambda, \eta) \coloneq
\begin{pmatrix}
\grad \psi(p) - \grad \psi(\pi_m) - \eta \, s_m + \lambda \1 \\
\innerprod{\1}{p} - 1
\end{pmatrix},
\end{equation*}
where the first block has $\abs{\Vocab}$ entries and the second block has one entry. Thus $F(p, \lambda, \eta) = 0$ if and only if $p$ sums to one and $(p, \lambda)$ satisfies Equation \eqref{eq:stationarity} with step size $\eta$. By Proposition \ref{prop:closed-form}, for every $\eta > 0$,
\begin{equation*}
F(p(\eta), \lambda(\eta), \eta) = 0.
\end{equation*}
This also holds at $\eta = 0$, since $F(\pi_m, 0, 0) = 0$. By Lemma \ref{lem:tsallis-facts}, $\psi$ is infinitely differentiable on $(0, \infty)^{\Vocab}$, so $F$ is continuously differentiable.

Next, we compute the Jacobian matrix $J \in \Re^{(\abs{\Vocab} + 1) \times (\abs{\Vocab} + 1)}$ of $F$ with respect to $(p, \lambda)$ at $(\pi_m, 0, 0)$. In block form, this gives that
\begin{equation*}
J =
\begin{pmatrix}
H & \1 \\
\1^\top & 0
\end{pmatrix},
\qquad
\text{for}
\ 
H \coloneq \diag\paren{\pi_m(v)^{\alpha - 2}}_{v \in \Vocab} \in \Re^{\Vocab \times \Vocab}.
\end{equation*}
We show that $J$ is invertible. Suppose that $x \in \Re^{\Vocab}$ and $\mu \in \Re$ satisfy $J \begin{pmatrix} x \\ \mu \end{pmatrix} = 0$. This gives the pair of equations
\begin{equation*}
H x + \mu \1 = 0,
\qquad
\1^\top x = 0.
\end{equation*}
Since $H$ is diagonal with positive entries, it is invertible, and the first equation gives $x = -\mu H^{-1} \1$. Substituting this into the second equation,
\begin{equation*}
0 = \1^\top x = -\mu \, \1^\top H^{-1} \1 = -\mu \sum_{v \in \Vocab} \pi_m(v)^{2 - \alpha}.
\end{equation*}
Since $\sum_v \pi_m(v)^{2 - \alpha} > 0$, we have $\mu = 0$, and thus $x = 0$. Thus $J \begin{pmatrix} x \\ \mu \end{pmatrix} = 0$ implies $x=0$ and $\mu=0$, so $J$ is invertible.

Since $F$ is continuously differentiable, $F(\pi_m, 0, 0) = 0$, and $J$ is invertible, the implicit function theorem gives $\epsilon > 0$ and a continuously differentiable map $\eta \mapsto (\tilde p(\eta), \tilde \lambda(\eta))$ from $(-\epsilon, \epsilon)$ to $(0, \infty)^{\Vocab} \times \Re$ with $(\tilde p(0), \tilde \lambda(0)) = (\pi_m, 0)$ and
\begin{equation*}
F(\tilde p(\eta), \tilde \lambda(\eta), \eta) = 0 \qquad \text{for every } \eta \in (-\epsilon, \epsilon).
\end{equation*}
For $\eta \in (0, \epsilon)$, the last entry of $F$ gives $\innerprod{\1}{\tilde p(\eta)} = 1$, so $\tilde p(\eta) \in \simplex$, and the first block shows that $(\tilde p(\eta), \tilde \lambda(\eta))$ satisfies Equation \eqref{eq:stationarity}. By the uniqueness of this pair in Proposition \ref{prop:closed-form},
\begin{equation*}
(\tilde p(\eta), \tilde \lambda(\eta)) = (p(\eta), \lambda(\eta)) \qquad \text{for every } \eta \in [0, \epsilon),
\end{equation*}
where the case $\eta = 0$ holds by construction. Thus $\eta \mapsto (p(\eta), \lambda(\eta))$ is continuously differentiable on $[0, \epsilon)$, with the derivative at $\eta = 0$ taken from the right.

We now compute the derivatives at $\eta = 0$. Differentiating $F(p(\eta), \lambda(\eta), \eta) = 0$ with respect to $\eta$ at $\eta = 0$ by the chain rule,
\begin{equation*}
J \begin{pmatrix} \dfrac{dp}{d\eta} \Big|_{\eta = 0} \\[4mm] \dfrac{d\lambda}{d\eta} \Big|_{\eta = 0} \end{pmatrix} + \frac{\partial F}{\partial \eta}(\pi_m, 0, 0) = 0.
\end{equation*}
By the definition of $F$,
\begin{equation*}
\frac{\partial F}{\partial \eta}(\pi_m, 0, 0) = \begin{pmatrix} -s_m \\ 0 \end{pmatrix}.
\end{equation*}
Further, multiplying out the blocks of $J$ gives the pair of equations
\begin{equation*}
H \, \frac{dp}{d\eta} \Big|_{\eta = 0} + \frac{d\lambda}{d\eta} \Big|_{\eta = 0} \1 = s_m,
\qquad
\1^\top \frac{dp}{d\eta} \Big|_{\eta = 0} = 0.
\end{equation*}
Multiplying the first equation by $H^{-1} = \diag\paren{\pi_m(v)^{2 - \alpha}}_{v \in \Vocab}$, for every $v \in \Vocab$,
\begin{equation}
\label{eq:app-flow-raw}
\frac{d}{d\eta} \, p(v) \Big|_{\eta = 0} = \pi_m(v)^{2 - \alpha} \paren{s_m(v) - \frac{d\lambda}{d\eta} \Big|_{\eta = 0}}.
\end{equation}
Substituting this into the second equation,
\begin{equation*}
0 = \sum_{v \in \Vocab} \pi_m(v)^{2 - \alpha} \, s_m(v) - \frac{d\lambda}{d\eta} \Big|_{\eta = 0} \sum_{v \in \Vocab} \pi_m(v)^{2 - \alpha}.
\end{equation*}
Rearranging, and using the definition \eqref{eq:sharp-mean-score} of $\pi_m^{[\alpha]}$,
\begin{equation*}
\frac{d\lambda}{d\eta} \Big|_{\eta = 0} = \frac{\sum_{v \in \Vocab} \pi_m(v)^{2 - \alpha} \, s_m(v)}{\sum_{u \in \Vocab} \pi_m(u)^{2 - \alpha}} = \EV{\Tok \sim \pi_m^{[\alpha]}}{s_m(\Tok)}.
\end{equation*}
Substituting this into Equation \eqref{eq:app-flow-raw},
\begin{equation}
\label{eq:app-flow-zero}
\frac{d}{d\eta} \, p(v) \Big|_{\eta = 0} = \pi_m(v)^{2 - \alpha} \paren{s_m(v) - \EV{\Tok \sim \pi_m^{[\alpha]}}{s_m(\Tok)}}.
\end{equation}
Since $A_m(p) = \innerprod{a_m}{p}$ and $\tilde A_m(p) = \innerprod{s_m}{p}$ are linear in $p$, by the chain rule both derivatives in Equation \eqref{eq:rates} exist, with
\begin{equation*}
\Gamma_m(\alpha) = \left\langle a_m, \, \frac{d}{d\eta} \, p \Big|_{\eta = 0} \right\rangle,
\qquad
\tilde \Gamma_m(\alpha) = \left\langle s_m, \, \frac{d}{d\eta} \, p \Big|_{\eta = 0} \right\rangle.
\end{equation*}
Let $c \coloneq \sum_{u \in \Vocab} \pi_m(u)^{2 - \alpha} > 0$, so that $\pi_m^{[\alpha]}(v) = \pi_m(v)^{2 - \alpha} / c$ by Equation \eqref{eq:sharp-mean-score}. Substituting Equation \eqref{eq:app-flow-zero},
\begin{align*}
\Gamma_m(\alpha)
&= \sum_{v \in \Vocab} a_m(v) \, \pi_m(v)^{2 - \alpha} \paren{s_m(v) - \EV{\Tok \sim \pi_m^{[\alpha]}}{s_m(\Tok)}} \\
&= c \sum_{v \in \Vocab} \pi_m^{[\alpha]}(v) \, a_m(v) \paren{s_m(v) - \EV{\Tok \sim \pi_m^{[\alpha]}}{s_m(\Tok)}} \\
&= c \, \EV{\Tok \sim \pi_m^{[\alpha]}}{a_m(\Tok) \paren{s_m(\Tok) - \EV{\Tok' \sim \pi_m^{[\alpha]}}{s_m(\Tok')}}} \\
&= c \, \Cov_{\Tok \sim \pi_m^{[\alpha]}}\paren{s_m(\Tok), \, a_m(\Tok)},
\end{align*}
where the last step uses $\Cov(X, Y) = \mathbb{E}\bigl[Y \paren{X - \mathbb{E}[X]}\bigr]$. By the same steps with $s_m$ in place of $a_m$,
\begin{equation*}
\tilde \Gamma_m(\alpha) = c \, \EV{\Tok \sim \pi_m^{[\alpha]}}{s_m(\Tok) \paren{s_m(\Tok) - \EV{\Tok' \sim \pi_m^{[\alpha]}}{s_m(\Tok')}}} = c \, \Var_{\Tok \sim \pi_m^{[\alpha]}}\paren{s_m(\Tok)}.
\end{equation*}
Since $\pi_m^{[\alpha]} \in \simplex$ has full support and $s_m$ is non-constant, the variance is positive, so $\tilde \Gamma_m(\alpha) > 0$. Dividing,
\begin{equation*}
\frac{\Gamma_m(\alpha)}{\tilde \Gamma_m(\alpha)} = \frac{c \, \Cov_{\Tok \sim \pi_m^{[\alpha]}}\paren{s_m(\Tok), \, a_m(\Tok)}}{c \, \Var_{\Tok \sim \pi_m^{[\alpha]}}\paren{s_m(\Tok)}} = \frac{\Cov_{\Tok \sim \pi_m^{[\alpha]}}\paren{s_m(\Tok), \, a_m(\Tok)}}{\Var_{\Tok \sim \pi_m^{[\alpha]}}\paren{s_m(\Tok)}}. \qedhere
\end{equation*}
\end{proof}

\subsubsection*{Proof of Corollary \ref{cor:sign}}
\label{app:pf-sign}

\begin{corollary*}
Let $\alpha \in [0, 1]$, and suppose $\Cov_{\Tok \sim \pi_m^{[\alpha]}}\paren{s_m(\Tok), \, a_m(\Tok)} > 0$. Then there exists $\eta_0 > 0$ such that $A_m\paren{\hat p_m(\placeholder \, ; \alpha, \eta)} > A_m(\pi_m)$ for every $\eta \in (0, \eta_0)$.
\end{corollary*}
\begin{proof}[Proof of Corollary \ref{cor:sign}]
A constant score has zero covariance with $a_m$, so the assumption implies that $s_m$ is non-constant, and Theorem \ref{thm:rate} applies. By its proof,
\begin{equation*}
\Gamma_m(\alpha) = c \, \Cov_{\Tok \sim \pi_m^{[\alpha]}}\paren{s_m(\Tok), \, a_m(\Tok)} > 0,
\qquad
c = \sum_{u \in \Vocab} \pi_m(u)^{2 - \alpha} > 0.
\end{equation*}
Define $h(\eta) \coloneq A_m\paren{\hat p_m(\placeholder \, ; \alpha, \eta)}$ for $\eta \geq 0$, so that $h(0) = A_m(\pi_m)$. By the definition \eqref{eq:rates} of $\Gamma_m(\alpha)$,
\begin{equation*}
\lim_{\eta \to 0^{+}} \frac{h(\eta) - h(0)}{\eta} = \Gamma_m(\alpha) > 0.
\end{equation*}
Thus there exists $\eta_0 > 0$ such that for every $\eta \in (0, \eta_0)$,
\begin{equation*}
\frac{h(\eta) - h(0)}{\eta} > \frac{\Gamma_m(\alpha)}{2} > 0.
\end{equation*}
Multiplying by $\eta > 0$ gives $h(\eta) > h(0)$, i.e., $A_m\paren{\hat p_m(\placeholder \, ; \alpha, \eta)} > A_m(\pi_m)$.
\end{proof}

\subsection{\texorpdfstring{Solving for $\hat \lambda$}{Solving for Lambda}}
\label{app:pf-lambda}
By Proposition \ref{prop:closed-form}, the Bregman tilting update is determined by the multiplier $\hat \lambda$, the unique scalar for which the right-hand side of Equation \eqref{eq:closed-form-kl} or \eqref{eq:closed-form} sums to one. We describe the properties of $\hat \lambda$ and the algorithm we use to compute it at each decoding step.

\paragraph{Properties of $\hat \lambda$.} From the proof of Proposition \ref{prop:closed-form} (Appendix \ref{app:pf-closed-form}), we have that $\hat p_m(\placeholder \, ; \alpha) = \nu_\alpha(\placeholder \, ; \hat \lambda)$, where $\nu_\alpha(\placeholder \, ; \lambda)$ and its total mass $Z_\alpha(\lambda)$ are defined in Equation \eqref{eq:Z-fn-def}, and $\hat \lambda$ solves $Z_\alpha(\lambda) = 1$. For $\alpha = 1$, this gives the closed form
\begin{equation*}
\hat \lambda = \ln \sum_{v \in \Vocab} \pi_m(v) \, e^{\eta \, s_m(v)}.
\end{equation*}
For $\alpha \in [0, 1)$, $\hat \lambda$ has no closed form. However, the same proof shows that $\lambda_{\mathrm{lo}} \leq \hat \lambda \leq \lambda_{\mathrm{hi}}$ for the endpoints $\lambda_{\mathrm{lo}}$ and $\lambda_{\mathrm{hi}}$ of Equation \eqref{eq:lambda-lo-hi}. This gives a computable bound for $\hat \lambda$ at inference time:
\begin{equation}
\label{eq:lambda-hat-interval}
\lambda_{\mathrm{lo}} = \paren{\max_{v \in \Vocab} \, \eta s_m(v) - \frac{\pi_m(v)^{\alpha - 1}}{1 - \alpha}} + \frac{1}{1-\alpha} \leq \hat \lambda \leq \eta \paren{\max_{v \in \Vocab} \, s_m(v)} = \lambda_{\mathrm{hi}}.
\end{equation}
The same proof also shows that $Z_\alpha$ is well defined and strictly decreasing on the interval $[\lambda_{\mathrm{lo}},\lambda_{\mathrm{hi}}]$ (Equation \ref{eq:Z-deriv}), so $\hat \lambda$ is the unique solution of $Z_\alpha(\lambda) = 1$ in this interval. Further, $Z_\alpha$ is strictly convex on $[\lambda_{\mathrm{lo}},\lambda_{\mathrm{hi}}]$, since differentiating Equation \eqref{eq:Z-deriv} again gives
\begin{equation*}
\frac{d^2}{d\lambda^2} Z_\alpha(\lambda) = (2 - \alpha) \sum_{v \in \Vocab} \nu_\alpha(v \, ; \lambda)^{3 - 2\alpha} > 0.
\end{equation*}

\paragraph{Root Finding Algorithm.} For $\alpha \in [0, 1)$, we use the properties above to approximate $\hat \lambda$ at each decoding step in three stages: bisection, Newton's method, and normalization.

First, we run bisection on an interval $[\ell, u]$, initialized at the endpoints of Equation \eqref{eq:lambda-hat-interval}, $\ell = \lambda_{\mathrm{lo}}$ and $u = \lambda_{\mathrm{hi}}$. Since $Z_\alpha$ is strictly decreasing with $Z_\alpha(\hat \lambda) = 1$, every step maintains $\ell \leq \hat \lambda \leq u$ while halving the width of the interval. We use 6 bisection steps in our experiments.

Second, we run Newton's method to find the root of $Z_\alpha(\lambda) - 1$, starting from the lower end $\lambda_0 = \ell$ of the final bisection interval. By Equation \eqref{eq:Z-deriv}, the Newton iterates are
\begin{equation*}
\lambda_{k+1} \coloneq \lambda_k - \frac{Z_\alpha(\lambda_k) - 1}{Z_\alpha'(\lambda_k)} = \lambda_k + \frac{Z_\alpha(\lambda_k) - 1}{\sum_{v \in \Vocab} \nu_\alpha(v \, ; \lambda_k)^{2 - \alpha}}.
\end{equation*}
Since $Z_\alpha$ is strictly decreasing and strictly convex on $[\lambda_{\mathrm{lo}},\lambda_{\mathrm{hi}}]$ and $\lambda_0 \leq \hat \lambda$, the Newton iterates increase monotonically toward $\hat \lambda$ without overshoot, i.e., $\lambda_k \leq \lambda_{k+1} \leq \hat \lambda$ for every $k \geq 0$. Further, Newton's method achieves a quadratic convergence rate when the iterates are sufficiently close to $\hat \lambda$ \citep{suli:2003:intro-numerical-analysis}. We use 6 Newton steps in our experiments.

Lastly, we normalize the final iterate to ensure a valid PMF. Writing $\lambda_K$ for the final Newton iterate, we sample the next token from the PMF $\nu_\alpha(v \, ; \lambda_K) / Z_\alpha(\lambda_K)$. This lies in $\simplex$ for any number of steps of our procedure, and it is equal to $\hat p_m(v \, ; \alpha)$ when $\lambda_K = \hat \lambda$ recovers the exact root.

This procedure is cheap relative to the LLM decoding step. Each bisection step evaluates $Z_\alpha$ once, and each Newton step evaluates $Z_\alpha$ and its derivative \eqref{eq:Z-deriv} once. Each evaluation is a single elementwise pass over the vectors $\pi_m$ and $s_m$, which the update at $\alpha = 1$ computes as well. Thus, finding $\hat \lambda$ requires no additional forward pass of the LLM, and its cost of $\mathcal{O}(\abs{\Vocab})$ operations per step, for a fixed number of steps, is negligible next to the two forward passes which produce $\pi_m$ and $s_m$.

\section{Experimental Details}
\label{app:exp-details}

We run all experiments on NVIDIA H100 GPUs. All of the models and benchmarks we use are open source. The rest of this section gives the full details of our experiments.

\subsection{Recursive Self-Aggregation}
\label{app:rsa}

RSA \citep{venkatraman:2025:rsa} scales the iterative refinement procedure along an additional parallel axis. It solves $Q > 1$ chains in parallel that interact through their prompts, so we write $h^{(t,j)}$, $r^{(t,j)}$, and $f^{(t,j)} = \Eval(r^{(t,j)})$ for the prompt, rollout, and feedback of chain $j \in [Q]$ at round $t$. At round 1, every chain is prompted with $h^{(1,j)} = h^{(1)}$. After round $t$, RSA draws for each chain $j$ a set $\S^{(t,j)} \subseteq [Q]$ of $K > 1$ distinct chains, uniformly at random from all subsets of $[Q]$ of size $K$ and independently across chains, so the same rollout can appear in the prompts of several chains. The prompt $h^{(t+1,j)}$ then contains the question, the $K$ rollouts $r^{(t,i)}$ for $i \in \S^{(t,j)}$, and instructions to aggregate them into a single improved solution. While Self-Refine independently refines one chain at a time, RSA mixes knowledge across chains, so a correct step found by one chain can spread to the others. We modify RSA to incorporate feedback by showing each rollout $r^{(t,i)}$ followed by its feedback $f^{(t,i)} = \Eval(r^{(t,i)})$, which reviews that rollout on its own. We use $K = 4$, following the prior work \citep{venkatraman:2025:rsa}.

\subsection{Benchmarks and Grading}
\label{app:benchmarks}

\paragraph{LiveCodeBench.} We use LiveCodeBench \citep{jain:2025:livecodebench}, which collects new problems from programming contests over time so that models can be tested on problems released after their training data. We use the 131 problems of its sixth release that were published from February to April 2025 following the evaluations reported in the model card of Qwen3-4B-Instruct-2507\footnote{\url{https://huggingface.co/Qwen/Qwen3-4B-Instruct-2507}} as well as prior work \citep{hubotter:2026:rl-self-dist}. Of these problems, 82 are from AtCoder, where the program reads standard input and writes standard output, and 49 are from LeetCode, where the program completes a given function. By the labels of the benchmark, 31 problems are easy, 39 are medium, and 61 are hard. Each problem has a few public tests that appear as examples in the problem statement (a median of 3) and a larger set of private tests (a median of 40). We take the last Python code block of each rollout and run it on every one of the exposed and hidden tests in a sandbox, with a limit of 6 seconds of CPU time per test. A rollout is correct if its program passes every test, and a rollout without a code block is incorrect. We reimplement the grading of the official LiveCodeBench evaluation code so that programs run in a sandbox and the feedback can use the result of every public test. On the zero-shot rollouts, our grader agrees with the official code on all but one of the 2069 programs; the exception has a test that runs at the 6 s limit.

\paragraph{MBPP+.} We use MBPP+ \citep{liu:2023:evalplus}, the 378 tasks of the EvalPlus release (v0.2.0) of MBPP \citep{austin:2021:mbpp}. Each task is a one-sentence description of a Python function to write, with its name, and 3 to 7 hand-written test assertions (a median of 3), which the prompt shows as examples; EvalPlus adds about 105 generated tests per task that probe edge cases, which the prompt never shows. We grade as the EvalPlus leaderboard does: the program must define the named function, and a rollout is correct if its function returns the reference output on every hand-written and generated test, compared with EvalPlus's rules (exact equality, with a tolerance for floating-point outputs), each test with EvalPlus's time allowance of max(1 s, 4 times the reference solution's time) of CPU time. Our grader agrees with the EvalPlus checker on 6031 of the 6048 zero-shot programs; the other 17 are cases where the checker's 4 GB memory limit fails its own reference solution. The solver already solves most of MBPP+ zero-shot: its mean pass@1 is 75.5\% (averaged over the four seeds), and at seed 0 all 16 of its zero-shot rollouts are correct on 223 of the 378 tasks (59\%), so on these tasks the mean pass@1 of every method is pinned near 100\% and only losses are possible. We therefore report MBPP+ on its \emph{hard set}: the 155 tasks that were not solved by all 16 rollouts of the seed-0 zero-shot run (its mean pass@1 on them is 39.6\%). The set is fixed before the runs and is the same for every method and seed; since it is chosen from the seed-0 zero-shot rollouts, its round-1 score at seed 0 is selected low, while seeds 1 to 3 draw their own round 1 on the same tasks. The results on the full set of 378 tasks are in Appendix \ref{app:mbpp-full}.

\paragraph{AIME 2025.} We use the 30 problems of AIME 2025 (I and II), as distributed by MathArena \citep{balunovic:2025:matharena}. Each answer is an integer between 0 and 999. We grade the last \texttt{\textbackslash boxed\{\}} expression of each rollout against the reference answer with Math-Verify,\footnote{\url{https://github.com/huggingface/Math-Verify}} and a rollout without one is incorrect.

\paragraph{HMMT February 2025.} We use the 30 problems of HMMT February 2025, also as distributed by MathArena. These are the problems of the three individual rounds of the February 2025 Harvard-MIT Mathematics Tournament, 10 each in algebra and number theory, combinatorics, and geometry. Only 14 of the 30 answers are integers; the others are fractions, radicals, other closed-form expressions, and for one problem a list of two numbers. We grade the last \texttt{\textbackslash boxed\{\}} expression of each rollout against the reference answer with the answer parser of MathArena, which is how its leaderboard grades this competition, and a rollout without one is incorrect.

\subsection{Solver and Feedback}
\label{app:feedback}

\paragraph{Solver.} We use Qwen3-4B-Instruct-2507 \citep{yang:2025:qwen3} as the solver on every benchmark, sampled at temperature 1 and at most 8192 new tokens per rollout on AIME, LiveCodeBench, and MBPP+ and 12288 on HMMT. On HMMT, we observed that an 8192 token cap led to a high truncation rate, so we increase the limit to 12288 in that setting.

\paragraph{Execution feedback.} On LiveCodeBench and MBPP+, the feedback comes from running the program rather than from an LLM. $\Eval(\rollt)$ runs the program of $\rollt$ on the exposed tests and reports the first test it fails. For a wrong answer, it shows the test input, the output of the program, and the expected output. For a runtime error, it shows the error and the line of the program that raised it, and for a time limit, it shows the test input. If the program passes every exposed test, the feedback says so and reminds the model that the hidden tests may still cover edge cases or larger inputs. The feedback never shows the hidden tests. The exact messages are given in Appendix \ref{app:exp-prompts-code}.

The benchmark only labels about 3 tests per problem as public, which can miss many errors. Following prior work \citep{hubotter:2026:rl-self-dist}, we construct the exposed test set as the benchmark's public tests along with a random half of the benchmark's private tests, drawn once and fixed for every run. Thus, we treat half of the benchmark's private tests as hidden, with the other half exposed. We still grade each program on all of its tests. Since the added tests are part of the grade, this setting measures how well each method uses informative feedback, not how well its programs generalize to unseen tests. On MBPP+ we expose a random half of the generated tests in the same way (a median of 55 exposed and 53 hidden tests per task).

\paragraph{Feedback model.} On AIME and HMMT, the feedback comes from an LLM that reviews each rollout. We use Qwen3.5-27B to implement the feedback mechanism $\Eval$, with thinking disabled and the sampling parameters recommended for it (temperature 1, top-$p$ 0.95, top-$k$ 20, presence penalty 1.5). The feedback model sees the question and a single rollout, but never the reference answer, and replies in two parts. The first part is a private verification of at most 8192 tokens, in which it may redo the rollout's computations itself. The second part, of at most 1024 tokens, is feedback to the author: it points to the first step whose conclusion is false or unjustified, explains why, and says what to reconsider, without stating an answer of its own. The solver sees only the second part as $\ft$. When we evaluate the feedback model zero-shot, we give it at most 32768 new tokens per rollout rather than 8192 as it is biased toward longer answers. HMMT uses the same feedback model, prompts, and token limits for the feedback as AIME. See Appendix \ref{app:exp-prompts-math} for the exact prompts we use.

\subsection{Protocol, Metrics, and Statistics}
\label{app:metrics}

We evaluate the solver zero-shot on every benchmark, using the prompt $h^{(1)}$ with $Q = 16$ rollouts per question, and the feedback model zero-shot on AIME and HMMT with $Q = 8$ rollouts per question. We then evaluate the prompting-only versions of Self-Refine and RSA with feedback ($\eta = 0$), and each of them with Bregman tilting at $\alpha = 0$ (Equation \eqref{eq:closed-form}) with $\eta = 48$ and at $\alpha = 1$ with $\eta = 1$. We chose $\eta$ by calibrating its value on a single seed of AIME 2025, balancing divergence to the input-prompting LLM, truncation rates, and performance. We then keep it unchanged for every benchmark and method. At $\alpha = 0$, we find $\hat \lambda$ at each position with 6 bisection steps followed by 6 Newton steps (Appendix \ref{app:pf-lambda}). On LiveCodeBench and MBPP+, the prompts are the code generation prompts of RSA \citep{venkatraman:2025:rsa}, and Self-Refine uses their aggregation prompt with a single candidate (Appendix \ref{app:exp-prompts-code}). The prompts with feedback show each candidate followed by the result of running its program on the exposed tests, and the prompt without feedback $\thisttt$ is the same prompt without these results. We run every method with four seeds on every benchmark. Within a seed, all methods share the same round-1 rollouts and the same feedback on them, so round 1 is the zero-shot solver for every method and the comparisons between methods are paired by question; each seed draws its own round 1. We report means over the four seeds. All standard errors treat the questions of a benchmark as fixed and measure the variation across independent runs on them. For mean pass@1 under Self-Refine, whose chains are independent, we compute them over the 64 chains of the four seeds, and for every other quantity over the four seeds.

On every benchmark, we perform each method for $T = 5$ rounds with $Q = 16$ chains per question, and then compute metrics on each round's $Q$ rollouts, averaged over the questions of the benchmark:
\begin{itemize}
    \item \textbf{mean pass@1}: the fraction of the $Q$ rollouts that are correct;
    \item \textbf{pass@$Q$}: whether at least one of the $Q$ rollouts is correct.
\end{itemize}
Both metrics are computed directly from the $Q$ rollouts of a round. Under Self-Refine, the $Q$ chains of a question are independent, so pass@$Q$ is the usual pass@$k$ with $k = Q$. Under RSA, the chains share rollouts through their prompts after round 1, so pass@$Q$ instead measures how many questions the population that RSA produces covers.

\subsection{Prompts}
\label{app:exp-prompts}
See below for the various prompts used in our experiments. LiveCodeBench and MBPP+ use the prompts of Appendix \ref{app:exp-prompts-code}, and AIME and HMMT use the same prompts (Appendix \ref{app:exp-prompts-math}). None of the prompts uses a system prompt, and the question, rollouts, and feedback are inserted as plain text.

\subsubsection{Code}
\label{app:exp-prompts-code}

The LiveCodeBench and MBPP+ prompts are the code generation prompts used in prior work \citep{venkatraman:2025:rsa}. Each ends with a format instruction, written $\mathit{fmt}$ below. On LiveCodeBench it is the format instruction of the LiveCodeBench evaluation code; for a problem whose program reads standard input, it is
\mypromptbox{
\small
Read the inputs from stdin solve the problem and write the answer to stdout (do not directly test on the sample inputs). Enclose your code within delimiters as follows. Ensure that when the python program runs, it reads the inputs, runs the algorithm and writes output to STDOUT.

\medskip
\texttt{\textasciigrave\textasciigrave\textasciigrave{}python}\\
\texttt{\# YOUR CODE HERE}\\
\texttt{\textasciigrave\textasciigrave\textasciigrave}
}
and for a problem whose program completes a given function, it is
\mypromptbox{
\small
You will use the following starter code to write the solution to the problem and enclose your code within delimiters.\\
\texttt{\textasciigrave\textasciigrave\textasciigrave{}python}\\
\textit{starter code}\\
\texttt{\textasciigrave\textasciigrave\textasciigrave}
}
On MBPP+, $\mathit{fmt}$ shows the task's hand-written test assertions and names the function to define:
\mypromptbox{
\small
Your code should pass these tests:

\medskip
\texttt{\textasciigrave\textasciigrave\textasciigrave{}python}\\
\textit{the task's test assertions, one per line}\\
\texttt{\textasciigrave\textasciigrave\textasciigrave}

\medskip
Enclose your code within delimiters as follows. Ensure that the code block defines the function \texttt{\textasciigrave}\textit{function name}\texttt{\textasciigrave} the tests call, with every import and helper it needs.

\medskip
\texttt{\textasciigrave\textasciigrave\textasciigrave{}python}\\
\texttt{\# YOUR CODE HERE}\\
\texttt{\textasciigrave\textasciigrave\textasciigrave}
}
The question of an MBPP+ task is its one-sentence description.

\paragraph{Zero-shot.} Round 1 of every method, and the zero-shot baseline, use the prompt $h^{(1)}$:
\mypromptbox{
\small
\textbf{User:}\quad
You will be given a question (problem specification) and will generate a correct Python program that matches the specification and passes all tests.

\medskip
Question:\\
$\mathit{question}$

\medskip
$\mathit{fmt}$
}

\paragraph{Self-Refine and RSA.} The prompt without feedback, $\thisttt$, is the aggregation prompt of RSA \citep{venkatraman:2025:rsa}. RSA shows $K = 4$ rollouts drawn from the previous round. Self-Refine uses the same prompt with $K = 1$, where the one rollout is the chain's own rollout $\rollt$.
\mypromptbox{
\small
\textbf{User:}\quad
You are given a python code implementation problem and several candidate code blocks with their reasoning. Some candidates may be incorrect or contain errors. Aggregate the useful ideas and produce a single, high-quality solution. Reason carefully; if candidates disagree, choose the correct path.\\
You will be given a question (problem specification) and will generate a correct Python program that matches the specification and passes all tests.

\medskip
Question:\\
$\mathit{question}$

\medskip
Candidate solutions (may contain mistakes):

\medskip
\texttt{-{}-{}-{}-} Solution 1 \texttt{-{}-{}-{}-}\\
\textit{rollout 1}

\medskip
$\vdots$

\medskip
\texttt{-{}-{}-{}-} Solution $K$ \texttt{-{}-{}-{}-}\\
\textit{rollout $K$}

\medskip
Now provide an improved and correct solution along with its reasoning.

\medskip
$\mathit{fmt}$
}
The prompt with feedback, $\histtt$, shows each of the same rollouts followed by its feedback. The text that appears only in this prompt is in bold:
\mypromptbox{
\small
\textbf{User:}\quad
You are given a python code implementation problem and several candidate code blocks with their reasoning\textbf{, each followed by the result of running its program on the problem's example tests}. Some candidates may be incorrect or contain errors. \textbf{Use the test results to tell which ideas are sound, aggregate the useful ones} and produce a single, high-quality solution. Reason carefully; if candidates disagree, choose the correct path.\\
You will be given a question (problem specification) and will generate a correct Python program that matches the specification and passes all tests.

\medskip
Question:\\
$\mathit{question}$

\medskip
Candidate solutions (may contain mistakes):

\medskip
\texttt{-{}-{}-{}-} Solution 1 \texttt{-{}-{}-{}-}\\
\textit{rollout 1}

\medskip
\textbf{\texttt{-{}-{}-{}-} Result of running solution 1 on the example tests \texttt{-{}-{}-{}-}}\\
\textbf{\textit{feedback on rollout 1}}

\medskip
$\vdots$

\medskip
\texttt{-{}-{}-{}-} Solution $K$ \texttt{-{}-{}-{}-}\\
\textit{rollout $K$}

\medskip
\textbf{\texttt{-{}-{}-{}-} Result of running solution $\boldsymbol{K}$ on the example tests \texttt{-{}-{}-{}-}}\\
\textbf{\textit{feedback on rollout $\boldsymbol{K}$}}

\medskip
Now provide an improved and correct solution along with its reasoning.

\medskip
$\mathit{fmt}$
}

\paragraph{Execution feedback.} The feedback $\ft = \Eval(\rollt)$ is one of the following messages, where the example tests are the public tests of the problem together with the added half of its private tests (Appendix \ref{app:feedback}), and $n$ is their number. If the program passes every example test, the feedback is
\mypromptbox{
\small
All $n$ example tests passed. The hidden tests remain, and the examples are few and small: a program that passes them can still be wrong on edge cases or too slow for the largest inputs allowed by the constraints. Re-check the constraints, boundary cases and time complexity before keeping this solution.
}
Otherwise, it reports the first example test $i$ that the program fails, with the number $m$ of example tests it passes. For a wrong answer, the feedback is
\mypromptbox{
\small
$m$ of $n$ example tests passed. Example test $i$: Wrong Answer

\medskip
Input\\
\textit{test input}

\medskip
Output\\
\textit{output of the program}

\medskip
Expected\\
\textit{expected output}
}
For a runtime error, the first line is the same with ``Runtime Error'' in place of ``Wrong Answer'', followed by the error message, ``Line $\ell$ of your program'' for the line $\ell$ that raised it, and the test input. For a time limit, it is the same with ``Time Limit Exceeded (6 s)'', followed by the test input. If the program does not compile, the feedback is ``Compilation Error (line $\ell$)'' followed by the error message. If the rollout has no code block, the feedback is
\mypromptbox{
\small
Incorrect Format: no code block was found in your previous response, so no program could be tested. Your complete program must appear in a single \texttt{\textasciigrave\textasciigrave\textasciigrave{}python ... \textasciigrave\textasciigrave\textasciigrave} code block, in the format the question specifies (reading standard input and writing standard output, or completing the given starter code).
}
and if the rollout was truncated before a complete code block, the feedback is
\mypromptbox{
\small
Truncated Attempt: your previous response ran past the length limit before a complete code block appeared, so no program could be tested. Keep the reasoning brief and finish with the complete program. Your complete program must appear in a single \texttt{\textasciigrave\textasciigrave\textasciigrave{}python ... \textasciigrave\textasciigrave\textasciigrave} code block, in the format the question specifies (reading standard input and writing standard output, or completing the given starter code).
}

\subsubsection{Math}
\label{app:exp-prompts-math}

\paragraph{Zero-shot.} Round 1 of every method, and the zero-shot baselines, use the prompt $h^{(1)}$:
\mypromptbox{
\small
\textbf{User:}\quad
$\mathit{question}$

\medskip
Put your final answer within \texttt{\textbackslash boxed\{\}}.
}

\paragraph{Self-Refine.} The Self-Refine prompt without feedback, $\thisttt$, is the self-refinement prompt of RSA \citep{venkatraman:2025:rsa}:
\mypromptbox{
\small
\textbf{User:}\quad
You are given a math problem and a candidate solution. The candidate may be incomplete or contain errors. Refine this trajectory and produce an improved, higher-quality solution. If it is entirely wrong, attempt a new strategy. End with the final result in \texttt{\textbackslash boxed\{\}}.

\medskip
Problem:\\
$\mathit{question}$

\medskip
Candidate solution (may contain mistakes):\\
\texttt{-{}-{}-{}-} Candidate \texttt{-{}-{}-{}-}\\
$\rollt$

\medskip
Now refine the candidate into an improved solution. Provide clear reasoning and end with the final answer in \texttt{\textbackslash boxed\{\}}.
}
The Self-Refine prompt with feedback, $\histtt$, shows the same rollout followed by its feedback. The text that appears only in this prompt is in bold:
\mypromptbox{
\small
\textbf{User:}\quad
You are given a math problem, a candidate solution\textbf{, and a reviewer's feedback on that candidate}. The candidate may be incomplete or contain errors\textbf{, and the reviewer may or may not have found a problem with it}. \textbf{Use the feedback to} refine this trajectory and produce an improved, higher-quality solution. If \textbf{the candidate} is entirely wrong, attempt a new strategy. End with the final result in \texttt{\textbackslash boxed\{\}}.

\medskip
Problem:\\
$\mathit{question}$

\medskip
Candidate solution (may contain mistakes):\\
\texttt{-{}-{}-{}-} Candidate \texttt{-{}-{}-{}-}\\
$\rollt$

\medskip
\textbf{\texttt{-{}-{}-{}-} Reviewer's feedback on the candidate \texttt{-{}-{}-{}-}}\\
$\bm{f^{(t)}}$

\medskip
Now refine the candidate into an improved solution \textbf{that addresses the feedback}. Provide clear reasoning and end with the final answer in \texttt{\textbackslash boxed\{\}}.
}

\paragraph{RSA.} Without feedback, we use the aggregation prompt of RSA \citep{venkatraman:2025:rsa}. At each round after the first, it shows $K = 4$ rollouts drawn from the previous round:
\mypromptbox{
\small
\textbf{User:}\quad
You are given a math problem and several candidate solutions. Some candidates may be incorrect or contain errors. Aggregate the useful ideas and produce a single, high-quality solution. Reason carefully; if candidates disagree, choose the correct path. If all are incorrect, then attempt a different strategy. End with the final result in \texttt{\textbackslash boxed\{\}}.

\medskip
Problem:\\
$\mathit{question}$

\medskip
Candidate solutions (may contain mistakes):\\
\texttt{-{}-{}-{}-} Solution 1 \texttt{-{}-{}-{}-}\\
\textit{rollout 1}

\medskip
$\vdots$

\medskip
\texttt{-{}-{}-{}-} Solution $K$ \texttt{-{}-{}-{}-}\\
\textit{rollout $K$}

\medskip
Now write a single improved solution. Provide clear reasoning and end with the final answer in \texttt{\textbackslash boxed\{\}}.
}
The RSA prompt with feedback shows each of the same rollouts followed by the feedback model's reply on it. The text that appears only in this prompt is in bold:
\mypromptbox{
\small
\textbf{User:}\quad
You are given a math problem, several candidate solutions\textbf{, and a reviewer's feedback on each candidate}. Some candidates may be incorrect or contain errors\textbf{, and the reviewer may or may not have found a problem with each one}. \textbf{Use the feedback to judge the candidates,} aggregate the useful ideas\textbf{,} and produce a single, high-quality solution. Reason carefully; if candidates disagree, choose the correct path. If all are incorrect, then attempt a different strategy. End with the final result in \texttt{\textbackslash boxed\{\}}.

\medskip
Problem:\\
$\mathit{question}$

\medskip
Candidate solutions (may contain mistakes)\textbf{, each followed by the reviewer's feedback on it}:\\
\texttt{-{}-{}-{}-} Solution 1 \texttt{-{}-{}-{}-}\\
\textit{rollout 1}

\medskip
\textbf{\texttt{-{}-{}-{}-} Reviewer's feedback on Solution 1 \texttt{-{}-{}-{}-}}\\
\textbf{\textit{feedback on rollout 1}}

\medskip
$\vdots$

\medskip
\texttt{-{}-{}-{}-} Solution $K$ \texttt{-{}-{}-{}-}\\
\textit{rollout $K$}

\medskip
\textbf{\texttt{-{}-{}-{}-} Reviewer's feedback on Solution $\boldsymbol{K}$ \texttt{-{}-{}-{}-}}\\
\textbf{\textit{feedback on rollout $\boldsymbol{K}$}}

\medskip
Now write a single improved solution \textbf{that addresses the feedback}. Provide clear reasoning and end with the final answer in \texttt{\textbackslash boxed\{\}}.
}

\paragraph{Feedback model.} The feedback model $\Eval(\rollt)$ reviews one rollout at a time with the following prompt:
\mypromptbox{
\small
\textbf{User:}\quad
You are reviewing a candidate solution to a math problem. Your job is to decide whether its final answer is correct and, if not, to tell the author where it went wrong. You are not told whether the candidate is correct. The candidate may also have been cut off before finishing.

\medskip
Work in two parts.

\medskip
PART 1: VERIFICATION. This part is for you alone; the author will not see it. Check the candidate as carefully as you need to, including redoing computations and derivations yourself.
\begin{itemize}[leftmargin=*,noitemsep,topsep=0pt,label=-]
    \item Restate what the problem asks and check that the candidate solves that problem, not a misreading of it.
    \item Go through the candidate's steps in order. For each important step, verify that its conclusion is true and follows from what came before. Recompute the key quantities rather than taking them on trust.
    \item Check for missed cases, unjustified assumptions and sign or counting errors.
    \item Check the final answer against the problem's constraints and the form it requires, by substitution or an independent route where possible.
\end{itemize}
A step that is terse but true is not an error. An error is a conclusion that is false, or a claim the solution depends on that does not follow.\\
Stop rule: as soon as you have confirmed one error that the final answer depends on, and that the candidate does not correct later in the solution, stop verifying and go to Part 2. Everything after a broken step is moot, and you do not need to work out the right answer yourself.

\medskip
PART 2: FEEDBACK TO THE AUTHOR. The author will see only this part. Do not state, hint at, or box a final answer of your own, and do not write out a full alternative solution; the author must redo the work. Point to the first step whose conclusion is false or unjustified (by the claim it makes, not by a step number), explain why it fails, and say what the author should check or reconsider there. If you found no error, say so and name the step most worth double-checking.

\medskip
Finish with exactly one line in this form:\\
Verdict: correct \textbar{} wrong \textbar{} missing\\
``correct'' means you verified the boxed answer; ``wrong'' means you believe it is not the right answer; ``missing'' means the candidate never reaches a final answer. Give ``correct'' only if Part 1 actually reproduced the key steps.

\medskip
Problem:\\
$\mathit{question}$

\medskip
Candidate solution:\\
$\rollt$

\medskip
Begin with ``PART 1: VERIFICATION''.
}
We generate the reply in two stages. The first stage generates Part 1 until the model begins Part 2 or reaches 8192 tokens. We then append the header ``PART 2: FEEDBACK TO THE AUTHOR'' and generate Part 2 for at most 1024 tokens. The feedback $\ft$ that the solver sees is Part 2 with its verdict line removed, so the solver reads the critique but not the feedback model's final judgment.

\clearpage
\subsection{Results by Round}
\label{app:by-round}

Tables \ref{tab:lcb-results}, \ref{tab:mbpp-results}, \ref{tab:main-results}, and \ref{tab:hmmt-results} give mean pass@1 and pass@16 at every round for the runs of Table \ref{tab:results}. On every benchmark and for both input-prompting methods, Bregman tilting at either value of $\alpha$ improves over input prompting at every round in both metrics.

\begin{table}[H]
\centering
\footnotesize
\setlength{\tabcolsep}{4pt}
\definecolor{tiltshade}{RGB}{251,231,214}
\setlength{\aboverulesep}{0pt}
\setlength{\belowrulesep}{0pt}
\renewcommand{\arraystretch}{1.35}
\caption{Mean pass@1 and pass@16 (\%) by round on LiveCodeBench, averaged over four seeds, with $Q = 16$ rollouts per question, along with the standard error. Zero-shot is the round-1 solver which all methods share within a seed. The shaded columns correspond to our {\setlength{\fboxsep}{1.5pt}\colorbox{tiltshade}{Bregman tilting}}. The best score in each row and metric is highlighted in \textcolor{green!50!black}{\textbf{green}}.}
\label{tab:lcb-results}
\begin{tabular}{llc>{\columncolor{tiltshade}}c>{\columncolor{tiltshade}}cc>{\columncolor{tiltshade}}c>{\columncolor{tiltshade}}c}
\toprule
 & & \multicolumn{3}{c}{Pass@1} & \multicolumn{3}{c}{Pass@16} \\
\cmidrule(lr){3-5} \cmidrule(lr){6-8}
\addlinespace[2pt]
 & & & \multicolumn{2}{c}{{\setlength{\fboxsep}{2.5pt}\colorbox{tiltshade}{Bregman tilting}}} & & \multicolumn{2}{c}{{\setlength{\fboxsep}{2.5pt}\colorbox{tiltshade}{Bregman tilting}}} \\
\cmidrule(lr){4-5} \cmidrule(lr){7-8}
\addlinespace[2pt]
\rowcolor{white}
Method & Round & Prompting & $\alpha = 1$ & $\alpha = 0$ & Prompting & $\alpha = 1$ & $\alpha = 0$ \\
\midrule
\multicolumn{2}{l}{Zero-shot (round 1)} & 34.1\,{\scriptsize$\pm$0.2} & \multicolumn{2}{c}{} & 47.5\,{\scriptsize$\pm$0.7} & \multicolumn{2}{c}{} \\
\midrule
\multirow{4}{*}{Self-Refine} & 2 & 36.2\,{\scriptsize$\pm$0.2} & 37.2\,{\scriptsize$\pm$0.2} & \textcolor{green!50!black}{\textbf{37.9}}\,{\scriptsize$\pm$0.2} & 48.9\,{\scriptsize$\pm$0.7} & 51.3\,{\scriptsize$\pm$0.7} & \textcolor{green!50!black}{\textbf{51.7}}\,{\scriptsize$\pm$0.6} \\
 & 3 & 37.3\,{\scriptsize$\pm$0.2} & 38.1\,{\scriptsize$\pm$0.2} & \textcolor{green!50!black}{\textbf{39.0}}\,{\scriptsize$\pm$0.2} & 50.8\,{\scriptsize$\pm$0.8} & 52.9\,{\scriptsize$\pm$0.7} & \textcolor{green!50!black}{\textbf{54.2}}\,{\scriptsize$\pm$0.7} \\
 & 4 & 38.0\,{\scriptsize$\pm$0.2} & 39.4\,{\scriptsize$\pm$0.2} & \textcolor{green!50!black}{\textbf{40.3}}\,{\scriptsize$\pm$0.2} & 51.5\,{\scriptsize$\pm$0.7} & 53.6\,{\scriptsize$\pm$0.8} & \textcolor{green!50!black}{\textbf{56.5}}\,{\scriptsize$\pm$0.7} \\
 & 5 & 38.6\,{\scriptsize$\pm$0.2} & 39.9\,{\scriptsize$\pm$0.2} & \textcolor{green!50!black}{\textbf{41.0}}\,{\scriptsize$\pm$0.3} & 52.1\,{\scriptsize$\pm$0.7} & 55.5\,{\scriptsize$\pm$0.8} & \textcolor{green!50!black}{\textbf{59.0}}\,{\scriptsize$\pm$0.7} \\
\midrule
\multirow{4}{*}{RSA} & 2 & 38.3\,{\scriptsize$\pm$0.3} & 40.0\,{\scriptsize$\pm$0.3} & \textcolor{green!50!black}{\textbf{40.4}}\,{\scriptsize$\pm$0.3} & 48.3\,{\scriptsize$\pm$0.8} & 49.2\,{\scriptsize$\pm$0.8} & \textcolor{green!50!black}{\textbf{50.4}}\,{\scriptsize$\pm$0.8} \\
 & 3 & 40.5\,{\scriptsize$\pm$0.4} & 42.9\,{\scriptsize$\pm$0.4} & \textcolor{green!50!black}{\textbf{43.3}}\,{\scriptsize$\pm$0.4} & 48.1\,{\scriptsize$\pm$0.7} & 49.4\,{\scriptsize$\pm$0.8} & \textcolor{green!50!black}{\textbf{50.4}}\,{\scriptsize$\pm$0.8} \\
 & 4 & 42.4\,{\scriptsize$\pm$0.4} & 44.8\,{\scriptsize$\pm$0.5} & \textcolor{green!50!black}{\textbf{45.2}}\,{\scriptsize$\pm$0.5} & 48.1\,{\scriptsize$\pm$0.7} & 49.2\,{\scriptsize$\pm$0.8} & \textcolor{green!50!black}{\textbf{51.0}}\,{\scriptsize$\pm$0.8} \\
 & 5 & 43.6\,{\scriptsize$\pm$0.5} & 45.9\,{\scriptsize$\pm$0.6} & \textcolor{green!50!black}{\textbf{47.0}}\,{\scriptsize$\pm$0.6} & 47.9\,{\scriptsize$\pm$0.8} & 49.6\,{\scriptsize$\pm$0.8} & \textcolor{green!50!black}{\textbf{51.5}}\,{\scriptsize$\pm$0.8} \\
\bottomrule
\end{tabular}
\end{table}

\begin{table}[H]
\centering
\footnotesize
\setlength{\tabcolsep}{4pt}
\definecolor{tiltshade}{RGB}{251,231,214}
\setlength{\aboverulesep}{0pt}
\setlength{\belowrulesep}{0pt}
\renewcommand{\arraystretch}{1.35}
\caption{Mean pass@1 and pass@16 (\%) by round on the MBPP+ hard set, averaged over four seeds, with $Q = 16$ rollouts per question, along with the standard error. Zero-shot is the round-1 solver which all methods share within a seed. The shaded columns correspond to our {\setlength{\fboxsep}{1.5pt}\colorbox{tiltshade}{Bregman tilting}}. The best score in each row and metric is highlighted in \textcolor{green!50!black}{\textbf{green}}.}
\label{tab:mbpp-results}
\begin{tabular}{llc>{\columncolor{tiltshade}}c>{\columncolor{tiltshade}}cc>{\columncolor{tiltshade}}c>{\columncolor{tiltshade}}c}
\toprule
 & & \multicolumn{3}{c}{Pass@1} & \multicolumn{3}{c}{Pass@16} \\
\cmidrule(lr){3-5} \cmidrule(lr){6-8}
\addlinespace[2pt]
 & & & \multicolumn{2}{c}{{\setlength{\fboxsep}{2.5pt}\colorbox{tiltshade}{Bregman tilting}}} & & \multicolumn{2}{c}{{\setlength{\fboxsep}{2.5pt}\colorbox{tiltshade}{Bregman tilting}}} \\
\cmidrule(lr){4-5} \cmidrule(lr){7-8}
\addlinespace[2pt]
\rowcolor{white}
Method & Round & Prompting & $\alpha = 1$ & $\alpha = 0$ & Prompting & $\alpha = 1$ & $\alpha = 0$ \\
\midrule
\multicolumn{2}{l}{Zero-shot (round 1)} & 40.8\,{\scriptsize$\pm$0.3} & \multicolumn{2}{c}{} & 62.3\,{\scriptsize$\pm$0.7} & \multicolumn{2}{c}{} \\
\midrule
\multirow{4}{*}{Self-Refine} & 2 & 55.4\,{\scriptsize$\pm$0.3} & 59.8\,{\scriptsize$\pm$0.3} & \textcolor{green!50!black}{\textbf{62.0}}\,{\scriptsize$\pm$0.3} & 76.1\,{\scriptsize$\pm$0.7} & 79.4\,{\scriptsize$\pm$0.5} & \textcolor{green!50!black}{\textbf{80.0}}\,{\scriptsize$\pm$0.5} \\
 & 3 & 59.4\,{\scriptsize$\pm$0.3} & 63.4\,{\scriptsize$\pm$0.3} & \textcolor{green!50!black}{\textbf{65.5}}\,{\scriptsize$\pm$0.3} & 79.5\,{\scriptsize$\pm$0.6} & 80.8\,{\scriptsize$\pm$0.4} & \textcolor{green!50!black}{\textbf{83.2}}\,{\scriptsize$\pm$0.5} \\
 & 4 & 62.4\,{\scriptsize$\pm$0.3} & 66.0\,{\scriptsize$\pm$0.3} & \textcolor{green!50!black}{\textbf{67.5}}\,{\scriptsize$\pm$0.3} & 81.6\,{\scriptsize$\pm$0.5} & 82.6\,{\scriptsize$\pm$0.5} & \textcolor{green!50!black}{\textbf{84.5}}\,{\scriptsize$\pm$0.5} \\
 & 5 & 64.3\,{\scriptsize$\pm$0.3} & 67.1\,{\scriptsize$\pm$0.3} & \textcolor{green!50!black}{\textbf{67.8}}\,{\scriptsize$\pm$0.3} & 83.4\,{\scriptsize$\pm$0.5} & 83.5\,{\scriptsize$\pm$0.5} & \textcolor{green!50!black}{\textbf{85.0}}\,{\scriptsize$\pm$0.5} \\
\midrule
\multirow{4}{*}{RSA} & 2 & 59.4\,{\scriptsize$\pm$0.4} & 63.1\,{\scriptsize$\pm$0.3} & \textcolor{green!50!black}{\textbf{65.0}}\,{\scriptsize$\pm$0.3} & 74.8\,{\scriptsize$\pm$0.7} & 77.4\,{\scriptsize$\pm$0.6} & \textcolor{green!50!black}{\textbf{77.9}}\,{\scriptsize$\pm$0.5} \\
 & 3 & 65.9\,{\scriptsize$\pm$0.5} & 71.0\,{\scriptsize$\pm$0.4} & \textcolor{green!50!black}{\textbf{73.0}}\,{\scriptsize$\pm$0.3} & 77.4\,{\scriptsize$\pm$0.7} & 80.2\,{\scriptsize$\pm$0.5} & \textcolor{green!50!black}{\textbf{80.8}}\,{\scriptsize$\pm$0.5} \\
 & 4 & 70.2\,{\scriptsize$\pm$0.5} & 75.4\,{\scriptsize$\pm$0.5} & \textcolor{green!50!black}{\textbf{76.5}}\,{\scriptsize$\pm$0.4} & 77.9\,{\scriptsize$\pm$0.6} & 80.8\,{\scriptsize$\pm$0.6} & \textcolor{green!50!black}{\textbf{81.9}}\,{\scriptsize$\pm$0.4} \\
 & 5 & 72.9\,{\scriptsize$\pm$0.6} & 77.5\,{\scriptsize$\pm$0.5} & \textcolor{green!50!black}{\textbf{78.9}}\,{\scriptsize$\pm$0.4} & 77.7\,{\scriptsize$\pm$0.5} & 81.1\,{\scriptsize$\pm$0.6} & \textcolor{green!50!black}{\textbf{82.6}}\,{\scriptsize$\pm$0.4} \\
\bottomrule
\end{tabular}
\end{table}

\begin{table}[H]
\centering
\footnotesize
\setlength{\tabcolsep}{4pt}
\definecolor{tiltshade}{RGB}{251,231,214}
\setlength{\aboverulesep}{0pt}
\setlength{\belowrulesep}{0pt}
\renewcommand{\arraystretch}{1.35}
\caption{Mean pass@1 and pass@16 (\%) by round on AIME 2025, averaged over four seeds, with $Q = 16$ rollouts per question, along with the standard error. Zero-shot is the round-1 solver which all methods share within a seed. The shaded columns correspond to our {\setlength{\fboxsep}{1.5pt}\colorbox{tiltshade}{Bregman tilting}}. The best score in each row and metric is highlighted in \textcolor{green!50!black}{\textbf{green}}. For reference, the zero-shot mean pass@1 of the Qwen3.5-27B feedback model is 75.8\%.}
\label{tab:main-results}
\begin{tabular}{llc>{\columncolor{tiltshade}}c>{\columncolor{tiltshade}}cc>{\columncolor{tiltshade}}c>{\columncolor{tiltshade}}c}
\toprule
 & & \multicolumn{3}{c}{Pass@1} & \multicolumn{3}{c}{Pass@16} \\
\cmidrule(lr){3-5} \cmidrule(lr){6-8}
\addlinespace[2pt]
 & & & \multicolumn{2}{c}{{\setlength{\fboxsep}{2.5pt}\colorbox{tiltshade}{Bregman tilting}}} & & \multicolumn{2}{c}{{\setlength{\fboxsep}{2.5pt}\colorbox{tiltshade}{Bregman tilting}}} \\
\cmidrule(lr){4-5} \cmidrule(lr){7-8}
\addlinespace[2pt]
\rowcolor{white}
Method & Round & Prompting & $\alpha = 1$ & $\alpha = 0$ & Prompting & $\alpha = 1$ & $\alpha = 0$ \\
\midrule
\multicolumn{2}{l}{Zero-shot (round 1)} & 41.4\,{\scriptsize$\pm$0.6} & \multicolumn{2}{c}{} & 67.5\,{\scriptsize$\pm$2.0} & \multicolumn{2}{c}{} \\
\midrule
\multirow{4}{*}{Self-Refine} & 2 & 56.2\,{\scriptsize$\pm$0.7} & 56.7\,{\scriptsize$\pm$0.7} & \textcolor{green!50!black}{\textbf{57.3}}\,{\scriptsize$\pm$0.7} & 81.7\,{\scriptsize$\pm$1.9} & 83.3\,{\scriptsize$\pm$1.9} & \textcolor{green!50!black}{\textbf{84.2}}\,{\scriptsize$\pm$2.0} \\
 & 3 & 62.6\,{\scriptsize$\pm$0.7} & 65.1\,{\scriptsize$\pm$0.7} & \textcolor{green!50!black}{\textbf{65.7}}\,{\scriptsize$\pm$0.7} & 84.2\,{\scriptsize$\pm$1.7} & 87.5\,{\scriptsize$\pm$1.4} & \textcolor{green!50!black}{\textbf{88.3}}\,{\scriptsize$\pm$1.5} \\
 & 4 & 67.2\,{\scriptsize$\pm$0.6} & 69.9\,{\scriptsize$\pm$0.7} & \textcolor{green!50!black}{\textbf{70.6}}\,{\scriptsize$\pm$0.7} & 86.7\,{\scriptsize$\pm$1.5} & 90.0\,{\scriptsize$\pm$1.2} & \textcolor{green!50!black}{\textbf{90.8}}\,{\scriptsize$\pm$1.3} \\
 & 5 & 70.3\,{\scriptsize$\pm$0.6} & 71.8\,{\scriptsize$\pm$0.6} & \textcolor{green!50!black}{\textbf{73.2}}\,{\scriptsize$\pm$0.6} & 87.5\,{\scriptsize$\pm$1.6} & 90.0\,{\scriptsize$\pm$1.4} & \textcolor{green!50!black}{\textbf{92.5}}\,{\scriptsize$\pm$0.8} \\
\midrule
\multirow{4}{*}{RSA} & 2 & 59.4\,{\scriptsize$\pm$0.9} & 62.7\,{\scriptsize$\pm$1.0} & \textcolor{green!50!black}{\textbf{64.0}}\,{\scriptsize$\pm$0.9} & 75.8\,{\scriptsize$\pm$1.9} & \textcolor{green!50!black}{\textbf{83.3}}\,{\scriptsize$\pm$1.7} & 82.5\,{\scriptsize$\pm$1.4} \\
 & 3 & 71.5\,{\scriptsize$\pm$1.3} & \textcolor{green!50!black}{\textbf{76.7}}\,{\scriptsize$\pm$0.9} & 76.5\,{\scriptsize$\pm$1.2} & 83.3\,{\scriptsize$\pm$1.8} & 85.8\,{\scriptsize$\pm$1.3} & \textcolor{green!50!black}{\textbf{86.7}}\,{\scriptsize$\pm$1.5} \\
 & 4 & 76.7\,{\scriptsize$\pm$1.3} & 81.5\,{\scriptsize$\pm$1.0} & \textcolor{green!50!black}{\textbf{82.8}}\,{\scriptsize$\pm$1.2} & 84.2\,{\scriptsize$\pm$1.4} & 88.3\,{\scriptsize$\pm$1.5} & \textcolor{green!50!black}{\textbf{90.0}}\,{\scriptsize$\pm$1.4} \\
 & 5 & 80.7\,{\scriptsize$\pm$0.9} & 84.5\,{\scriptsize$\pm$1.0} & \textcolor{green!50!black}{\textbf{85.9}}\,{\scriptsize$\pm$1.2} & 85.8\,{\scriptsize$\pm$1.6} & 89.2\,{\scriptsize$\pm$1.4} & \textcolor{green!50!black}{\textbf{90.0}}\,{\scriptsize$\pm$1.5} \\
\bottomrule
\end{tabular}
\end{table}

\begin{table}[H]
\centering
\footnotesize
\setlength{\tabcolsep}{4pt}
\definecolor{tiltshade}{RGB}{251,231,214}
\setlength{\aboverulesep}{0pt}
\setlength{\belowrulesep}{0pt}
\renewcommand{\arraystretch}{1.35}
\caption{Mean pass@1 and pass@16 (\%) by round on HMMT February 2025, averaged over four seeds, with $Q = 16$ rollouts per question, along with the standard error. Zero-shot is the round-1 solver which all methods share within a seed. The shaded columns correspond to our {\setlength{\fboxsep}{1.5pt}\colorbox{tiltshade}{Bregman tilting}}. The best score in each row and metric is highlighted in \textcolor{green!50!black}{\textbf{green}}. For reference, the zero-shot mean pass@1 of the Qwen3.5-27B feedback model is 68.3\%.}
\label{tab:hmmt-results}
\begin{tabular}{llc>{\columncolor{tiltshade}}c>{\columncolor{tiltshade}}cc>{\columncolor{tiltshade}}c>{\columncolor{tiltshade}}c}
\toprule
 & & \multicolumn{3}{c}{Pass@1} & \multicolumn{3}{c}{Pass@16} \\
\cmidrule(lr){3-5} \cmidrule(lr){6-8}
\addlinespace[2pt]
 & & & \multicolumn{2}{c}{{\setlength{\fboxsep}{2.5pt}\colorbox{tiltshade}{Bregman tilting}}} & & \multicolumn{2}{c}{{\setlength{\fboxsep}{2.5pt}\colorbox{tiltshade}{Bregman tilting}}} \\
\cmidrule(lr){4-5} \cmidrule(lr){7-8}
\addlinespace[2pt]
\rowcolor{white}
Method & Round & Prompting & $\alpha = 1$ & $\alpha = 0$ & Prompting & $\alpha = 1$ & $\alpha = 0$ \\
\midrule
\multicolumn{2}{l}{Zero-shot (round 1)} & 29.8\,{\scriptsize$\pm$0.5} & \multicolumn{2}{c}{} & 52.5\,{\scriptsize$\pm$2.3} & \multicolumn{2}{c}{} \\
\midrule
\multirow{4}{*}{Self-Refine} & 2 & 38.3\,{\scriptsize$\pm$0.6} & 40.4\,{\scriptsize$\pm$0.7} & \textcolor{green!50!black}{\textbf{40.9}}\,{\scriptsize$\pm$0.7} & 65.0\,{\scriptsize$\pm$1.8} & 66.7\,{\scriptsize$\pm$1.7} & \textcolor{green!50!black}{\textbf{70.0}}\,{\scriptsize$\pm$1.7} \\
 & 3 & 45.2\,{\scriptsize$\pm$0.7} & \textcolor{green!50!black}{\textbf{48.7}}\,{\scriptsize$\pm$0.7} & \textcolor{green!50!black}{\textbf{48.7}}\,{\scriptsize$\pm$0.7} & 70.8\,{\scriptsize$\pm$2.0} & 75.8\,{\scriptsize$\pm$1.9} & \textcolor{green!50!black}{\textbf{80.0}}\,{\scriptsize$\pm$1.9} \\
 & 4 & 52.1\,{\scriptsize$\pm$0.7} & \textcolor{green!50!black}{\textbf{54.5}}\,{\scriptsize$\pm$0.7} & 54.3\,{\scriptsize$\pm$0.7} & 77.5\,{\scriptsize$\pm$1.7} & 80.8\,{\scriptsize$\pm$1.3} & \textcolor{green!50!black}{\textbf{82.5}}\,{\scriptsize$\pm$1.7} \\
 & 5 & 55.3\,{\scriptsize$\pm$0.7} & 57.8\,{\scriptsize$\pm$0.7} & \textcolor{green!50!black}{\textbf{58.2}}\,{\scriptsize$\pm$0.7} & 78.3\,{\scriptsize$\pm$1.5} & \textcolor{green!50!black}{\textbf{83.3}}\,{\scriptsize$\pm$1.5} & \textcolor{green!50!black}{\textbf{83.3}}\,{\scriptsize$\pm$1.5} \\
\midrule
\multirow{4}{*}{RSA} & 2 & 42.2\,{\scriptsize$\pm$0.9} & 45.6\,{\scriptsize$\pm$1.0} & \textcolor{green!50!black}{\textbf{47.0}}\,{\scriptsize$\pm$0.9} & 62.5\,{\scriptsize$\pm$1.7} & 64.2\,{\scriptsize$\pm$1.7} & \textcolor{green!50!black}{\textbf{65.0}}\,{\scriptsize$\pm$1.7} \\
 & 3 & 55.0\,{\scriptsize$\pm$1.2} & 59.8\,{\scriptsize$\pm$1.1} & \textcolor{green!50!black}{\textbf{60.9}}\,{\scriptsize$\pm$1.1} & 67.5\,{\scriptsize$\pm$2.0} & 73.3\,{\scriptsize$\pm$1.8} & \textcolor{green!50!black}{\textbf{76.7}}\,{\scriptsize$\pm$1.5} \\
 & 4 & 63.0\,{\scriptsize$\pm$1.3} & 69.0\,{\scriptsize$\pm$1.2} & \textcolor{green!50!black}{\textbf{69.8}}\,{\scriptsize$\pm$1.0} & 74.2\,{\scriptsize$\pm$1.7} & \textcolor{green!50!black}{\textbf{79.2}}\,{\scriptsize$\pm$1.7} & 78.3\,{\scriptsize$\pm$1.2} \\
 & 5 & 68.7\,{\scriptsize$\pm$1.6} & 74.5\,{\scriptsize$\pm$1.4} & \textcolor{green!50!black}{\textbf{74.6}}\,{\scriptsize$\pm$1.0} & 75.0\,{\scriptsize$\pm$1.9} & 80.8\,{\scriptsize$\pm$1.6} & \textcolor{green!50!black}{\textbf{83.3}}\,{\scriptsize$\pm$1.5} \\
\bottomrule
\end{tabular}
\end{table}

Table \ref{tab:majority-results} gives the majority vote accuracy at every round on AIME 2025 and HMMT February 2025, where a question counts as correct if the most common final answer among its $Q = 16$ rollouts is correct. As with the other two metrics, Bregman tilting at either value of $\alpha$ improves over input prompting at every round for both input-prompting methods.

\begin{table}[H]
\centering
\footnotesize
\setlength{\tabcolsep}{4pt}
\definecolor{tiltshade}{RGB}{251,231,214}
\setlength{\aboverulesep}{0pt}
\setlength{\belowrulesep}{0pt}
\renewcommand{\arraystretch}{1.35}
\caption{Majority vote accuracy (\%) by round on AIME 2025 and HMMT February 2025, averaged over four seeds, with $Q = 16$ rollouts per question, along with the standard error. Ties go to the answer that appears first. Zero-shot is the round-1 solver which all methods share within a seed. The shaded columns correspond to our {\setlength{\fboxsep}{1.5pt}\colorbox{tiltshade}{Bregman tilting}}. The best score in each row and benchmark is highlighted in \textcolor{green!50!black}{\textbf{green}}.}
\label{tab:majority-results}
\begin{tabular}{llc>{\columncolor{tiltshade}}c>{\columncolor{tiltshade}}cc>{\columncolor{tiltshade}}c>{\columncolor{tiltshade}}c}
\toprule
 & & \multicolumn{3}{c}{AIME 2025} & \multicolumn{3}{c}{HMMT Feb 2025} \\
\cmidrule(lr){3-5} \cmidrule(lr){6-8}
\addlinespace[2pt]
 & & & \multicolumn{2}{c}{{\setlength{\fboxsep}{2.5pt}\colorbox{tiltshade}{Bregman tilting}}} & & \multicolumn{2}{c}{{\setlength{\fboxsep}{2.5pt}\colorbox{tiltshade}{Bregman tilting}}} \\
\cmidrule(lr){4-5} \cmidrule(lr){7-8}
\addlinespace[2pt]
\rowcolor{white}
Method & Round & Prompting & $\alpha = 1$ & $\alpha = 0$ & Prompting & $\alpha = 1$ & $\alpha = 0$ \\
\midrule
\multicolumn{2}{l}{Zero-shot (round 1)} & 57.5\,{\scriptsize$\pm$2.5} & \multicolumn{2}{c}{} & 34.2\,{\scriptsize$\pm$1.4} & \multicolumn{2}{c}{} \\
\midrule
\multirow{4}{*}{Self-Refine} & 2 & 70.0\,{\scriptsize$\pm$1.8} & 70.8\,{\scriptsize$\pm$2.0} & \textcolor{green!50!black}{\textbf{72.5}}\,{\scriptsize$\pm$1.7} & 43.3\,{\scriptsize$\pm$2.5} & 52.5\,{\scriptsize$\pm$2.3} & \textcolor{green!50!black}{\textbf{53.3}}\,{\scriptsize$\pm$2.2} \\
 & 3 & 77.5\,{\scriptsize$\pm$1.6} & \textcolor{green!50!black}{\textbf{80.8}}\,{\scriptsize$\pm$1.7} & \textcolor{green!50!black}{\textbf{80.8}}\,{\scriptsize$\pm$0.8} & 54.2\,{\scriptsize$\pm$1.3} & 62.5\,{\scriptsize$\pm$1.7} & \textcolor{green!50!black}{\textbf{64.2}}\,{\scriptsize$\pm$1.9} \\
 & 4 & 80.0\,{\scriptsize$\pm$1.4} & 84.2\,{\scriptsize$\pm$1.4} & \textcolor{green!50!black}{\textbf{85.0}}\,{\scriptsize$\pm$1.5} & 65.8\,{\scriptsize$\pm$1.4} & 66.7\,{\scriptsize$\pm$1.4} & \textcolor{green!50!black}{\textbf{68.3}}\,{\scriptsize$\pm$1.9} \\
 & 5 & 81.7\,{\scriptsize$\pm$1.2} & 84.2\,{\scriptsize$\pm$0.8} & \textcolor{green!50!black}{\textbf{85.8}}\,{\scriptsize$\pm$1.3} & 67.5\,{\scriptsize$\pm$1.4} & \textcolor{green!50!black}{\textbf{70.8}}\,{\scriptsize$\pm$1.7} & \textcolor{green!50!black}{\textbf{70.8}}\,{\scriptsize$\pm$1.4} \\
\midrule
\multirow{4}{*}{RSA} & 2 & 64.2\,{\scriptsize$\pm$2.2} & 69.2\,{\scriptsize$\pm$2.3} & \textcolor{green!50!black}{\textbf{70.8}}\,{\scriptsize$\pm$2.0} & 47.5\,{\scriptsize$\pm$2.2} & 52.5\,{\scriptsize$\pm$2.5} & \textcolor{green!50!black}{\textbf{55.8}}\,{\scriptsize$\pm$2.2} \\
 & 3 & 74.2\,{\scriptsize$\pm$1.9} & \textcolor{green!50!black}{\textbf{83.3}}\,{\scriptsize$\pm$1.7} & 82.5\,{\scriptsize$\pm$1.7} & 58.3\,{\scriptsize$\pm$2.0} & 65.8\,{\scriptsize$\pm$1.7} & \textcolor{green!50!black}{\textbf{66.7}}\,{\scriptsize$\pm$1.5} \\
 & 4 & 78.3\,{\scriptsize$\pm$1.9} & 85.0\,{\scriptsize$\pm$1.5} & \textcolor{green!50!black}{\textbf{87.5}}\,{\scriptsize$\pm$1.6} & 66.7\,{\scriptsize$\pm$2.0} & 72.5\,{\scriptsize$\pm$1.7} & \textcolor{green!50!black}{\textbf{73.3}}\,{\scriptsize$\pm$1.8} \\
 & 5 & 82.5\,{\scriptsize$\pm$0.8} & \textcolor{green!50!black}{\textbf{87.5}}\,{\scriptsize$\pm$1.3} & \textcolor{green!50!black}{\textbf{87.5}}\,{\scriptsize$\pm$1.6} & 70.0\,{\scriptsize$\pm$2.3} & 76.7\,{\scriptsize$\pm$1.9} & \textcolor{green!50!black}{\textbf{77.5}}\,{\scriptsize$\pm$1.3} \\
\bottomrule
\end{tabular}
\end{table}

\subsection{Reference Prompt by Round}
\label{app:reference-by-round}

\begin{figure}[H]
\centering
\includegraphics[width=0.9\linewidth]{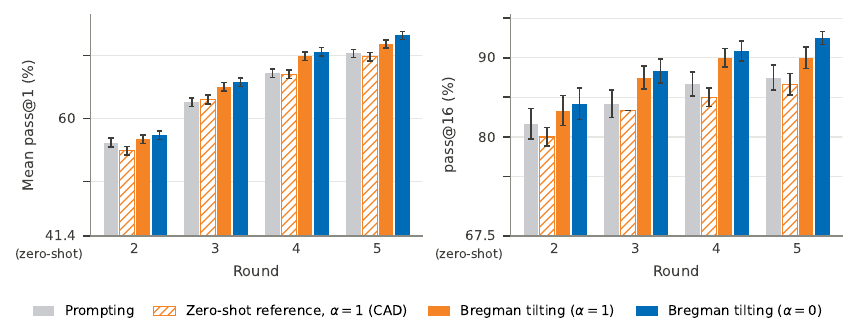}
\caption{Mean pass@1 (left) and pass@16 (right) by round on AIME 2025 with Self-Refine and feedback from Qwen3.5-27B, for input prompting, Bregman tilting at $\alpha = 1$ with the zero-shot reference prompt $h^{(1)}$ (CAD), and Bregman tilting with our no-feedback reference prompt $\thisttt$ at $\alpha = 1$ and $\alpha = 0$. Bars are means over four seeds, with standard error bars. Within each seed, all methods use the same round-1 rollouts and feedback, so each y-axis starts at the round-1 score, marked ``(zero-shot)''. At round 3, the zero-shot reference reaches the same questions at every seed, so its pass@16 has zero standard error.}
\label{fig:reference-by-round}
\end{figure}

Figure \ref{fig:reference-by-round} gives mean pass@1 and pass@16 at every round for the runs of Table \ref{tab:reference}.

\subsection{MBPP+ Full Set}
\label{app:mbpp-full}

Table \ref{tab:mbpp-full-results} gives mean pass@1 and pass@16 at every round on the full set of 378 MBPP+ tasks, for the runs of Table \ref{tab:mbpp-results}. The full set adds the 223 tasks which the solver already solves in all 16 of its seed-0 zero-shot rollouts (Appendix \ref{app:benchmarks}).

\begin{table}[H]
\centering
\footnotesize
\setlength{\tabcolsep}{4pt}
\definecolor{tiltshade}{RGB}{251,231,214}
\setlength{\aboverulesep}{0pt}
\setlength{\belowrulesep}{0pt}
\renewcommand{\arraystretch}{1.35}
\caption{Mean pass@1 and pass@16 (\%) by round on the full set of 378 MBPP+ tasks, averaged over four seeds, with $Q = 16$ rollouts per question, along with the standard error. Zero-shot is the round-1 solver which all methods share within a seed. The shaded columns correspond to our {\setlength{\fboxsep}{1.5pt}\colorbox{tiltshade}{Bregman tilting}}. The best score in each row and metric is highlighted in \textcolor{green!50!black}{\textbf{green}}.}
\label{tab:mbpp-full-results}
\begin{tabular}{llc>{\columncolor{tiltshade}}c>{\columncolor{tiltshade}}cc>{\columncolor{tiltshade}}c>{\columncolor{tiltshade}}c}
\toprule
 & & \multicolumn{3}{c}{Pass@1} & \multicolumn{3}{c}{Pass@16} \\
\cmidrule(lr){3-5} \cmidrule(lr){6-8}
\addlinespace[2pt]
 & & & \multicolumn{2}{c}{{\setlength{\fboxsep}{2.5pt}\colorbox{tiltshade}{Bregman tilting}}} & & \multicolumn{2}{c}{{\setlength{\fboxsep}{2.5pt}\colorbox{tiltshade}{Bregman tilting}}} \\
\cmidrule(lr){4-5} \cmidrule(lr){7-8}
\addlinespace[2pt]
\rowcolor{white}
Method & Round & Prompting & $\alpha = 1$ & $\alpha = 0$ & Prompting & $\alpha = 1$ & $\alpha = 0$ \\
\midrule
\multicolumn{2}{l}{Zero-shot (round 1)} & 75.5\,{\scriptsize$\pm$0.1} & \multicolumn{2}{c}{} & 84.5\,{\scriptsize$\pm$0.3} & \multicolumn{2}{c}{} \\
\midrule
\multirow{4}{*}{Self-Refine} & 2 & 81.3\,{\scriptsize$\pm$0.1} & 83.1\,{\scriptsize$\pm$0.1} & \textcolor{green!50!black}{\textbf{84.0}}\,{\scriptsize$\pm$0.1} & 90.2\,{\scriptsize$\pm$0.3} & 91.5\,{\scriptsize$\pm$0.2} & \textcolor{green!50!black}{\textbf{91.8}}\,{\scriptsize$\pm$0.2} \\
 & 3 & 82.9\,{\scriptsize$\pm$0.1} & 84.1\,{\scriptsize$\pm$0.1} & \textcolor{green!50!black}{\textbf{84.8}}\,{\scriptsize$\pm$0.1} & 91.6\,{\scriptsize$\pm$0.2} & 92.1\,{\scriptsize$\pm$0.2} & \textcolor{green!50!black}{\textbf{93.1}}\,{\scriptsize$\pm$0.2} \\
 & 4 & 84.2\,{\scriptsize$\pm$0.1} & 85.1\,{\scriptsize$\pm$0.1} & \textcolor{green!50!black}{\textbf{85.2}}\,{\scriptsize$\pm$0.1} & 92.5\,{\scriptsize$\pm$0.2} & 92.9\,{\scriptsize$\pm$0.2} & \textcolor{green!50!black}{\textbf{93.6}}\,{\scriptsize$\pm$0.2} \\
 & 5 & 85.0\,{\scriptsize$\pm$0.1} & \textcolor{green!50!black}{\textbf{85.6}}\,{\scriptsize$\pm$0.1} & 85.3\,{\scriptsize$\pm$0.1} & 93.2\,{\scriptsize$\pm$0.2} & 93.3\,{\scriptsize$\pm$0.2} & \textcolor{green!50!black}{\textbf{93.8}}\,{\scriptsize$\pm$0.2} \\
\midrule
\multirow{4}{*}{RSA} & 2 & 83.0\,{\scriptsize$\pm$0.2} & 84.6\,{\scriptsize$\pm$0.1} & \textcolor{green!50!black}{\textbf{85.4}}\,{\scriptsize$\pm$0.1} & 89.7\,{\scriptsize$\pm$0.3} & 90.7\,{\scriptsize$\pm$0.3} & \textcolor{green!50!black}{\textbf{90.9}}\,{\scriptsize$\pm$0.2} \\
 & 3 & 85.7\,{\scriptsize$\pm$0.2} & 87.9\,{\scriptsize$\pm$0.2} & \textcolor{green!50!black}{\textbf{88.8}}\,{\scriptsize$\pm$0.1} & 90.7\,{\scriptsize$\pm$0.3} & 91.9\,{\scriptsize$\pm$0.2} & \textcolor{green!50!black}{\textbf{92.1}}\,{\scriptsize$\pm$0.2} \\
 & 4 & 87.5\,{\scriptsize$\pm$0.2} & 89.8\,{\scriptsize$\pm$0.2} & \textcolor{green!50!black}{\textbf{90.3}}\,{\scriptsize$\pm$0.2} & 90.9\,{\scriptsize$\pm$0.3} & 92.1\,{\scriptsize$\pm$0.2} & \textcolor{green!50!black}{\textbf{92.6}}\,{\scriptsize$\pm$0.2} \\
 & 5 & 88.7\,{\scriptsize$\pm$0.2} & 90.6\,{\scriptsize$\pm$0.2} & \textcolor{green!50!black}{\textbf{91.2}}\,{\scriptsize$\pm$0.2} & 90.9\,{\scriptsize$\pm$0.2} & 92.3\,{\scriptsize$\pm$0.2} & \textcolor{green!50!black}{\textbf{92.9}}\,{\scriptsize$\pm$0.2} \\
\bottomrule
\end{tabular}
\end{table}

\subsection[Runtime Experiment Details]{Runtime Experiment Details}
\label{app:runtime}

We give the details of the runtime comparison of Table \ref{tab:runtime}. The rollouts are the round-2 rollouts of the seed-0 input-prompting Self-Refine run on AIME 2025. For each of them, the prompt $h^{(2)}$ contains the question, the round-1 rollout, and its feedback, and the reference prompt $\thistsup{(2)}$ is the same prompt without the feedback. With one chain, we decode the first chain of each of the 30 questions, one at a time (97{,}138 tokens in total). With 8 chains, we decode the first 8 chains of 10 questions drawn at random, with the 8 chains of a question together in one batch (234{,}424 tokens in total), where a chain leaves the batch when its recorded rollout ends. All methods use the same decoding implementation with key-value caching, each on an NVIDIA H100 NVL GPU. Input prompting decodes $h^{(2)}$ alone, while Bregman tilting decodes $h^{(2)}$ and $\thistsup{(2)}$ together, so each chain takes two sequences of the batch. CAD does the same with the zero-shot prompt $h^{(1)}$, which contains the question alone, in place of $\thistsup{(2)}$. As in our experiments, we use $\eta = 1$ at $\alpha = 1$, and $\eta = 48$ with 6 bisection steps followed by 6 Newton steps for $\hat \lambda$ at $\alpha = 0$ (Appendix \ref{app:metrics}). At every position, each method computes its next-token PMF and samples a token from it, and we then replace the sampled token with the recorded one, so the timed computation is that of Algorithm \ref{alg:bregman-tilting} while the prefixes are identical across methods.

For each method and number of chains, we make one untimed pass over the rollouts followed by two timed passes, and report the mean of the two, which differ by less than 0.3\%. The time per token is the wall-clock decoding time divided by the number of generated tokens, summed over the chains of a batch, and it excludes the initial processing of the prompts, which Table \ref{tab:runtime} reports separately. The fixed cost of a decoding step, which is shared by all sequences in the batch, includes loading the model weights from GPU memory and launching the GPU kernels, while each sequence adds its own attention over its key-value cache and its own next-token PMF. The prompt processing time doubles under Bregman tilting, since the reference prompt is about as long as the prompt, but barely changes under CAD, since its reference prompt is more than 20 times shorter. The key-value cache, which is allocated in advance for a context of 20{,}480 tokens per sequence, doubles under both CAD and Bregman tilting, while the $7.5$ GiB of model weights are stored once.

\clearpage
\printbibliography

\end{document}